\documentclass[onefignum,onetabnum]{siamart220329}

\usepackage{lipsum}
\usepackage{amsfonts}
\usepackage{graphicx}
\usepackage{epstopdf}
\usepackage{algorithmic}
\usepackage{comment}
\usepackage{ulem}
\usepackage{cancel}

\newcommand{\algorithmicnote}{\textbf{Note: }}

\ifpdf
  \DeclareGraphicsExtensions{.eps,.pdf,.png,.jpg}
\else
  \DeclareGraphicsExtensions{.eps}
\fi

\definecolor{red2}{RGB}{204,0,0}
\definecolor{blue2}{RGB}{0,103,165}
\hypersetup{colorlinks,linkcolor=[RGB]{0,103,165},citecolor=[RGB]{180,0,0}}

\newcommand{\TP}[1]{{\color{blue2} #1}}

\newsiamremark{remark}{Remark}
\newsiamremark{hypothesis}{Hypothesis}
\crefname{hypothesis}{Hypothesis}{Hypotheses}
\newsiamthm{claim}{Claim}

\def\be{\begin{equation}}
\def\ee{\end{equation}}

\def\x{\mathbf{x}}
\def\xt{\widetilde{\x}}
\def\y{\mathbf{y}}

\def\f{\mathbf{f}}
\def\z{\mathbf{z}}

\def\I{\mathbf{I}}

\def\xt{\widetilde{\x}}
\def\at{\hat{\mathbf{a}}}
\def\bt{\hat{\mathbf{b}}}
\def\X{\mathbf{X}}

\def\Rs{\mathbb{R}}

\def\G{\mathbf{G}}
\def\Gt{\mathbf{\widetilde{G}}}
\def\N{\mathbf{N}}
\def\P{\mathbb{P}}
\def\Pi{\mathbf{\Phi}}
\def\PPh{\mathbf{\Phi}}

\def\D{\mathbf{D}}
\def\S{\mathbf{S}}
\def\Dt{\widetilde{\D}}
\def\St{\widetilde{\S}}

\def\Ct{\boldsymbol{\mathcal{C}}}
\def\PPhi{\boldsymbol{\Phi}}

\headers{Flow Map Learning of SDE}{Yuan Chen and Dongbin Xiu}

\title{Learning Stochastic Dynamical System via Flow Map Operator}

\author{Yuan Chen\footnotemark[1] \and Dongbin Xiu\footnotemark[1]\thanks{E-mail addresses: \texttt{\{chen.11050, xiu.16\}@osu.edu}. Department of Mathematics, The Ohio State University, Columbus, OH 43210, USA. Funding: This work was partially supported by AFOSR FA9550-22-1-0011.}}

\usepackage{amsopn}

\ifpdf
\hypersetup{
  pdftitle={Flow-Map Based Learning of Stochastic Dynamical System},
  pdfauthor={Yuan Chen \and Dongbin Xiu}
}
\fi

\begin{document}

\maketitle

\begin{abstract}
We present a numerical framework for learning unknown stochastic dynamical systems using measurement data.
Termed stochastic flow map learning (sFML), the new framework is an extension of flow map learning (FML) that was developed for learning deterministic dynamical systems.
For learning stochastic systems, we define a stochastic flow map that is
a superposition of two sub-flow maps: a deterministic sub-map and a stochastic sub-map. The stochastic training data are used
to construct the deterministic sub-map first, followed by the stochastic sub-map. The deterministic sub-map takes the form of residual network (ResNet), similar to the work of FML for deterministic
systems. For the stochastic sub-map, we employ a generative model, particularly generative adversarial networks (GANs) in this paper.
The final constructed stochastic flow map then defines a stochastic evolution model that is a weak approximation, in term of distribution, of the unknown stochastic system. A comprehensive set of numerical examples are presented to demonstrate the flexibility and effectiveness of the proposed sFML method for various types of
stochastic systems.
\end{abstract}



\section{Introduction}

In the recent years, data-driven modeling of unknown dynamical systems has become a prominent area of research among the scientific computing community.
Methods for modeling various problems have been developed. See, for example, \cite{brunton2016discovering,owhadi2021computational,raissi2019physics,li2020fourier,kang2019identifying,schaeffer2017sparse,schaeffer2018extracting,raissi2018multistep}, among many others.
Most of the work focused on deterministic problems, as modeling stochastic systems using noisy data poses additional challenges.

For modeling of stochastic systems, most of the existing efforts focus on learning It\^{o}-type stochastic differential equations (SDE).
These include Gaussian process approximation (\cite{yildiz2018learning,pmlr-v1-archambeau07a,darcy2022one,opper2019variational}), Gaussian mixture distribution model (\cite{infante2016approximations}), deep neural networks (DNNs) model (\cite{dridi2021learning,dietrich2021learning,gu2021stationary,yang2020physics,zhu2022learning}), polynomial approximations (\cite{wang2022data, li2021data}), sparsity promoting learning (\cite{boninsegna2018sparse}), reconstruction of the corresponding Fokker-Planck equations (\cite{friedrich2011approaching}),
to name a few. Methods based on physics-informed neural networks (PINNs) were also developed. For example,
the work of \cite{yang2022generative,chen2021solving} seeks to recover the evolution of the probability density function of the SDEs,
while the work of  \cite{chen2023data} leverages PINNs to recover the governing SDE with transition pathway data.

The focus, as well as the contribution, of this paper is on the development of a numerical framework for data driven modeling
of general stochastic dynamical systems.
More specifically, the method presented in this paper is applicable to the learning of a wide class of
stochastic systems beyond It\^{o}-type SDEs and Gaussian noises.
To accomplish this, we propose a method to learn a stochastic flow map from the data.
This is an extension of the deterministic flow map learning (FML), which was first
developed to learn unknown autonomous dynamical systems (\cite{qin2019data}), in conjunction with residual network (ResNet), and
was later extended to non-autonomous systems (\cite{qin2021data}), systems with missing variables (\cite{FuChangXiu_JMLMC20}), as well as
partial differential equations (\cite{wu2020data, chen2022deep}).
For stochastic systems, the mapping between the solutions at different time instances, i.e., the flow map, is perturbed by an external stochastic inputs, which are unknown and often modeled via Wiener process. Consequently, the flow map can not be approximated via any deterministic function.

The proposed stochastic flow map learning (sFML) method seeks to construct a stochastic flow map consisting of two components: a deterministic sub-flow map and a stochastic
sub-flow map. Using the noisy training data, we first construct the deterministic sub-map by fitting a deterministic function to the data. This
is accomplished via the use of ResNet, in the same manner as FML for learning deterministic dynamical systems (\cite{qin2019data}).
This ResNet fitting obviously results in non-negligible mismatch to the data, as a deterministic function can not fit noisy data perfectly (unless one
grossly over-parameterize the function). We then model the mismatch, which is the noisy component of the training data, using a stochastic sub-map
in the form of a generative model. In this paper, we explore the use of generative adversarial networks (GANs), which
are a class of generative DNN models designed for generating data with the same distribution as that of the training data, cf., \cite{goodfellow2020generative}. While successful for problems such as image generation (\cite{berthelot2017began,wang2018high,kazeminia2020gans}), text generation (\cite{zhang2017stackgan,reed2016generative}), etc., GANs also showed promises in scientific computing problems. See, for example, inverse problems (\cite{xu2021solving,liu2020rode}), SDEs (\cite{yang2022generative,chen2021solving,yang2020physics}), uncertainty quantification (\cite{yang2019adversarial}), to name a few.
In this paper, we employ Wasserstein GANs with gradient penalty (WGAN-GP) (\cite{pmlrv70arjovsky17a,gulrajani2017improved}) to construct the stochastic sub-map.
Upon learning both the deterministic sub-map and stochastic sub-map, we obtain the complete stochastic flow map.
Using this stochastic flow map an evolution operator, we define an iterative scheme in time as a predictive model of the unknown stochastic dynamical system. The resulting sFML model is a weak approximation, in distribution, of the unknown stochastic system, and allows us to analyze the long-term system behavior under different initial conditions (not in the training data set).
The effectiveness of the approach is then demonstrated via a comprehensive set of numerical examples.

This article is organized as following: In Section \ref{sec:setup}, we introduce the setup of the problem. In Section \ref{sec:prelim}, we give a brief review of the related
materials, including the deterministic FML and GANs. The detailed description of the proposed sFML method is presented in Section \ref{sec:method}, followed by the numerical examples in Section \ref{sec:numerical}. We then conclude the paper by a brief summary in Section \ref{sec:conclu}.
\section{Problem Statement}
\label{sec:setup}
We consider a general stochastic dynamical system,
\begin{equation}
\label{equ:gSDE}
    \frac{d \mathbf{x}_t}{dt}(\omega) = \mathbf{f}(\mathbf{x}_t,\omega), \qquad \x_{t_0}(\omega)\sim \P_0,
  \end{equation}
  where $\omega\in\Omega$, an event space in a properly defined
  probability space, represents the random inputs, $\P_0$ is the
  distribution for the initial condition $\x_{t_0}$,
and the solution
$\mathbf{x}_t:=\mathbf{x}(\omega,t):\Omega\times[0,T]\mapsto \mathbb{R}^d$, $d\geq 1$, is a $d$-dimensional stochastic process for some finite time $T>0$.
The right-hand-side (RHS) 
$\mathbf{f}:\mathbb{R}^d\times \Omega\mapsto \mathbb{R}^d$ is unknown. Consequently, the solution $\x_t$ can not be obtained by
solving \eqref{equ:gSDE}. 

\subsection{Assumptions}

We assume the dynamics of \eqref{equ:gSDE} is well-defined and time-homogeneous (cf. \cite{oksendal2003stochastic}). That is, for any $\Delta\geq 0$, the following property holds:
	\begin{equation} \label{autonomous}
		\mathbb{P}(\mathbf{x}_{s+\Delta}|\mathbf{x}_{s})=\mathbb{P}(\mathbf{x}_{\Delta}|\mathbf{x}_{0}),\qquad s\geq 0.
              \end{equation}

Let $\PPhi_{t,t_0}(\x_0)=\x(t+t_0)$ be the time independent flow of
\eqref{equ:gSDE}. More precisely,
\be \label{flowmap}
\PPhi:(\Rs^d\times\Rs)\times \Rs\to \Rs^d\times \Rs; \qquad
\PPhi((\x_0, t_0), t) = (\PPhi_{t,t_0}(\x_0), t+t_0).
\ee
The time homogeneous assumption \eqref{autonomous} implies that the
$\Delta$-shift flow map from $\mathbf{x}_{s}$ to
$\mathbf{x}_{s+\Delta}$ produces a conditional distribution that
depends only on the time lag $\Delta$ but not on the time variable $s$. 
Therefore, hereafter we use the following simplified notation for the
flow map:
For $\Delta\geq 0$, $s\geq 0$,
        \be \label{G}
        \G_\Delta:\Rs^d\times \Rs \to \Rs^d; \qquad \x _{s+\Delta}=\G_{\Delta}(\x_{s}).
        \ee
        Naturally, we have $\G_0=\I d$, the identity function, and
$$
\G_{\Delta}(\mathbf{x}_{s}) {\sim}
\mathbb{P}(\mathbf{x}_{s+\Delta}|\mathbf{x}_{s})
=\mathbb{P}(\mathbf{x}_{\Delta}|\mathbf{x}_{0}).
$$
Note that $\G_\Delta$ is equivalent to $\PPhi$ in \eqref{flowmap}. We
merely suppressed its explicit dependence on the time variable,
because its distribution is the conditional distribution that
depends only on the time lag $\Delta$. Hereafter we shall refer to
$\G_\Delta$ the stochastic flow map.

\begin{remark}
If the right-hand-side of \eqref{equ:gSDE} takes the following form
\begin{equation}
	\mathbf{f}(\mathbf{x}_t,\omega) = \mathbf{a}\left(\mathbf{x}_t\right)+\mathbf{b}\left(\mathbf{x}_t\right) \dot{\mathbf{W}}_t(\omega),
\end{equation}
where $\mathbf{a}$ is the drift function, $\mathbf{b}$ the diffusion
function, and $\mathbf{W}_{t}$ a multivariate Wiener process, we
obtain the classical form of SDE, which  is often written as
\begin{equation}
\label{Ito}
d \mathbf{x}_t = \mathbf{a}\left(\mathbf{x}_t\right) dt+\mathbf{b}\left(\mathbf{x}_t\right) d \mathbf{W}_{t}.
\end{equation}
It satisfies time-homogeneous assumption \eqref{autonomous}. In this case, our basic assumption of $\f$ being unknown mounts to the
drift $\mathbf{a}$ and the diffusion $\mathbf{b}$ being unknown. Note
that the
discussion in the paper does not require the unknown SDE
\eqref{equ:gSDE} to be  in the classical form \eqref{Ito}.
\end{remark}

\subsection{Setup and Objective}

Our goal is to construct a numerical model for the system \eqref{equ:gSDE}, whose governing equations are unknown, by using measurement data
of $\x_t$, such that the dynamics of \eqref{equ:gSDE} can be
accurately simulated by the constructed numerical model.

Suppose observation data of the system states $\x_t$ are available over a sequence of discrete time instances $0\leq t_0<t_1<...$.
For simplicity, we assume the time instances are uniformly distributed with a constant time lag, $\Delta = t_{n}-t_{n-1}, \forall n \geq 1$.
Suppose we have observation data of $N\geq 1$ number of solution trajectories. That is, for each $i=1,\dots, N$, we have the following
solution sequence
\be \label{traj}
\x\left(t_0^{(i)}\right),\x\left(t_1^{(i)}\right),\dots,\x\left(t_{L_i}^{(i)}\right), 
\ee
where $(L_i+1)$ is the length of the sequence. This is the $i$-th solution trajectory, in the sense that it contains the solution of
\eqref{equ:gSDE} from the initial condition $\x(t_0^{(i)})$. For notational convenience, hereafter we assume each of the
trajectories is of the same length, i.e., $L_i\equiv L, \forall i$.

Under the time-homogeneous assumption \eqref{autonomous}, the differences between the time instances are important, as opposed to the actual time values.
Therefore, we denote the available data as a set of $N\geq 1$ solution trajectories of length $(L+1)$,
\begin{equation}
    \label{data}
    \X^{(i)} = \left(\mathbf{x}_{0}^{(i)},\mathbf{x}_{1}^{(i)},...,\mathbf{x}_{L}^{(i)}\right),\qquad i=1,\dots, N,
  \end{equation}
  where $\x^{(i)}_k = \x(t^{(i)}_k)$, $1\leq i\leq N$, $0\leq k\leq L$, and the time variables
  $t^{(i)}_k$ are neglected. We remark that this is done for more than just 
  notational convenience. It implies that the actual time variables do
  not need to be recorded in the data set during data collection.  It
  makes our proposed method ``coordinate free'' and have a wider
  applicability in practice.
  In another word, each data sequence
  $\X^{(i)}$, $i=1,\dots,N$, can be considered as the solutions of the system \eqref{equ:gSDE} over  time instances
  $t_k = k\Delta$, $k=0,\dots, L$, with an initial condition $\x_0^{(i)}$.

  Given a training data set consisting sufficiently large number $N>1$ trajectory data \eqref{data}, our goal is to construct
  a numerical stochastic flow map $\Gt_\Delta$ as an approximation to the true (and unknown) stochastic flow map
  $\G_\Delta$ \eqref{G}. More specifically, we seek
  \be \label{Gweak}
  {\Gt_\Delta (\x)\stackrel{d}{\approx}  \G_\Delta (\x)},
  \ee
  where $ \stackrel{d}{\approx}$ stands for approximation in distribution. The numerical flow map $\Gt_\Delta$ thus defines a
  numerical model such that given an initial condition $\xt_0$,
  \be \label{model}
  \xt_{n+1} = \Gt_\Delta (\xt_n), \qquad n\geq 0,
  \ee
  where we use subscript $n$ to denote the solution at time $t_n=n\Delta$ hereafter. 
  The numerical model \eqref{model} becomes a weak approximation to
  the true unknown system \eqref{equ:gSDE}, in the sense that,
given an arbitrary initial condition $\xt_0 = \x_0 \sim \P_0$,
  \be
  \xt_n \stackrel{d}{\approx} \x_n, \qquad n=1,2,\dots.
  \ee
  Naturally, the condition \eqref{Gweak} can be not accomplished by any
  deterministic function. We therefore propose to develop a stochastic
  flow map learning method to construct $\Gt_\Delta$.

  \section{Preliminaries} \label{sec:prelim}
  
  The proposed stochastic flow map learning (sFML) method is an extension of the flow map learning (FML) method for deterministic systems.
  It involves the use of generative models. More specifically in this
  paper, we focus on the use of
  generated adversarial networks (GANs). Here, we briefly review the ideas of the deterministic FML method and GANs. 
  
\subsection{Deterministic Flow Map Learning}

Consider an autonomous system,
\be\label{eq:ODE}
\frac{d\x}{dt} = \f(\x), \qquad \x\in\Rs^d,
\ee
where $\f:\Rs^d\to \Rs^d$ is not known. Its flow map is a mapping
$\PPh:\Rs^d\times \Rs\to\Rs^d$ such that
$\PPh(\x(t),s) = \x(t+s)$. 
Over the uniform time stencils with the constant time lag $\Delta$ we consider in this paper, the flow map defines the evolution of the solution
\be \label{Phi}
\x_{n+1} = \PPh_{\Delta}(\x_n),
\ee
where we move the parameter $\Delta$ into the subscript. When the governing equation \eqref{eq:ODE} is unknown, this evolutionary
equation is also unknown.

The FML method developed in \cite{qin2019data} utilizes data to learn the evolutionary relation
\eqref{Phi} by constructing an approximate flow map $\widetilde{\PPh}\approx \PPh_\Delta$ such that
\be \label{Phi_t}
\xt_{n+1} = \widetilde{\PPh}_\Delta(\xt_n)
\ee
is an accurate approximation to \eqref{Phi}. This is accomplished by training the approximate evolution
\eqref{Phi_t} to match the training data set \eqref{data}, which are deterministic data in this case.

Consider the $i$-th trajectory data from \eqref{data}, $i=1,\dots,N$. Using $\x_0^{(i)}$ as the initial condition, i.e.,
$\widetilde{\x}_0^{(i)} = \x_0^{(i)}$,
the approximate flow map evolution equation \eqref{Phi_t}  generates a trajectory,
\be
    \widetilde{\X}^{(i)} = \left(\widetilde{\x}_{0}^{(i)},\widetilde{\x}_{1}^{(i)},...,\widetilde{\x}_{L}^{(i)}\right).
    \ee
    The approximate flow map $\widetilde{\PPh}_\Delta$ is then determined by minimizing the mean squared loss
\be 
\min \sum_{i=1}^N \left\|\widetilde{\X}^{(i)} - \X^{(i)}\right\|_F^2,
\ee
where the Frobenius norm becomes vector 2-norm for scalar systems with $d=1$.

The work of \cite{qin2019data} proposed the use of residual network (ResNet)
\be
\widetilde{\PPh}_\Delta =  \mathbf{I}+ \N,
\ee
where $\mathbf{I}$ is the identity operator and $\N:\Rs^d\to\Rs^d$ stands for the mapping operator of a standard
feedforward fully connected DNN. The minimization of the mean squared
loss thus becomes a minimization problem for the hyper parameters 
defining the DNN. Once the training is completed, one obtain a
predictive model for the underlying unknown dynamical system, 
such that, for a given initial condition $\x_0$,
\be \label{ResNet}
\xt_{n+1} =  \xt_n + \N(\xt_n), \qquad n=0,1,\dots,
\ee
where $\xt_0 = \x_0$. This framework has shown to be highly effective and accurate for many systems
(\cite{qin2019data, qin2021data, FuChangXiu_JMLMC20}).

\subsection{Generative Adversarial Networks}
\label{sec:gans}

GANs (cf.~\cite{goodfellow2014generative}) are widely used, among many
other tasks, to generate data with a desired target distribution.
%
Consider a target distribution $\mathbb{P}_r$, with observed data
$\mathbf{x}_{\texttt{data}}\sim \mathbb{P}_r$. Let $\mathbf{z}$ be
random variables with a fixed known distribution
$\mathbb{P}_{\mathbf{z}}$. For example, a normal distribution. 
In GANs, one defines a neural network, called generator,
$G(\mathbf{z},\theta_g)$ with its hyper parameter set $\theta_g$ to
map $\mathbf{z}$ 
into a new distribution $\mathbb{P}_{\theta_g}$. The goal is to tune
the parameter $\theta_g$ such that the output distribution matches the
target distribution $\mathbb{P}_r$ . 
At the same time, one defines another neural network, called 
  discriminator,  $D(\mathbf{x},\theta_d)$ with its hyper parameter
set $\theta_d$ to 
assign a score to the input $\x$ such that it is able to distinguish
whether $\x$ is from the target distribution $\mathbb{P}_r$ or the
``fake'' distribution 
$\mathbb{P}_{\theta_g}$.

The two networks $D$ and $G$ are trained by solving a a zero-sum
two-player game. The generator $G$ aims to ``fool'' the discriminator
$D$ and let it  believe the samples of $G(\z)$ are from the target
distribution $\mathbb{P}_r$.  At the same time, the discriminator $D$
aims to distinguish the samples from $G(\mathbf{z})$ and the real
$\mathbb{P}_r$. The optimization problem can be  written in the
following minimax form: 
\begin{equation}
\min _{G} \max _{D} L_{\texttt{GANs}}(D, G):=\mathbb{E}_{\mathbf{x}
  \sim \mathbb{P}_r}[\log D(\mathbf{x})]+\mathbb{E}_{\mathbf{z} \sim
  \mathbb{P}_z}[\log (1-D(G(\boldsymbol{z})))].
\end{equation}
%
In this paper, we adopt the popular  WGANs
(\cite{pmlrv70arjovsky17a}), which
{trains the discriminator to approximate}
Wasserstein-1 distance
{between the real and fake samples}.
The definition of Wasserstein-$p$ distance is:
\begin{equation} \label{Wp}
    W_p(\mu, \nu)=\left(\inf _{\gamma \in \Gamma(\mu, \nu)} \mathbb{E}_{(x, y) \sim \gamma} d(x, y)^p\right)^{1 / p}, \qquad p\geq 1,
\end{equation}
where $\Gamma(\mu, \nu)$ is the set of  all couplings of $\mu$ and $\nu$, and $d(\cdot,\cdot)$ is a metric.
More specifically, we employ WGANs with gradient penalty (WGANs-GP),
which introduces a penalty on the gradient of the discriminator to
enhance numerical stability (\cite{gulrajani2017improved}). The loss
function for WGANs-GP is: 
\begin{equation} \label{GP}
L_{\texttt{WGANs-GP}}(D, G) = \underset{\tilde{\boldsymbol{x}} \sim
  \mathbb{P}_g}{\mathbb{E}}[D(\tilde{\boldsymbol{x}})]-\underset{\boldsymbol{x}
  \sim \mathbb{P}_r}{\mathbb{E}}[D(\boldsymbol{x})]+\lambda
\underset{\hat{\boldsymbol{x}} \sim
  \mathbb{P}_{\hat{\boldsymbol{x}}}}{\mathbb{E}}\left[\left(\left\|\nabla_{\hat{\boldsymbol{x}}}
      D(\hat{\boldsymbol{x}})\right\|_2-1\right)^2\right], 
\end{equation}
where the distribution $\mathbb{P}_{\hat{\boldsymbol{x}}}$ is defined
to sample uniformly along straight lines between pairs of points
sampled from the data distribution $\mathbb{P}_r$ and $\mathbb{P}_g$,
and $\lambda \geq 0$ is a penalty constant.

\section{Stochastic Flow Map Learning}
\label{sec:method}

We now discuss the proposed stochastic flow map learning (sFML)
method. The method consists of two steps: (1) construction of a deterministic sub-flow map to
model the averaged dynamics of the unknown system; and (2)
construction of a stochastic sub-flow map to model the noisy dynamics of the unknown system.
The two sub-flow maps are learned from the same training data set \eqref{data}.

\subsection{Method Description}

\begin{figure}[htbp]
  \centering
  \label{fig:decompose}
  \includegraphics[width=.65\textwidth]{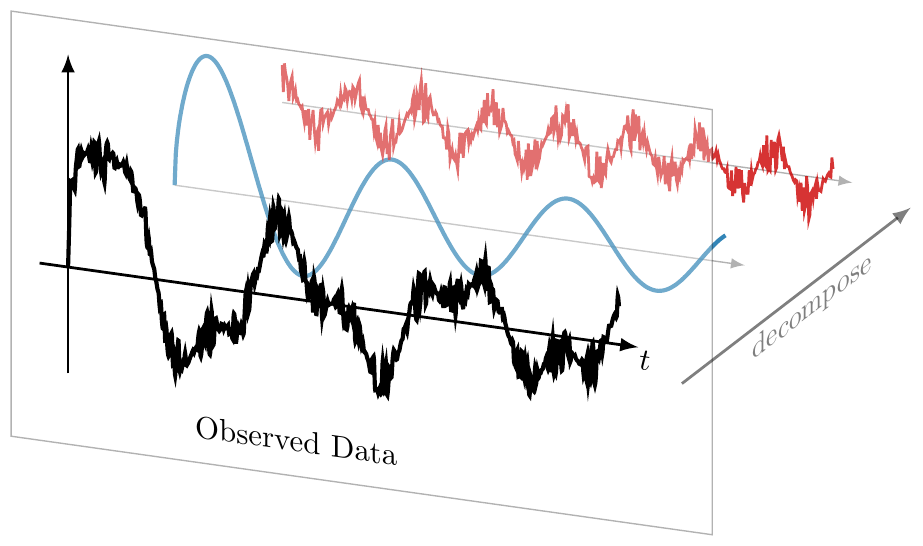}
  \caption{An illustration of the decomposition of stochastic dynamics
    into the superposition of a smooth dynamics, governed by the
    conditional expectation, and a noisy dynamics..
  }
\end{figure}

The solution of the unknown system \eqref{equ:gSDE} is governed by the
(unknown) stochastic flow map $\G_\Delta$ \eqref{G}. Its evolution
over two consecutive time steps can be decomposed into two parts:
\be
\x_{n+1} = \G_\Delta (\x_n)  = \mathbb{E}\left[\x_{n+1}|\x_n\right] + {\x}_{n+1}',
\ee
where the conditional mean $\mathbb{E}\left[\x_{n+1}|\x_n\right]$ is a
deterministic function of $\x_n$ and ${\x}_{n+1}'$ contains the
stochastic component. 
Therefore, we write the stochastic flow map as a composition of two parts,
\be 
\G_\Delta = \D_\Delta + \S_\Delta,
\ee
where
\be \label{DS}
\D_\Delta: \Rs^d\times \Rs\to \Rs^d = \mathbb{E}(\x_{n+1}|\x_n), \qquad
\S_\Delta: \Rs^d\times \Rs\to \Rs^d
\ee
are the deterministic part and stochastic part, respectively. Hereafter, we refer to $\D_\Delta$ as deterministic sub-map, and $\S_\Delta$ as 
stochastic sub-map. Obviously, neither $\D_\Delta$ nor $\S_\Delta$ is
known. Figure \ref{fig:decompose} illustrates the idea of the
decomposition.

Our proposed learning method seeks to construct numerical approximations to $\D_\Delta$ and $\S_\Delta$ by using the trajectory
data \eqref{data}. That is, we seek to construct a numerical stochastic flow map
\be \label{Gt}
\Gt_\Delta = \Dt_\Delta + \St_\Delta,
\ee
where $\Dt_\Delta$ and $\St_\Delta$ 
are the approximation to $\D_\Delta$ and $\S_\Delta$, respectively. Specifically, we seek $\Gt$ to be an approximation of $\G$ in distribution, in the sense that,
for any $\x\in\mathbb{R}^d$,
\be
\Gt_\Delta(\x) \stackrel{d}{\approx} \G_\Delta(\x).
\ee

\subsection{Learning Deterministic Sub-map}

Given the trajectory data \eqref{data}, we first construct a numerical deterministic sub-map
\be \label{Dt}
\Dt_\Delta: \mathbb{R}^d\times \Rs\to \mathbb{R}^d
\ee
as an accurate approximation of $\D_\Delta$.
This is accomplished by finding a deterministic operator such that it generates deterministic trajectories that
are the best fit to the stochastic training data \eqref{data}.

Let us consider the $i$-th trajectory data from \eqref{data}, $i=1,\dots,N$. Using $\x_0^{(i)}$ as the initial condition,
the deterministic sub-map generates a deterministic trajectory,
\be
    \overline{\X}^{(i)} = \left(\bar{\x}_{0}^{(i)},\bar{\x}_{1}^{(i)},...,\bar{\x}_{L}^{(i)}\right),
    \ee
    where $\bar{\x}_0^{(i)} = \x_0^{(i)}$, and 
\be
\bar{\x}_{n+1}^{(i)}  = \Dt_\Delta\left(\bar{\x}_n^{(i)}\right), \qquad n=0,\dots, L-1.
\ee
The deterministic sub-map in our method is defined as the one that minimizes the total mis-fit between the deterministic trajectories $\bar{\X}^{(i)}$ and the stochastic training
trajectories $\X^{(i)}$. More specifically, we use the common mean squared loss as the mis-fit measure and find
$\Dt_\Delta$ to accomplish
\be \label{msq}
\min \sum_{i=1}^N \left\|\overline{\X}^{(i)} - \X^{(i)}\right\|_F^2,
\ee
where the Frobenius norm becomes vector 2-norm for scalar systems with $d=1$.

To make the minimization problem tractable, we confine the search for $\Dt_\Delta$ in a finite dimensional space. This
is accomplished by employing a class of functions parameterized by a finite number of parameters. For low-dimensional
dynamical systems, one may employ a polynomial space. See, for example, \cite{WuXiu_JCPEQ18}.
For more general systems whose dimensions $d$ may not be small, DNNs have become a more popular choice.
In this paper, we also utilize DNN as the deterministic sub-map. Similarly to the work of deterministic FML (\cite{qin2019data}), we employ
ResNet structure and define
\be
\label{de_model}
\Dt_\Delta(\cdot; \Theta) := \mathbf{I} + \mathbf{N}_\Delta(\cdot; \Theta),
\ee
where $\mathbf{I}$ is the identity matrix of size $(d\times d)$ and $\mathbf{N}_\Delta(\cdot; \Theta):\mathbb{R}^d\to\mathbb{R}^d$ is the mapping operator
of a connected feedforward DNN with its hyperparameter set $\Theta$. Note that $\Theta$ is a finite set, whose cardinality is determined by the
width and depth of the DNN. The minimization problem \eqref{msq} thus becomes a minimization problem for the DNN parameter $\Theta$, i.e.,
\be \label{D_loss}
\min_\Theta \sum_{i=1}^N \sum_{n=1}^{L} \left\| \x_n^{(i)} - \Dt_\Delta^{[n]}\left(\x_0^{(i)}; \Theta\right)\right\|_2^2,
\ee
where $\Dt_\Delta^{[n]}$ stands for $n$ times composition of the operator $\Dt_\Delta$. Upon solving the minimization problem,
the hyperparameters $\Theta$ in the DNN are fixed. We thus neglect $\Theta$ for notational convenience, unless confusion arises otherwise.

Note that in the loss function \eqref{D_loss}, the case of $L>1$
corresponds to the well-known multi-step loss, which is known to
improve forecast accuracy for time series analysis,
cf. \cite{WeissA84, ChevillonH05, FransesL09}. It has also been used
in FML for improving numerical stability of long-term predictions
(\cite{FuChangXiu_JMLMC20,  chen2022deep}). Setting $L=1$, it is
straightforward to see that the loss function \eqref{D_loss} ensures
that, for sufficiently large number of samples $N\gg 1$, we have
\be
\Dt_\Delta(\x_0) \approx \mathbb{E}(\x_1|\x_0),
\ee
which implies, upon using the time homogeneous assumption \eqref{autonomous},
\be
\Dt_\Delta(\x_n) \approx \D_\Delta(\x_n) = \mathbb{E}(\x_{n+1}|\x_n).
\ee

\subsection{Learning Stochastic Sub-map}

Once the deterministic sub-map $\Dt_{\Delta}$ is constructed, we
proceed to construct the stochastic sub-map $\St_\Delta$ in the
numerical stochastic flow map \eqref{Gt}, as an approximation to the
exact stochastic sub-map $\S_\Delta:\Rs^d\times \Rs\to\Rs^d$ in \eqref{DS}.
Note that as the stochastic component of the flow map
$\G_\Delta$, $S_\Delta$ is subject to random input that is a function
of the time lag $\Delta$ and unknown. In order to approximate this
unknown random input, we define a generative model that incorporates
an explicit random input. More precisely, we define
\be \label{St}
\St_\Delta:\mathbb{R}^d\times \Rs\times \mathbb{R}^{n_s}\to \mathbb{R}^d, \qquad n_s\geq 1,
\ee
where $n_s$ defines the stochastic dimension of the stochastic sub-map.
Note that the true stochastic dimension of the sub-map $\S_\Delta$ is
unknown.
Upon choosing a finite number $n_s\geq 1$, we seek to construct the numerical stochastic flow map
\be \label{GGt}
\Gt_\Delta=\Dt_\Delta+\St_\Delta: \mathbb{R}^d\times \Rs\times \mathbb{R}^{n_s}\to \mathbb{R}^d
\ee
as a weak approximation (in distribution) to the true (unknown) stochastic flow map $\G_\Delta$
\eqref{G}. 
Note that at this point the deterministic sub-map $\Dt_\Delta$ is
fixed, as it is already constructed by the procedure in the previous
section. The only task is to construct the stochastic sub-map $\St_\Delta$.

To construct the stochastic sub-map $\St_\Delta$, we again utilize the training data set \eqref{data}. For each of the $i$-th trajectory data in the set,
$i=1,\dots,N$, we use $\x_0^{(i)}$ as the initial condition and generate
\be \label{Xhat}
    \widehat{\X}^{(i)} = \left(\hat{\x}_{0}^{(i)},\hat{\x}_{1}^{(i)},...,\hat{\x}_{L}^{(i)}\right),
    \ee
    where $\hat{\x}_0^{(i)} = \x_0^{(i)}$, and 
\be \label{x_generate}
\hat{\x}_{n+1}^{(i)}  = \Gt_\Delta\left(\hat{\x}_n^{(i)}, \z^{(i)}_n\right), \qquad n=0,\dots, L-1,
\ee
where $\z_n^{(i)}\in\mathbb{R}^{n_s}$ are i.i.d. random variables following a prescribed known distribution. For example, unit Gaussian.

We then minimize the distance between $\left\{\widehat{\X}^{(i)}\right\}_{i=1}^N$ and $\left\{\X^{(i)}\right\}_{i=1}^N$, where the distance function
is chosen to measure probability distribution. Since $\Dt_\Delta$ in $\Gt_\Delta$ is fixed, the minimization procedure thus determines the
stochastic sub-map $\St_\Delta$, which in turn becomes a minimization problem for the hyperparameters after we express $\St_\Delta$ in a parameterized
family of functions.

In this paper, we adopt GANs for the
construction of the stochastic sub-map $\St_\Delta$. Specifically, we
use DNN to represent $\St_\Delta$, i.e.,
\be
\St_\Delta := \St_\Delta(\x, \z; \Theta),
\ee
where
$
\x\in\mathbb{R}^d$, $\z\in\mathbb{R}^{n_s},
$
and $\Theta$ represents the hyperparameters in the DNN. Here $\z$ is a
random vector within known distribution. In our examples, we use
i.i.d. unit normal distribution. The dimension of $\z$, $n_s\geq 1$,
is a choice made by user, in order to approximate the true probability
space $\Omega$ whose dimension is unknown.
Consequently, the stochastic flow map $\Gt_\Delta= \Gt_\Delta(\x, \z;
\Theta)$ by using \eqref{GGt}.

By using the GANs approach described in Section \ref{sec:gans}, the generator
is the stochastic flow map $\Gt_\Delta(\x,\z;\Theta)$, which generates the
``fake'' data sequence $\left\{\widehat{\X}^{(i)}\right\}_{i=1}^N$ in
\eqref{Xhat}. The generator is {expected to minimize the distance} between the ``fake'' data
$\widehat{\X}:=\left\{\widehat{\X}^{(i)}\right\}_{i=1}^N$ \eqref{Xhat} and the training data
$\X:=\left\{\X^{(i)}\right\}_{i=1}^N$ \eqref{data}. {To achieve this goal}, a
discriminator network $\Ct$, which is another DNN with its own
hyperparameters, is trained to distinguish the two sets of data {by approximating the
Wasserstein-1 distance \eqref{Wp}}. For
the network training, we 
adopt the WGANs with gradient penalty (WGANs-GP), whose loss function
\eqref{GP} takes the following form in our case:
\begin{equation} \label{Dis}
L(\Ct, \Gt_\Delta) = \underset{\tilde{\x} \sim
  \widehat{\X}}{\mathbb{E}}[\Ct (\tilde{\x})]-\underset{\x
  \sim \X}{\mathbb{E}}[\Ct (\x)]+\lambda
\underset{\hat{\x} \sim
  \mathbb{P}(\widehat{\X}, \X)}{\mathbb{E}}\left[\left(\left\|\nabla_{\hat{\x}}
      \Ct(\hat{\x})\right\|_2-1\right)^2\right], 
\end{equation}
where $\mathbb{P}(\widehat{\X},\X)$ is 
uniform sampling distribution along the straight lines between pairs of points
from $\widehat\X$ and $\X$,
and $\lambda \geq 0$ is a penalty constant.

\subsection{DNN Structures and Algorithm}

In this section, we summarize the sFML algorithm and its corresponding
DNN structures.

Given the training data set \eqref{data}, the sFML algorithm consists
of the following two steps:
\begin{itemize}
  \item[(1)] Constructing the deterministic sub-map {$\Dt_\Delta$} by
    minimizing the loss function \eqref{D_loss}. The multi-step loss
    (when $L>1$) requires recurrent use of the deterministic sub-map
   {$\Dt_\Delta$}, as illustrated in Fig.~\ref{fig:RNNNet}.
    \begin{figure}[htbp]
  \centering
  \label{fig:RNNNet}
  \includegraphics[width=.6\textwidth]{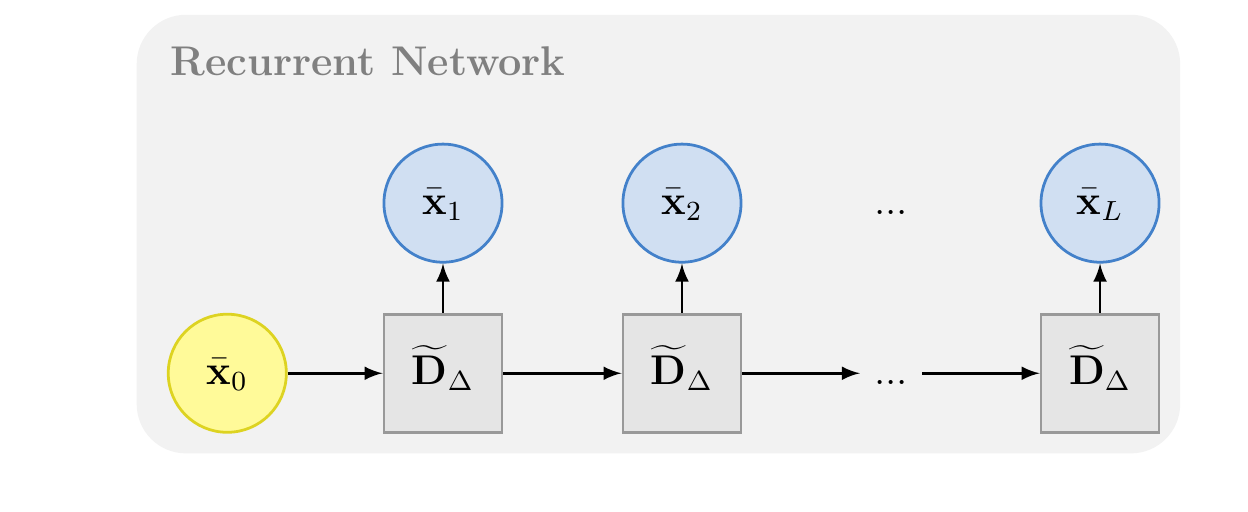}
  \caption{An illustration of the recurrent network structure for
    the deterministic sub-map $\Dt_{\Delta}$.
  }
\end{figure}

\item[(2)] Constructing the stochastic sub-map $\St_\Delta$, which
  defines the generator $\Gt_\Delta = \Dt_\Delta +
  \St_\Delta$. Subsequently, the generator generates the ``fake'' data
  $\widehat{\X}$ in \eqref{Xhat} via \eqref{x_generate}. In the spirit
  of ResNet, we write \eqref{x_generate} in an equivalent form
  \be \label{xx_generate}
\hat{\x}_{n+1}^{(i)}  = \hat{\x}_n^{(i)} +
\Gt_\Delta\left(\hat{\x}_n^{(i)}, \z^{(i)}_n\right), \qquad n=0,\dots,
L-1, \quad i=1,\dots,N,
\ee
where we use the same notation $\Gt_\Delta$ here to refer to the core DNN
part of the operation. Therefore, $\Gt_\Delta$ corresponds to the
difference
$\hat{\y}_n = \hat{\x}_{n} - \hat{\x}_{n-1}$, $n=1,\dots,L$.

The generator is illustrated on the left of Fig.~\ref{fig:SDENet}. 
The right of Fig.~\ref{fig:SDENet} illustrates the discriminator
$\Ct$, which is a standard DNN that computes a score $\mathbf{s}$ to
the ``fake'' data. It also computes a score for the training data
$\X$.  These are then used in the minimization of the loss function \eqref{Dis}.
\begin{figure}[htbp]
  \centering
  \label{fig:SDENet}
  \includegraphics[width=\textwidth]{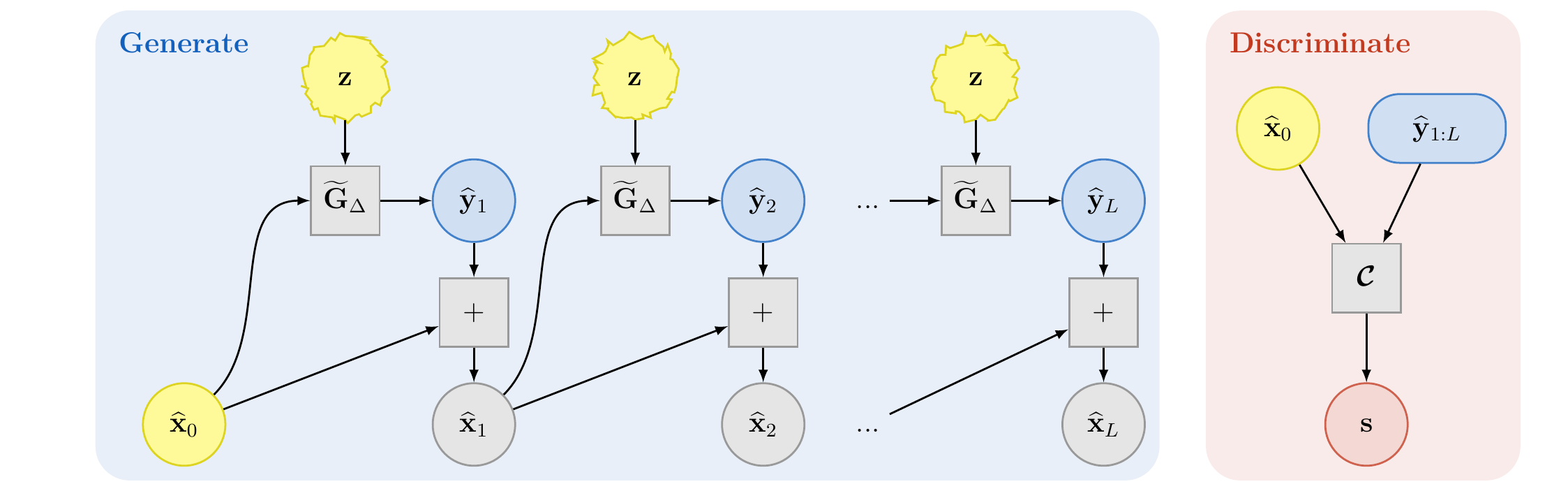}
  \caption{Left: An illustration of the generator in the stochastic
    flow map. Right: Discriminator that computes a score for the data.
  }
\end{figure}

\end{itemize}

The algorithm of the sFML method is summarized in Algorithm \ref{alg:wgangp}.
\begin{algorithm}
\caption{Learning SDE via GANs}
\label{alg:wgangp}
\begin{algorithmic}[1]
\REQUIRE Trajectory data of SDE: $\{(\mathbf{x}_{0},\mathbf{y}_{1},...,\mathbf{y}_{L})^{(i)}\}_{i=1}^N$; Deterministic model $\Dt_\Delta$.
\ENSURE  Recurrent length $L$; Noise input dimension $Z$ of generator; Layers and neurons each layer, Adam optimizer hyper-parameters $\beta_1$, $\beta_2$, base learning rate $l_r$ of DNN $\St_\Delta$, $\boldsymbol{\mathcal{C}}$ with training parameter $\Theta_{\boldsymbol{\S}}$ and $\Theta_{\boldsymbol{\mathcal{C}}}$; Iterations of training critic per generator iteration $n_{\texttt{ct}}$; Batch size $B$ and number of batches $n_B=N/B$ per epoch; Number of training epochs $n_{E}$; Gradient penalty constant $\lambda$.

\hspace*{-1.6\algorithmicindent}\algorithmicnote In the following algorithm, we apply operator $\Dt$, $\St$ and $\boldsymbol{\mathcal{C}}$ by rows when imputing a matrix.
\vspace{1mm}

\FOR{$n=1,2,...,n_E$}
    \FOR{$i=1,2,...,n_{B}$}
        \STATE{Slice $i$th group of data $\{(\mathbf{x}_{0},\mathbf{y}_{1},...,\mathbf{y}_{L})^{(i)}\}_{i=1+(i-1)B}^{iB}$;~~ $\mathbf{\widehat{x}}_{0} \gets \mathbf{x}_{0}$;}
        \FOR{$j=0,1,...,L-1$}
            \STATE{Sample $B$ $Z$-dimensional random vectors $\mathbf{z}=\{\mathbf{z}_k\}_{k=1}^{B}$, $\mathbf{z}_k \sim \mathcal{N}(\mathbf{0},\mathbf{I})$;}
            \STATE{$\mathbf{\widehat{y}}_{j+1} \gets \Dt_{\Delta}(\mathbf{\widehat{x}}_{j})-\mathbf{\widehat{x}}_{j}+\St_{\Delta}(\mathbf{\widehat{x}}_{j},\mathbf{z})$;}
            \STATE{$\mathbf{\widehat{x}}_{j+1} \gets \mathbf{\widehat{x}}_{j}+\mathbf{\widehat{y}}_{j+1}$;}
            \hfill\COMMENT{Generate fake SDE data}
        \ENDFOR
        \FOR{$k=1,2,...,B$}
            \STATE{Sample $n_B$ random numbers $\boldsymbol{\epsilon}=\{\epsilon_k\}_{k=1}^{n_B}$, $\epsilon_k \sim \mathcal{U}(0,1)$;}
            \STATE{$\mathbf{\widetilde{y}}_{1:L}^{(k)} \gets \epsilon_k \mathbf{y}_{1:L}^{(k)}+(1-\epsilon_k) \mathbf{\widehat{y}}_{1:L}^{(k)}$;}
            \STATE{$P_{\boldsymbol{\mathcal{C}}}^{(k)} \gets \left(\left\|\nabla_{(\mathbf{x}_0^{(k)},\mathbf{\widetilde{y}}_{1:L}^{(k)})} \boldsymbol{\mathcal{C}}(\mathbf{x}_0^{(k)},\mathbf{\widetilde{y}}_{1:L}^{(k)})\right\|_2-1\right)^2$;}
            \STATE{$L_{\boldsymbol{\mathcal{C}}}^{(k)} \gets \boldsymbol{\mathcal{C}}(\mathbf{x}_0^{(k)},\mathbf{\widehat{y}}_{1:L}^{(k)})-\boldsymbol{\mathcal{C}}(\mathbf{x}_0^{(k)},\mathbf{y}_{1:L}^{(k)})+\lambda P_{\boldsymbol{\mathcal{C}}}^{(k)}$;}
        \ENDFOR
        \STATE{$\Theta_{\boldsymbol{\mathcal{C}}} \gets \texttt{Adam}(\frac{1}{B} \sum_{k=1}^B L_{\boldsymbol{\mathcal{C}}}^{(k)},l_r,\beta_1,\beta_2)$;}
        \hfill\COMMENT{Update Critic}
        \IF{\texttt{mod}($n_B,{n_{\texttt{ct}}}) = 0$}
            \FOR{$k=1,2,...,B$}
                 \STATE{$L_{\S}^{(k)} \gets -\boldsymbol{\mathcal{C}}(\mathbf{x}_0^{(k)},\mathbf{\widehat{y}}_{1:L}^{(k)})$;}
            \ENDFOR
            \STATE{$\Theta_{\S} \gets \texttt{Adam}(\frac{1}{B} \sum_{k=1}^B L_{\S}^{(k)},l_r,\beta_1,\beta_2)$;}
            \hfill\COMMENT{Update Generator}
        \ENDIF
    \ENDFOR
\ENDFOR

\end{algorithmic}
\end{algorithm}

\subsection{Model Prediction and Analysis}

Upon satisfactory training of the DNNs,
we obtain a predictive sFML
model for the underlying unknown SDE. For a given initial condition
$\x_0$,
  \be \label{sFML}
\xt_{n+1}  = 
\Gt_\Delta\left(\xt_n, \z_n\right) =
\Dt_\Delta(\xt_n) + \St_\Delta(\xt_n, \z_n), \qquad n=0,1,2\dots,
\ee
with $\xt_0 = \x_0$. The model can be marched forward in time far
beyond the length of the training data.

Note that the sFML model resembles the form of the classical SDE
\eqref{Ito}, regardless whether the underlying unknown SDE is
 in the form of \eqref{Ito} or not. In fact, we can be re-write the
 sFML model as
   \be 
   {\xt_{n+1}  = \at (\xt_n)  \Delta  + \bt (\xt_n) \delta
     \mathbf{W}_n,}
   \ee
   where
   \be \label{abt}
   \at (\x) =
   \frac{\mathbb{E}_{\z}(\Gt_\Delta(\x,\z)-\x)}{\Delta},\qquad
   \bt (\x)
   = \frac{\text{Std}_{\z}(\Gt_\Delta(\x,\z))}{\sqrt{\Delta}},
   \ee
   and can be considered the effective drift and diffusion, respectively.
  Therefore, if the underlying unknown true SDE is of the classical
  form \eqref{Ito}, we
   expect the effective drift and diffusion \eqref{abt} of the sFML model \eqref{sFML} to be good approximations
   of the true drift and diffusion, respectively.

\section{Numerical Results}
\label{sec:numerical}

In this section, we present several numerical examples to demonstrate the performance of the proposed sFML method. The examples cover the following
cases:
\begin{itemize}
\item Linear SDEs. These include an Ornstein-Uhlenbeck (OU) process and a geometric Brownian motion;
 \item Nonlinear SDEs. These include SDEs with exponential and trigonometric
   drift or diffusions, as well as the well-known stochastic double-well potential problem;
 \item SDEs with non-Gaussian noise, including exponential distribution and lognormal distribution.
   \item Two-dimensional SDEs, which include a 2-dimensional OU process
     and a stochastically driven harmonic oscillator.
   \end{itemize}
   For all of these problems, the exact SDEs are known. They are used to generate the training data sets, which 
   are generated by solving the SDEs with Euler-Maruyama method, with initial conditions uniformly distributed in a region (specified later for each example).
   The SDEs are solved using a time step $\Delta  = 0.01$ for 100 steps, from which we randomly choose a length $L=40$ sequence to form our training data set \eqref{data}.
   In other words, the training data are solution sequences of length $T=0.4$.
   Our training data set \eqref{data} consists of $N=10,000$ such trajectories.

   For the DNN structure, we use a simple fully connect feedforward DNN for both the  deterministic sub-map $\Dt_\Delta$ \eqref{de_model} (see Figure \ref{fig:RNNNet}) and the discriminator $\Ct$ in Figure \ref{fig:SDENet}
  in the generator. In most examples,  we use 3 layers, where each
  layer has 20 nodes. The exceptions are the 2-dimensional
   examples, where each layer has 40 nodes.
   In the training process of the GANs, we follow Algorithm \ref{alg:wgangp}  and set $n_{\texttt{ct}}=5$, $\beta_1=0.5$, $\beta_2=0.999$, $l_r = 5\times 10^{-5}$.
   All examples are trained up to $100,000$ epochs.


   To examine the performance of the constructed sFML models, we provide the following ways:
   \begin{itemize}
   \item Simulated trajectories and their mean and standard deviation, for visual comparisons as well as the basic statistics;
   \item Conditional distribution the stochastic flow map
     $\Gt_\Delta$, in comparison to that of the true flow map $\G_\Delta$;
   \item Evolution of the predicted solution distribution at certain
     times beyond the training data time domain.
   \item Comparison of the effective drift and diffusion \eqref{abt}
     against the true drift and diffusion to those of 
     true SDEs.
   \end{itemize}

   \subsection{Linear SDEs}

   For scalar linear SDEs, we present the results for learning an OU
   process and a geometric Brownian motion.

\subsubsection{Ornstein–Uhlenbeck process}
\begin{figure}[htbp]
  \centering
  \label{fig:OU_data}
  \includegraphics[width=.43\textwidth]{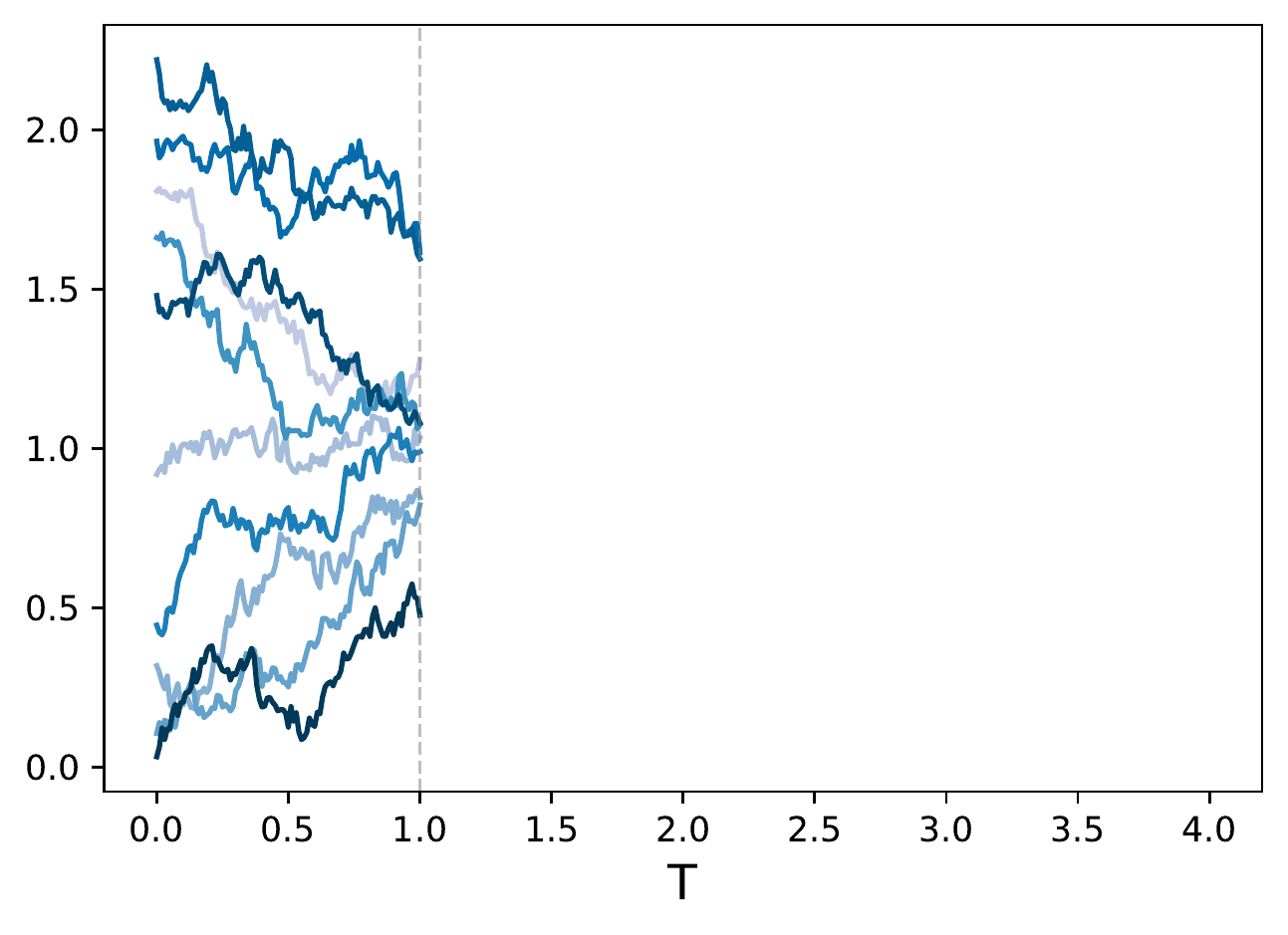}
  \includegraphics[width=.43\textwidth]{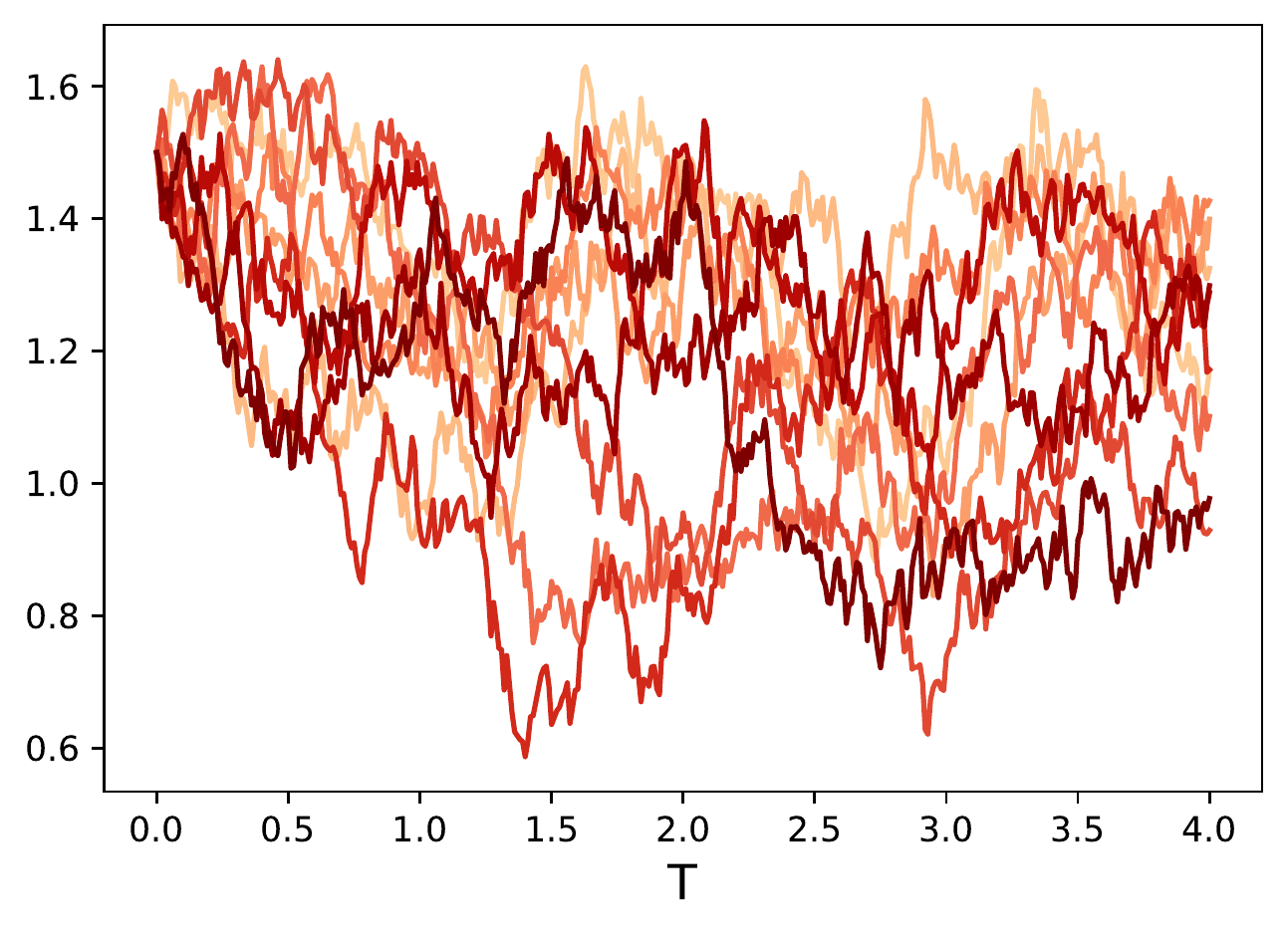}
  \caption{Solutions of the OU process. Left: samples of the training data; Right: simulated samples by the sFML model for up to $T=4.0$, with an initial condition $x_0=1.5$.}
\end{figure}

\begin{figure}[htbp]
  \centering
  \label{fig:OU_stat}
  \includegraphics[width=.9\textwidth]{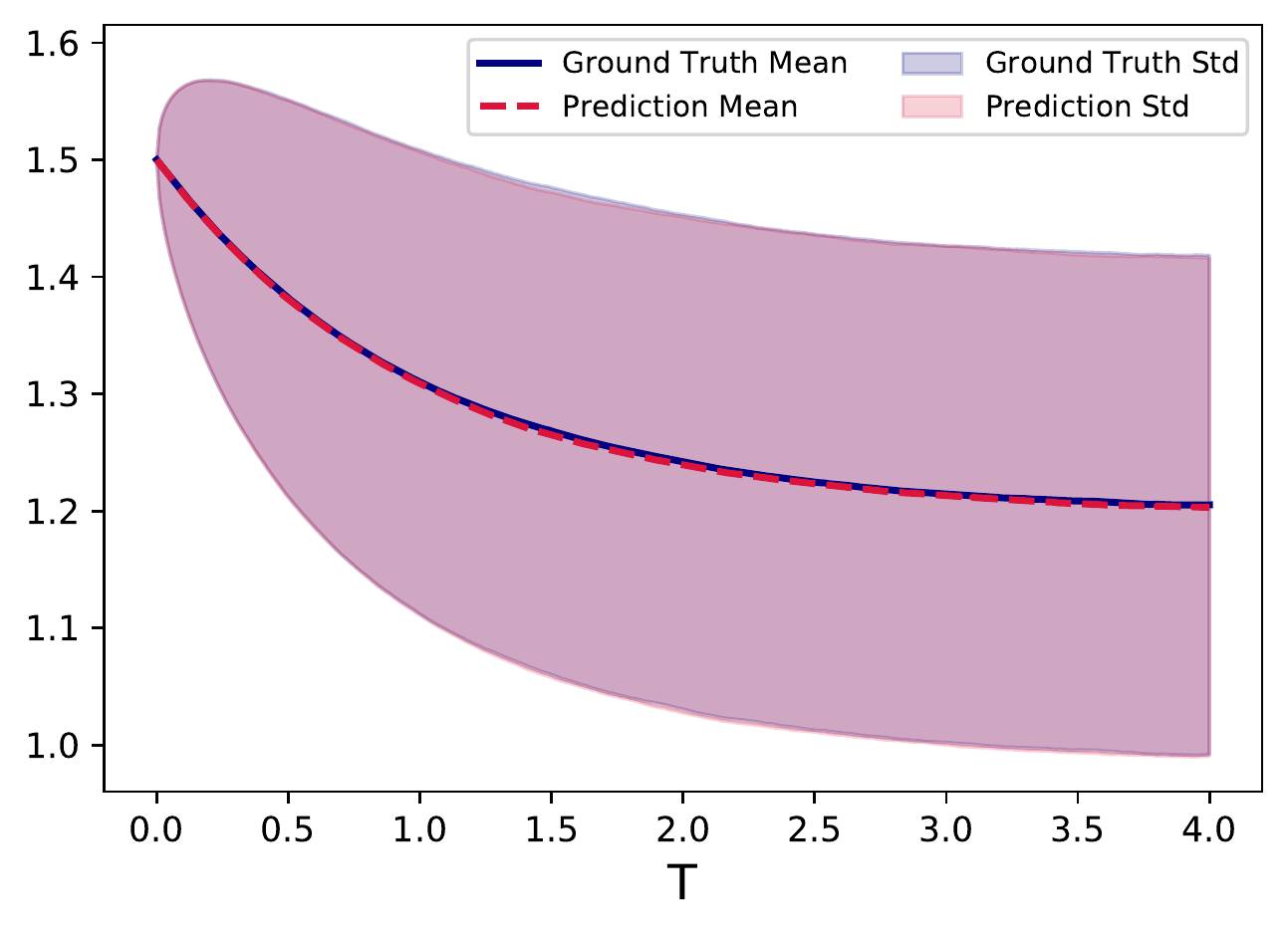}
  \caption{Mean and standard deviation (STD) of the OU process by the sFML model.}
\end{figure}

We first consider the following Ornstein–Uhlenbeck (OU) process,
\begin{equation}
    dx_t = \theta(\mu-x_t)dt+\sigma dW_t,
\end{equation}
where $\theta=1.0$, $\mu=1.2$, and $\sigma=0.3$. The training data are
generated by solving the SDE with initial conditions uniformly sampled from $\mathcal{U}(0,0.25)$.
Some of these samples are shown on the left of Figure \ref{fig:OU_data}. Upon training the DNNs and creating the sFML model for the OU process, we generat simulated
trajectories from the learned model for up to $T=4.0$. On the right of
Figure \ref{fig:OU_data}, we show the sFML model simulation results with an initial condition $x_0=1.5$ for time up to $T=4.0$.
We also compute the mean and standard deviation of the sFML model
prediction, averaged over $100,000$ simulation samples. The results
are shown in Figure \ref{fig:OU_stat}, along with the reference mean and variance from the true OU process. We observe
good agreement between the learned sFML model and the true model. Note that the agreement goes beyond the time horizon of the training data by a factor of 4.
%
%

\begin{figure}[htbp]
  \centering
  \label{fig:OU_ab}
  \includegraphics[width=.43\textwidth]{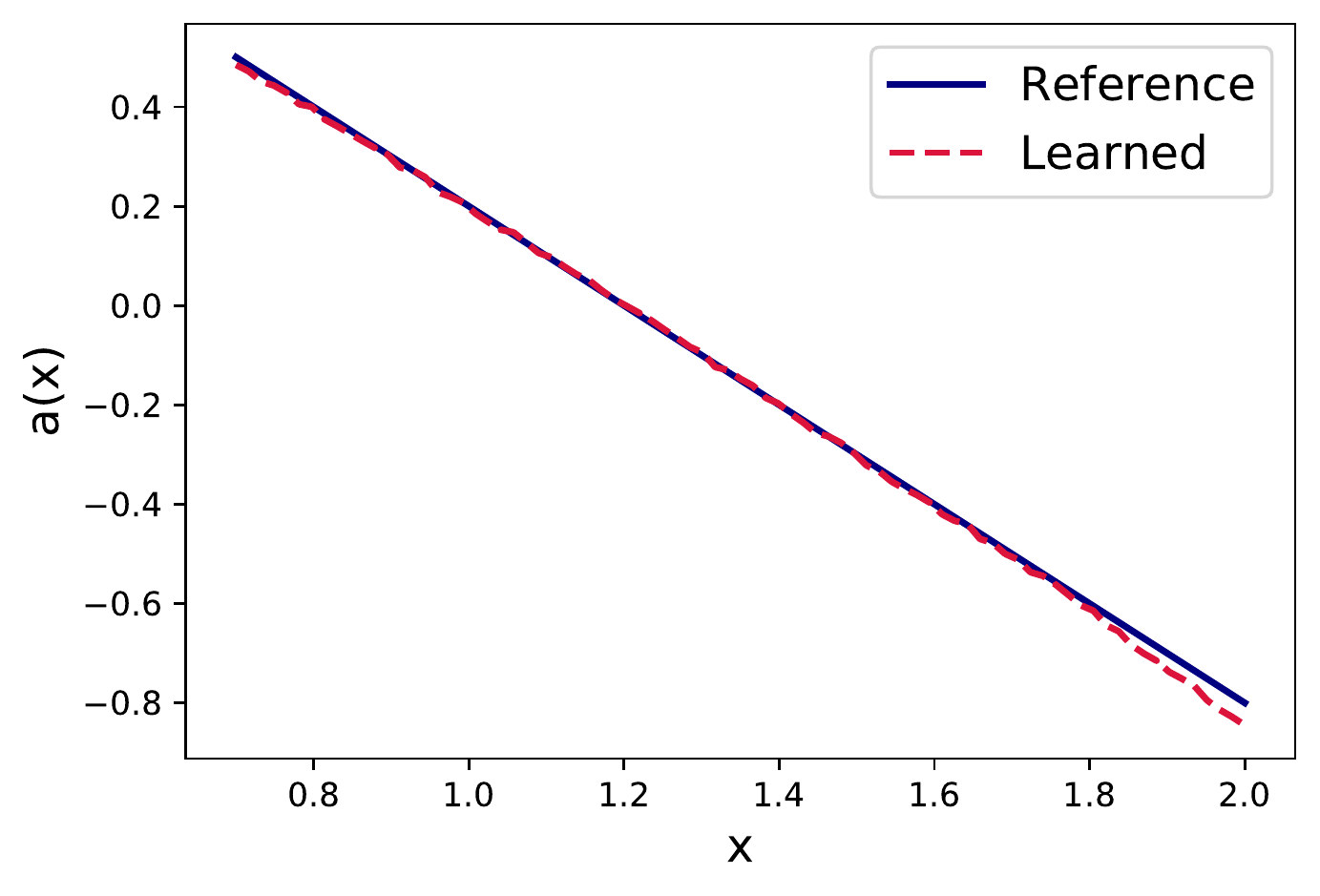}
  \includegraphics[width=.43\textwidth]{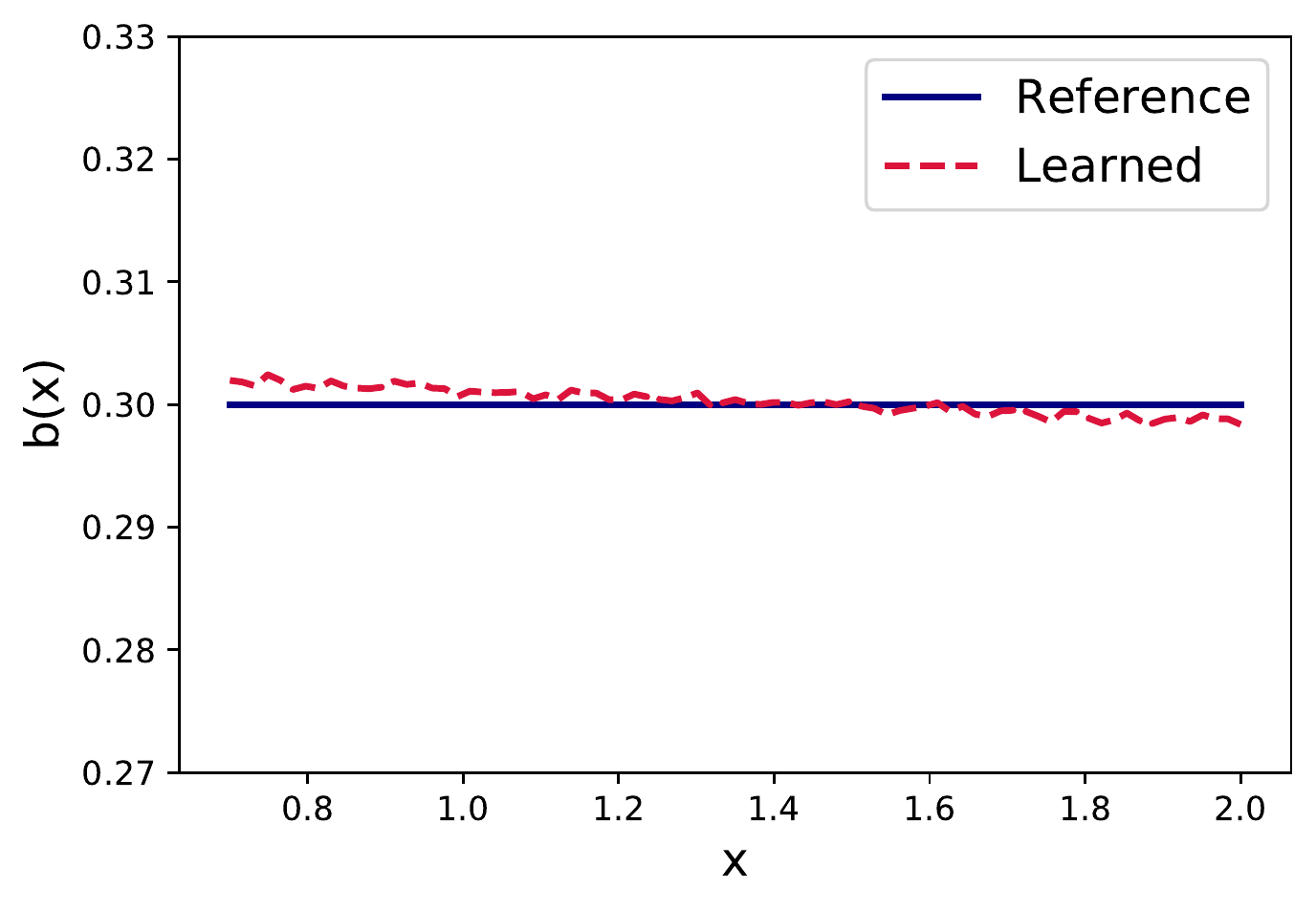}
  \caption{Effective drift and diffusion of the OU process. Left: drift $a(x)$; Right: diffusion $b(x)$.}
\end{figure}

\begin{figure}[htbp]
  \centering
  \label{fig:OU_pdf}
  \includegraphics[width=.9\textwidth]{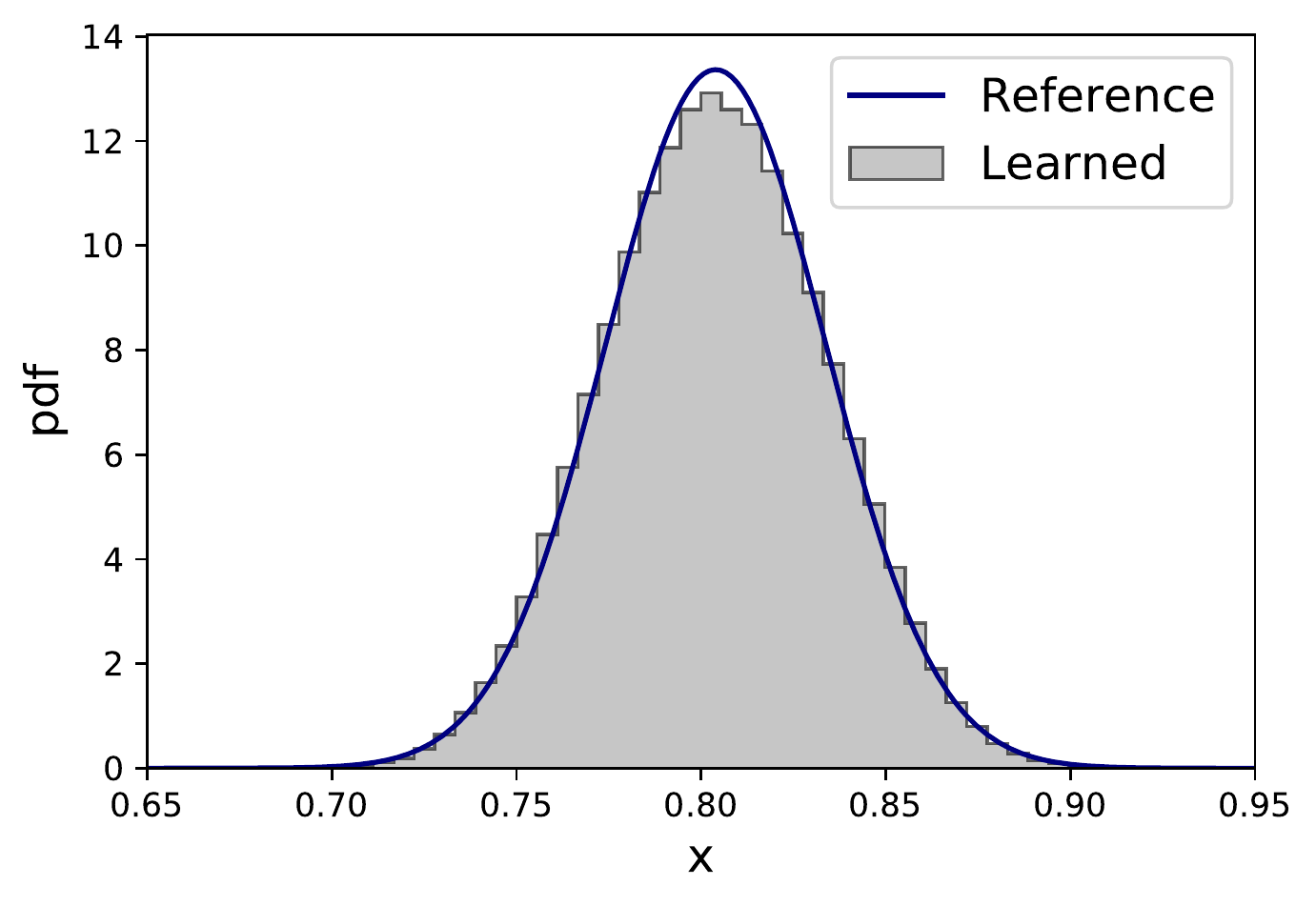}
  \caption{OU process at steady state: comparison of the conditional
    distribution of the sFML model $\Gt_\Delta(0.8)$ against that of
    the true model $\G_\Delta(0.8)$.}
\end{figure}

In Figure \ref{fig:OU_ab}, we show the recovered effective drift and
diffusion functions for the OU process, by following the procedure
\eqref{abt} using $100,000$ sFML model
simulations. Comparing them to the true drift and diffusion,
we observe good accuracy of the recovered functions, with relative errors on the order of $10^{-2}$.
Note that the discrepancy is relatively larger towards the end points
of the comparison domain. This is expected
because there are far less number of samples in these regions to gather highly accurate statistical results. Such a behavior is presented throughout all the examples in this paper.
The OU process eventually reaches a steady state.
In Figure \ref{fig:OU_pdf}, we plot the histogram for the samples of $\Gt_\Delta(0.8)$, as a comparison of the conditional distribution against that of the
true stochastic flow map $\G_\Delta(0.8)$. Good agreement can be observed.

\begin{figure}[htbp]
  \centering
  \label{fig:OU_train}
  \includegraphics[width=.6\textwidth]{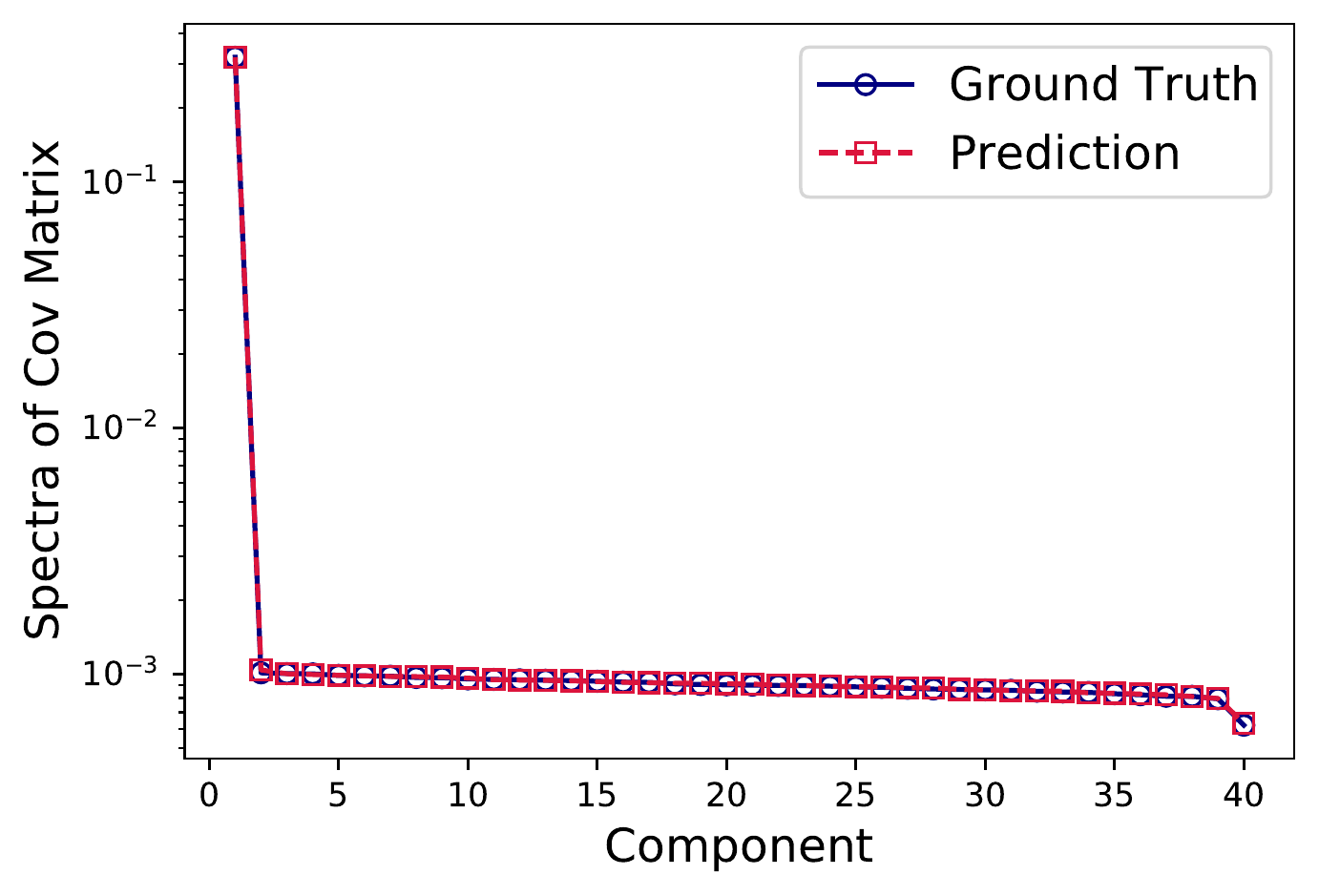}
  \caption{OU process: comparison of covariance matrix spectra of the
    sFML model solution sequence (``Prediction'') against that of the
    true solution (``Ground Truth'').}
\end{figure}

We conduct further examination of the solution time sequence produced
by the sFML model. In Figure \ref{fig:OU_train}, we present the
spectra of the covariance function of the simulated sequences by the
sFML model, along with that of the true solution sequence under the same
initial condition. We observe good agreement between the sFML model and the true model. 
%

\subsubsection{Geometric Brownian motion}
We now consider geometric Brownian motion, which, unlike the OU
process, does not have steady state. In particular, we consider 
\begin{equation}
    d x_t = \mu x_t d t+\sigma x_t d W_{t},
\end{equation}
where $\mu$ and $\sigma$ are constant parameters. We set $\mu=2.0$, $\sigma=1.0$ in our test.

\begin{figure}[htbp]
  \centering
  \label{fig:GBM_data}
  \includegraphics[width=.43\textwidth]{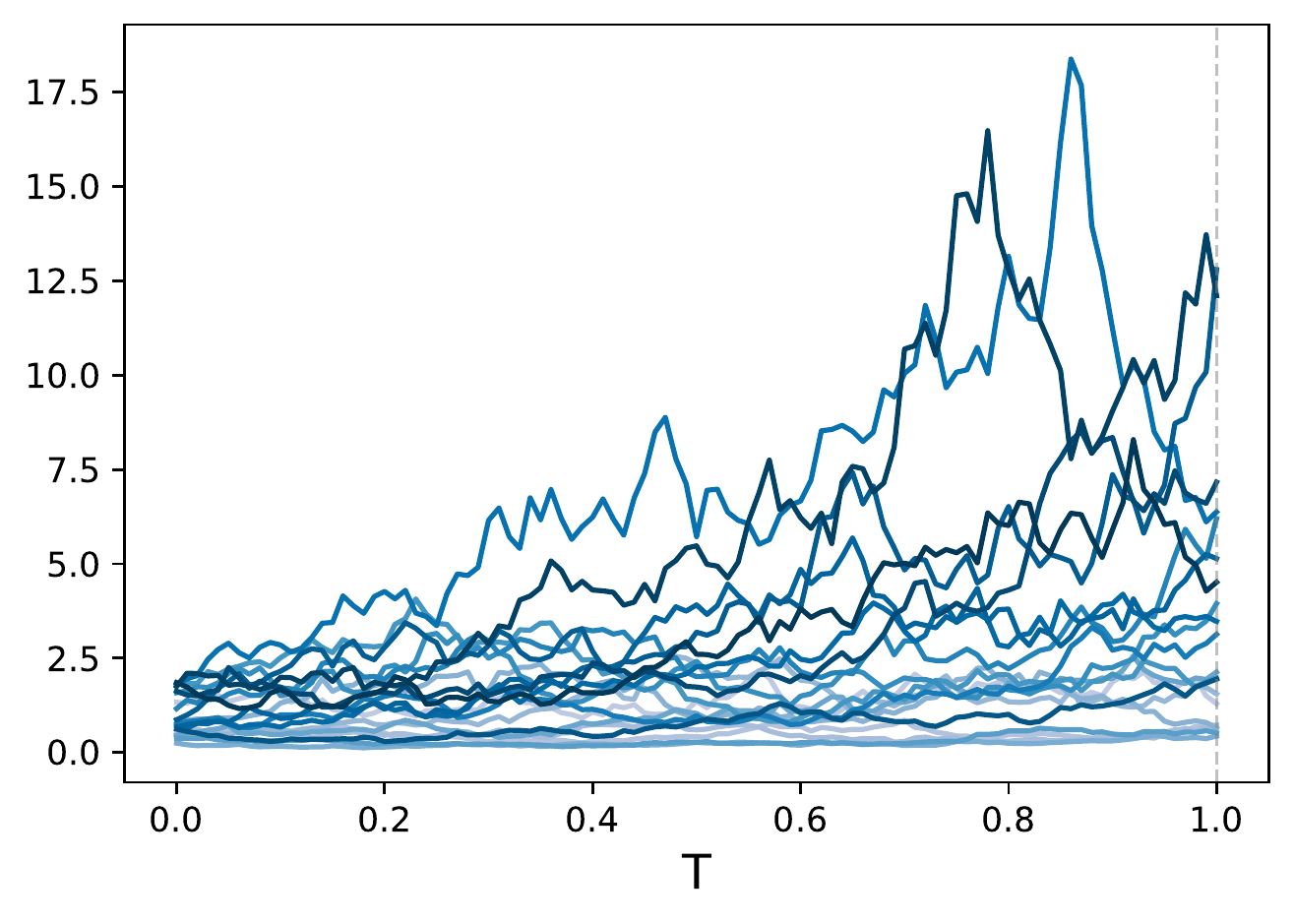}
  \includegraphics[width=.43\textwidth]{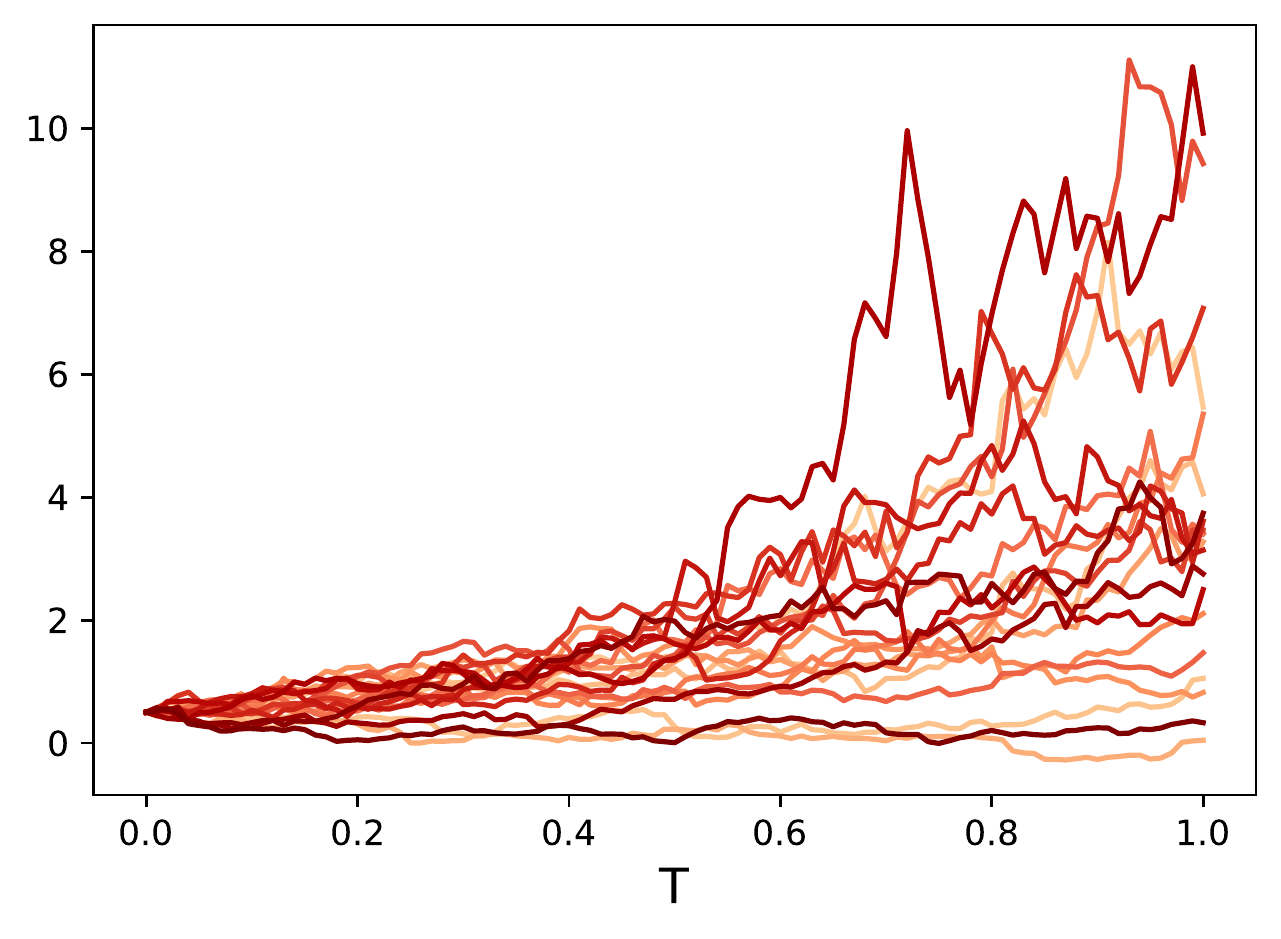}
  \caption{Solutions of the geometric Brownian motion. Left: samples
    of true solution for the training data; Right: simulated samples by the sFML model with an initial condition $x_0=0.5$.}
\end{figure}

\begin{figure}[htbp]
  \centering
  \label{fig:GBM_pdf}
  \includegraphics[width=.9\textwidth]{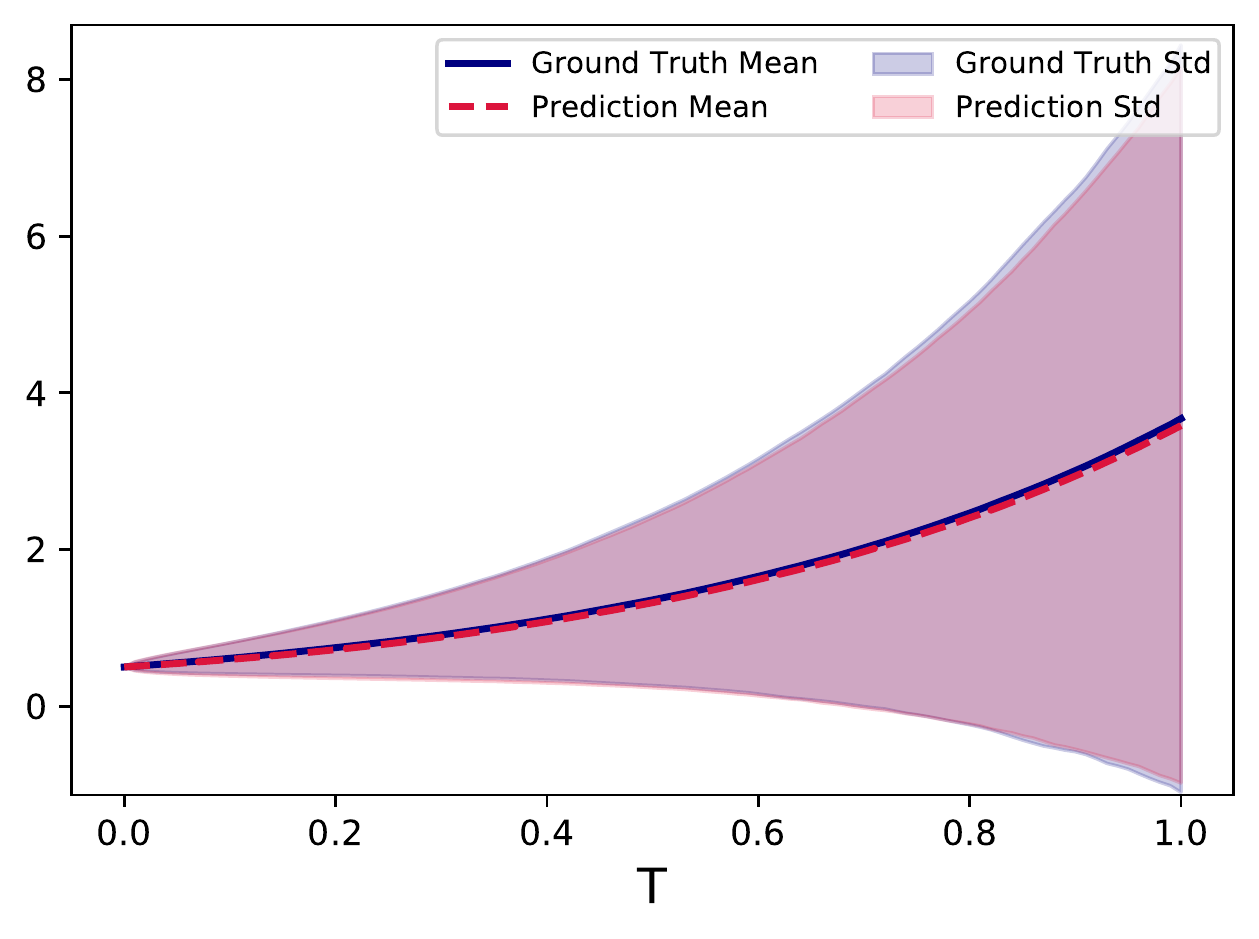}
  \caption{Mean and standard deviation of the geometric Brownian motion by the sFML model.}
\end{figure}

On the left of Figure \ref{fig:GBM_data}, some samples of the true
solutions for the training data are shown. These are generated from initial conditions uniformly distributed $\mathcal{U}(0,2)$.
On the right of Figure \ref{fig:GBM_data}, we show the simulated solution samples by the trained sFML model with an initial condition $x_0=0.5$.
The mean and STD of the solution comparison are shown in Figure \ref{fig:GBM_pdf}. Good agreement can be observed. Since this geometric Brownian motion grows exponentially fast
over time, the simulation is stopped at $T=1.0$.
%

\begin{figure}[htbp]
  \centering
  \label{fig:GBM_show}
  \includegraphics[width=.43\textwidth]{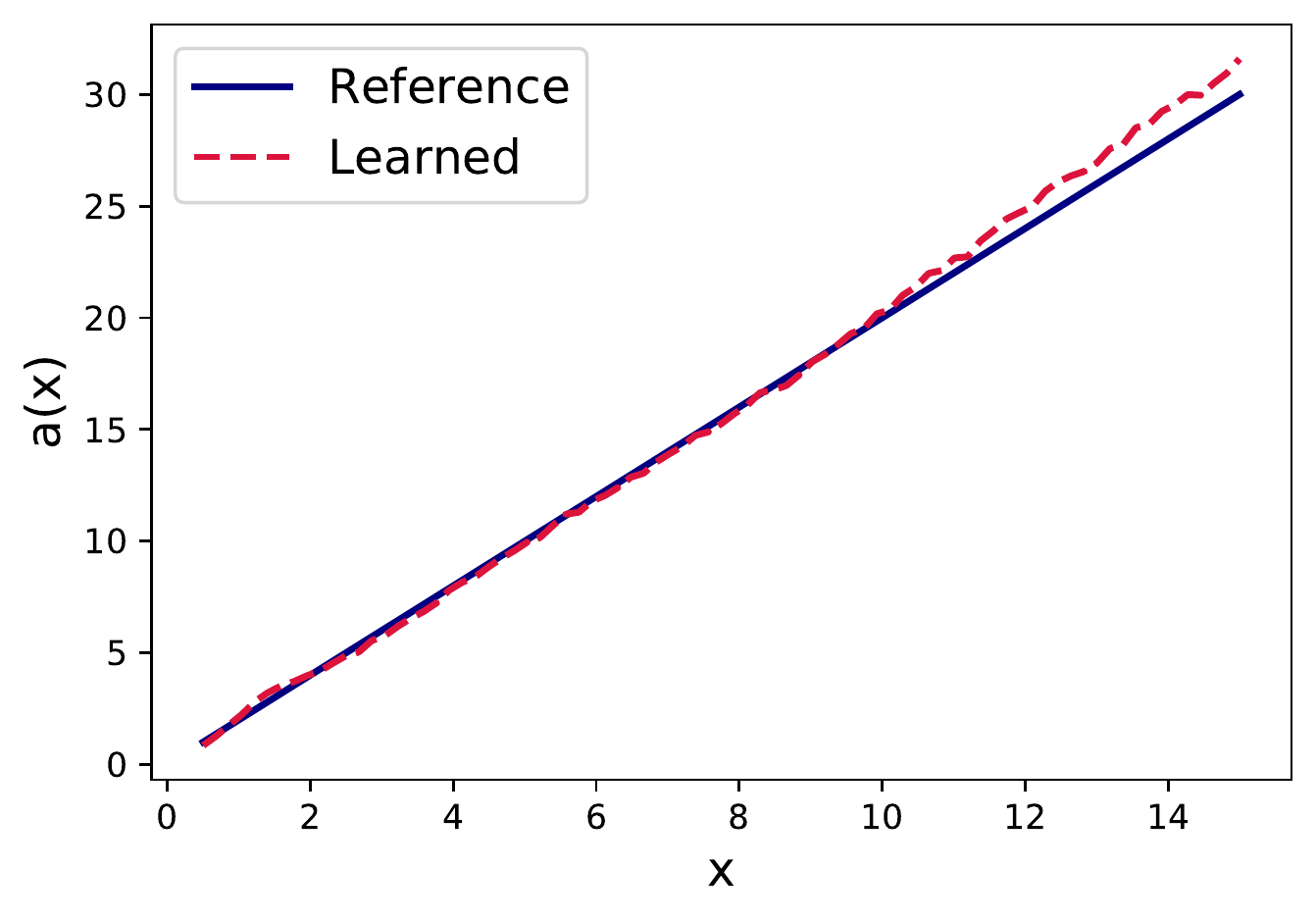}
  \includegraphics[width=.43\textwidth]{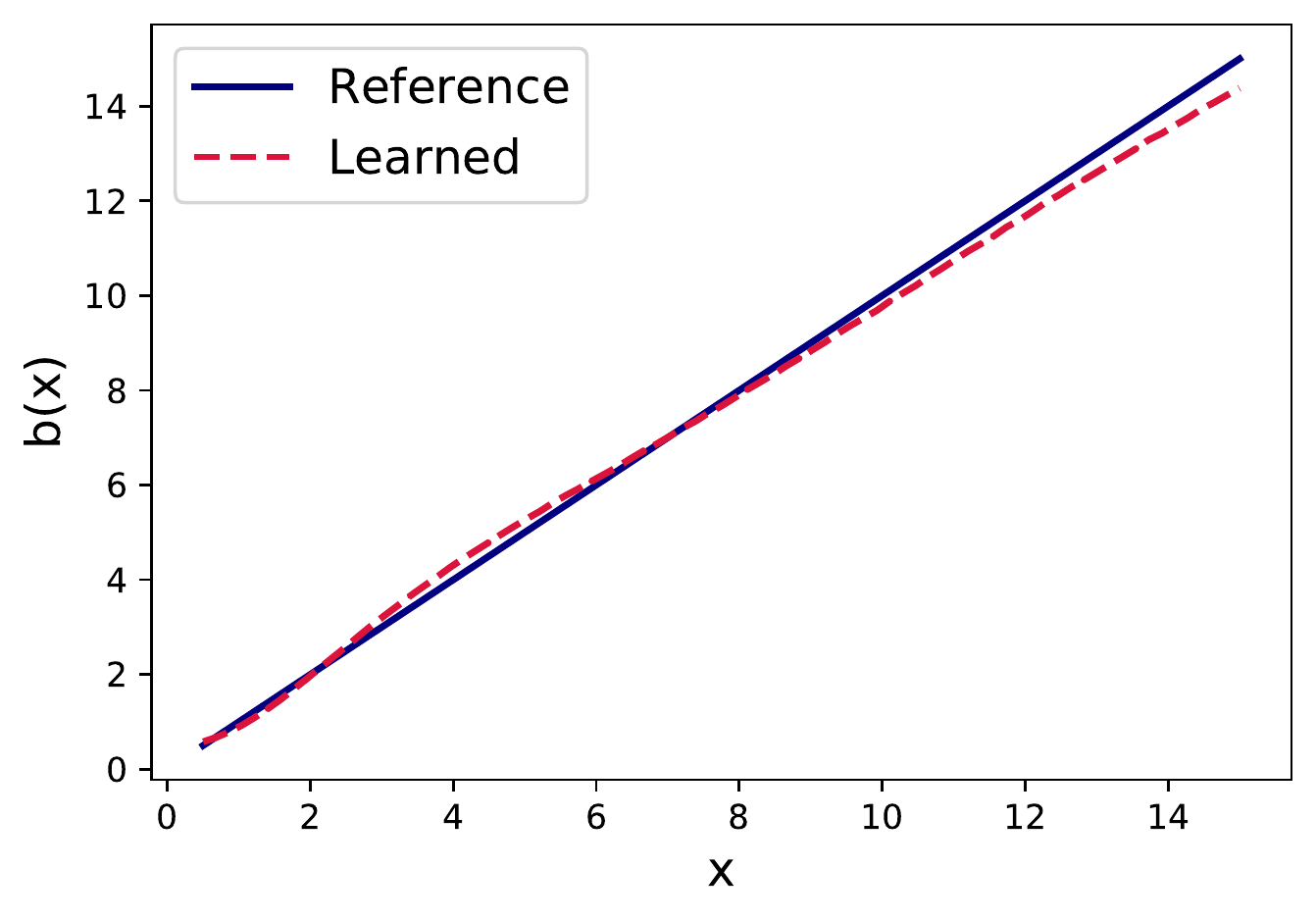}
  \caption{Geometric Brownian motion. Left: recovery of the drift $a(x)=\mu x$; Right: recovery of the diffusion $b(x)=\sigma x$.}
\end{figure}

The effective drift and diffusion functions recovered by the sFML
model are shown in 
Figure \ref{fig:GBM_show}. Good agreement with the true drift and diffusion can be observed. The conditional distribution produced by the stochastic
flow map $\Gt_\Delta(x)$ is plotted at $x=6$ (an arbitrary choice for
illustration purpose). We note that it agrees quite well with the true
distribution at $\G_\Delta(6)$.
\begin{figure}[htbp]
  \centering
  \label{fig:GBM_G}
  \includegraphics[width=.9\textwidth]{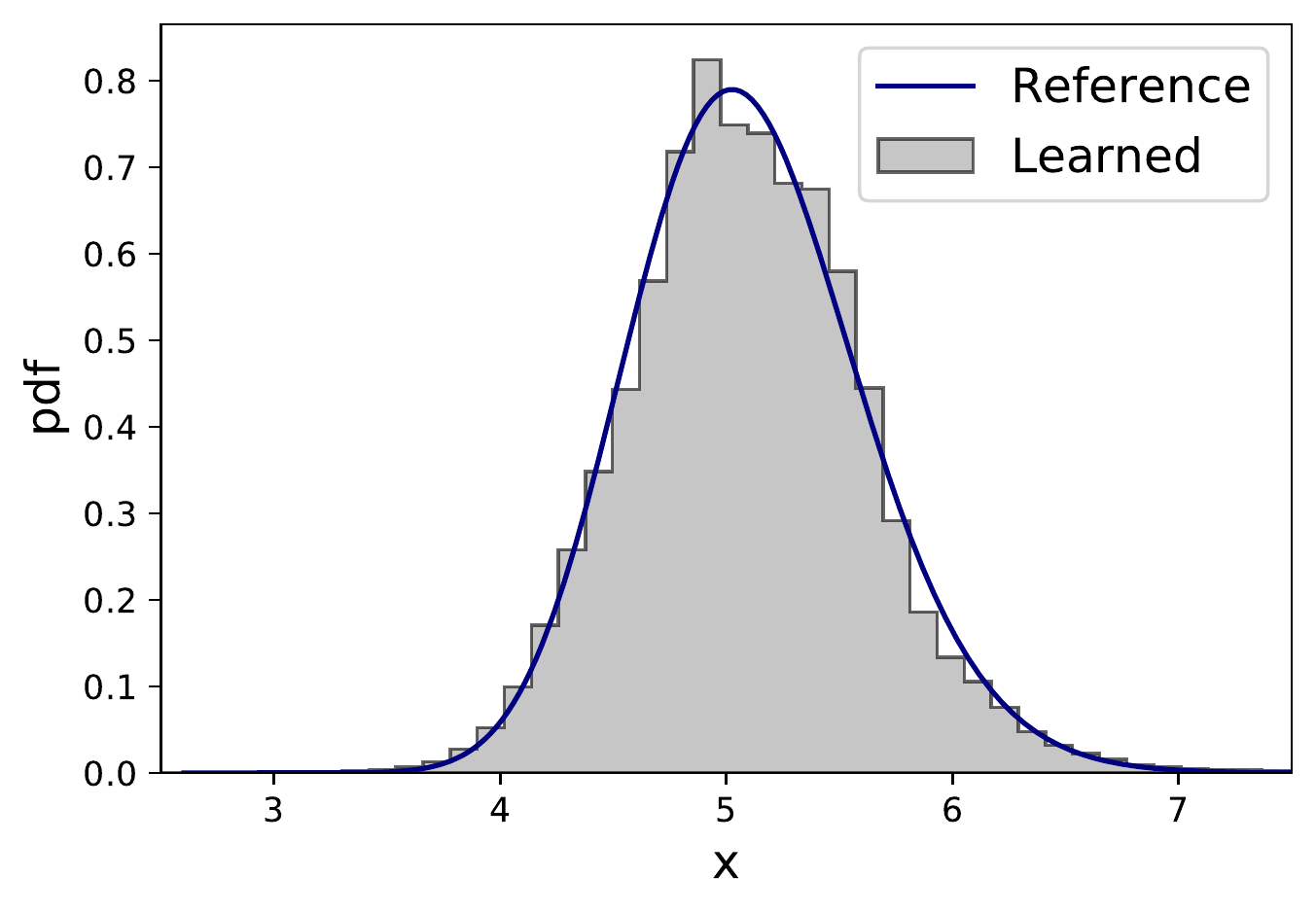}
  \caption{Geometric Brownian motion: comparison of the conditional distribution $\G_\Delta(6)$ and $\Gt_\Delta(6)$.}
\end{figure}

\subsection{Nonlinear SDE}

Similar to the work of \cite{darcy2022one}, we present two It\^{o} type SDEs with non-linear drift and diffusion functions. 

\subsubsection{SDE with nonlinear diffusion}
\begin{figure}[htbp]
  \centering
  \label{fig:Expdiff_data}
  \includegraphics[width=.43\textwidth]{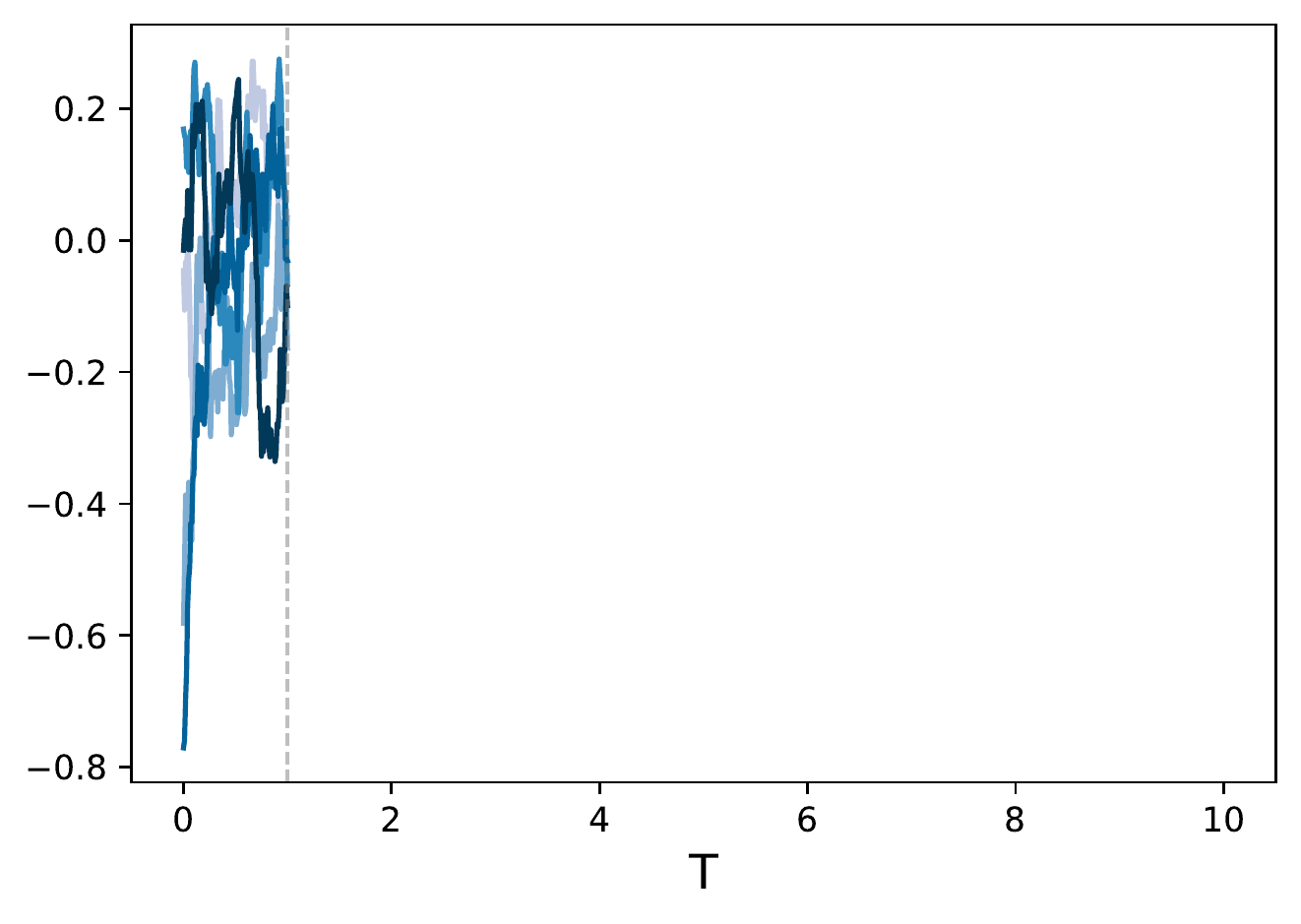}
  \includegraphics[width=.43\textwidth]{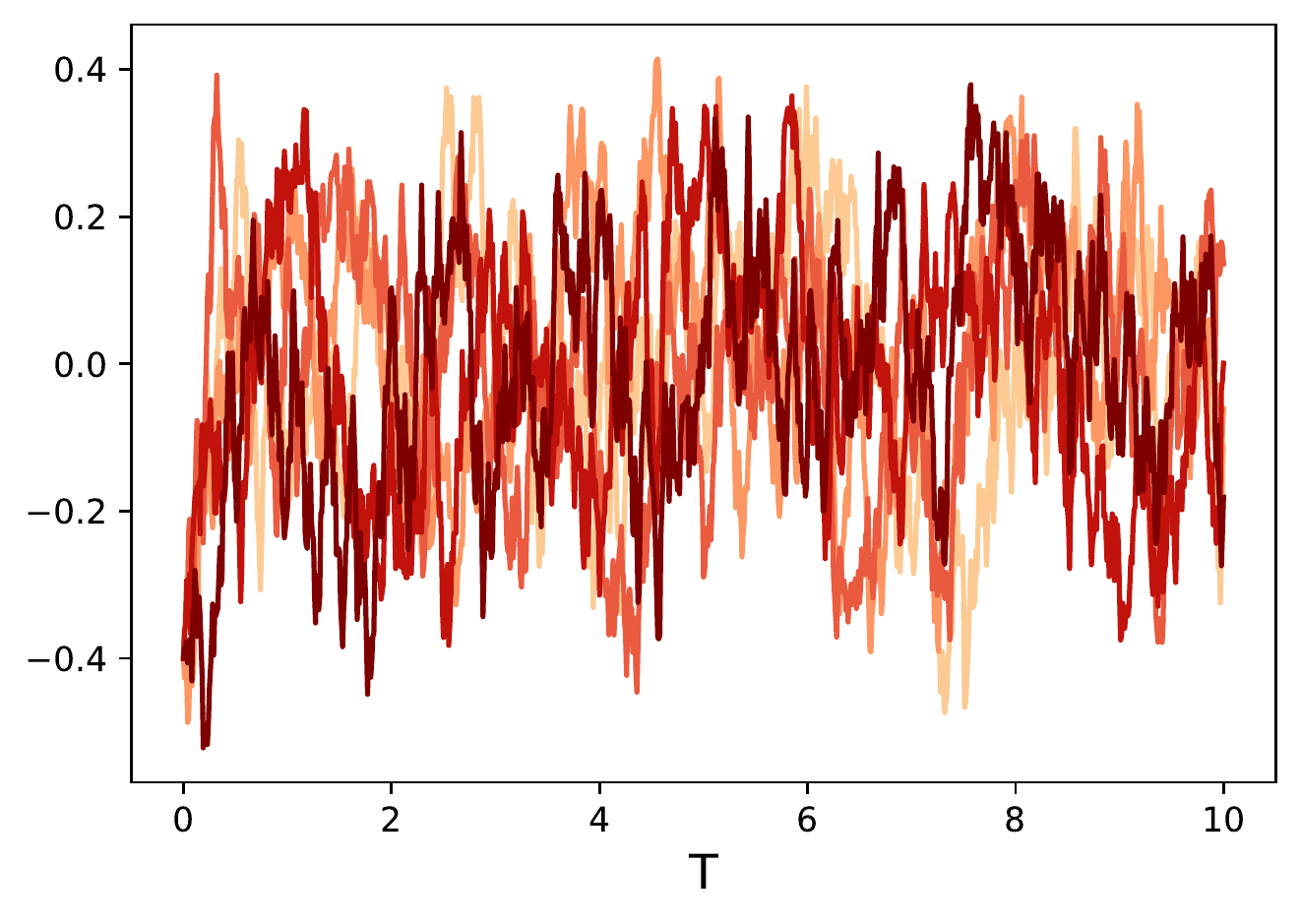}
  \caption{SDE with nonlinear diffusion \eqref{nonlinear1}. Left:
    samples of the training data; Right: samples of the learned sFML
    model simulations for up to $T=10.0$ with an initial condition $x_0=-0.4$.}
\end{figure}

\begin{figure}[htbp]
  \centering
  \label{fig:Expdiff_std}
  \includegraphics[width=.9\textwidth]{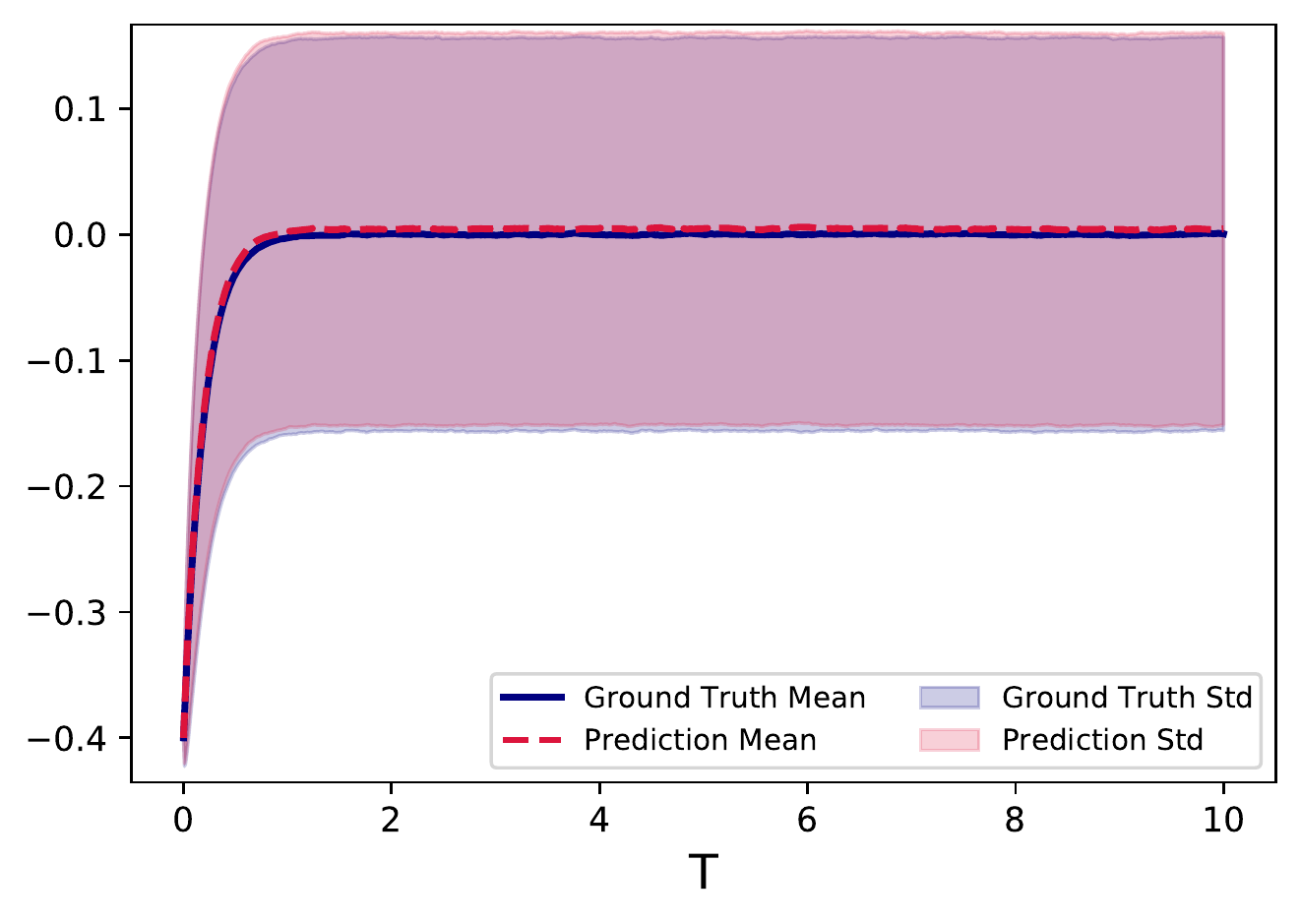}
  \caption{Mean and standard deviation of the learned sFML model for \eqref{nonlinear1}.}
\end{figure}

We first consider a SDE with a nonlinear diffusion,
\begin{equation} \label{nonlinear1}
    d x_t = -\mu x_t d t+\sigma e^{-x_t^2} d W_{t},
\end{equation}
where $\mu$ and $\sigma$ are constants. In this example, we set $\mu=5$ and $\sigma=0.5$.
The SDE is solved with initial conditions from a uniform distribution $\mathcal{U}(-1,1)$, for up to time $T=1.0$. Some of the solution samples are shown on the left of
Figure \ref{fig:Expdiff_data}. 
The simulated solutions of the learned sFML model are shown on the right of Figure \ref{fig:Expdiff_data}, with an initial condition $x_0=-0.4$ and for time up to $T=10$, for visual comparison.
The evolutions of mean and STD of the sFML model are shown in Figure
\ref{fig:Expdiff_std}, where good agreement with those of the true solution can be observed.

\begin{figure}[htbp]
  \centering
  \label{fig:Expdiff_show}
  \includegraphics[width=.43\textwidth]{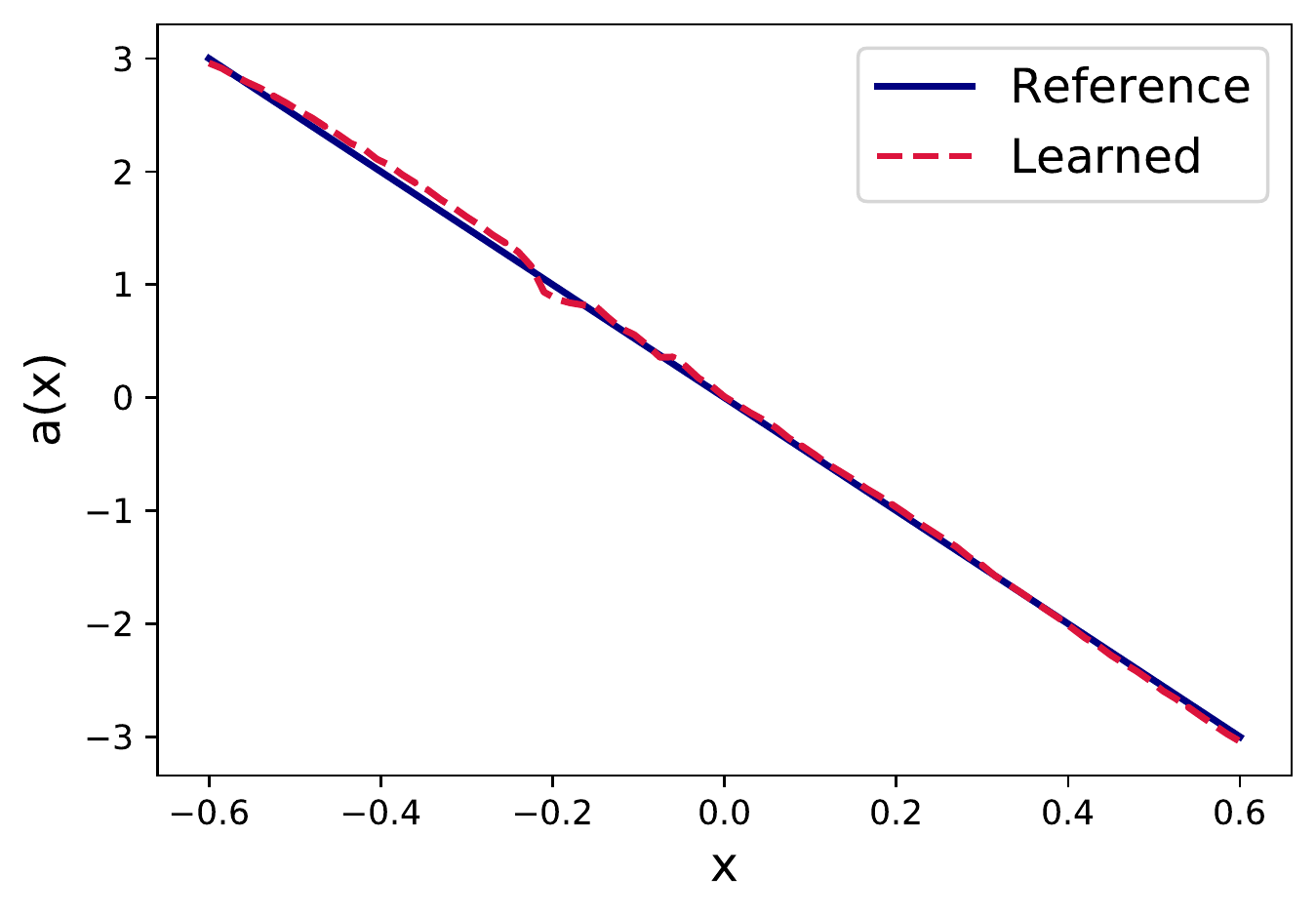}
  \includegraphics[width=.43\textwidth]{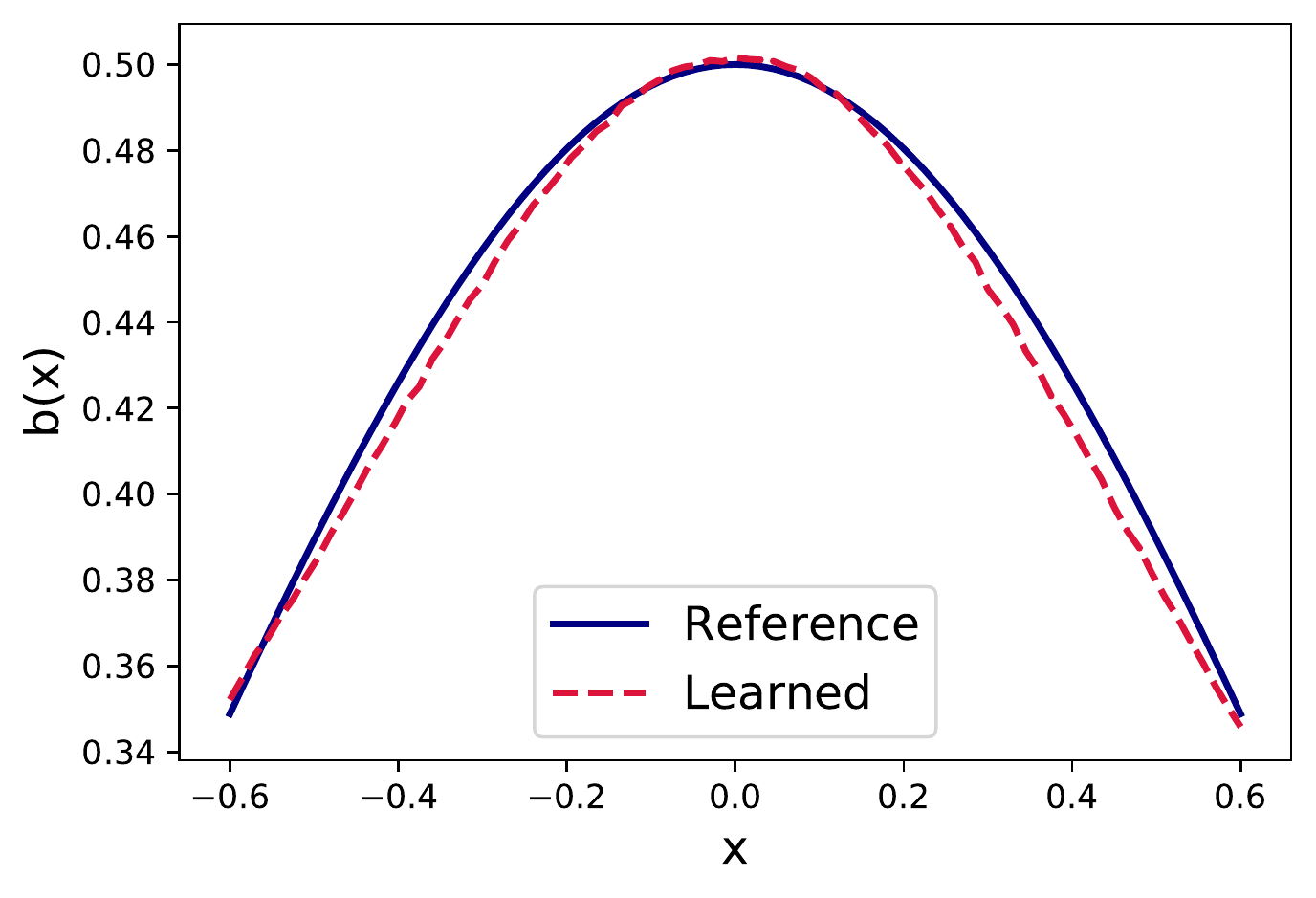}
  \caption{Nonlinear SDE \eqref{nonlinear1}. Left: recovery of the drift $a(x)=-\mu x$; Right: recovery of the diffusion $b(x)=\sigma e^{-x^2}$.}
\end{figure}

\begin{figure}[htbp]
  \centering
  \label{fig:Expdiff_pdf}
  \includegraphics[width=.9\textwidth]{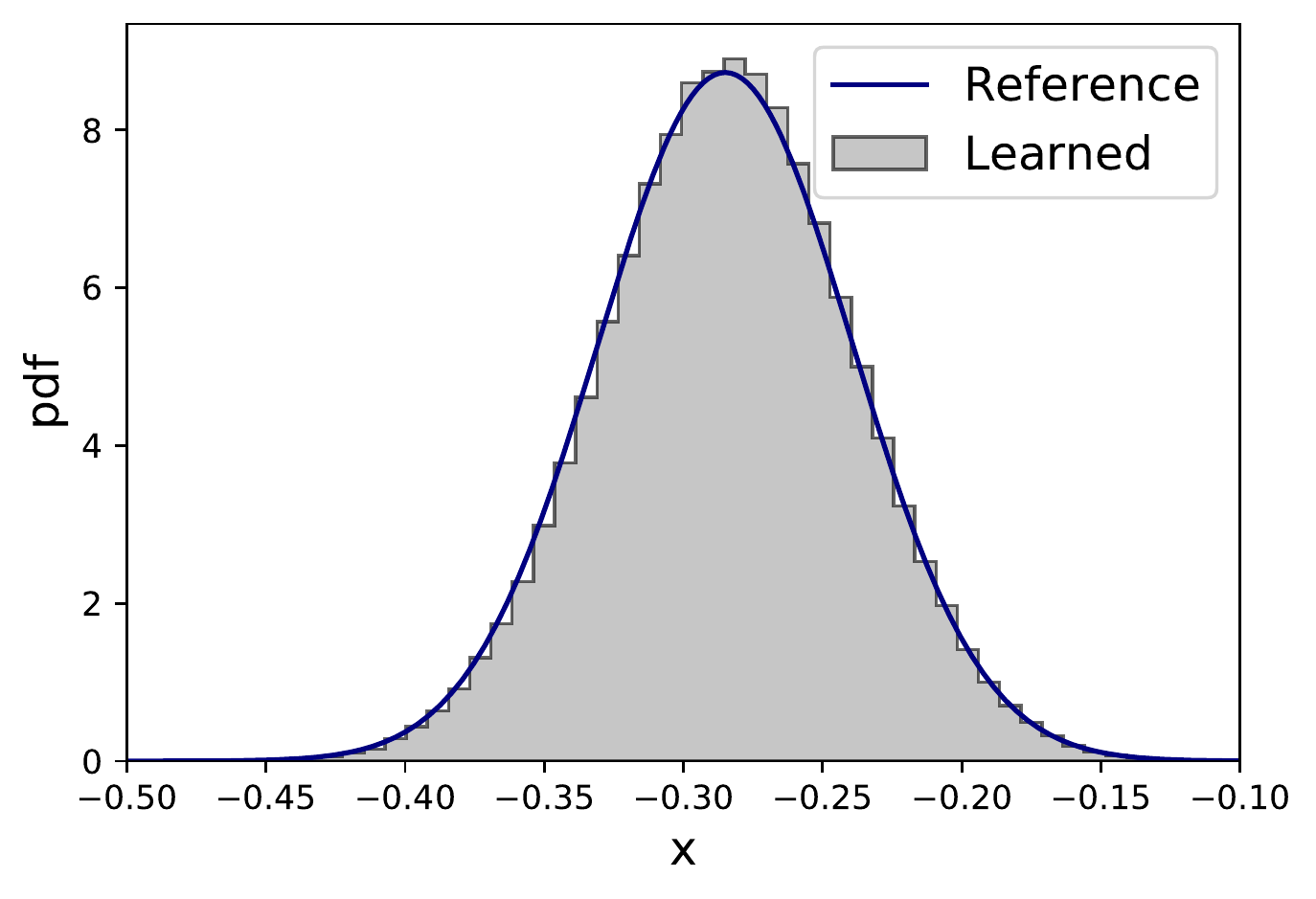}
  \caption{Nonlinear SDE \eqref{nonlinear1}. Comparison of the conditional distribution for $\G_\Delta(-0.3)$ and $\Gt_\Delta(-0.3)$.}
\end{figure}

The recovered effective drift and diffusion functions are plotted in Figure \ref{fig:Expdiff_show}, whereas the conditional distribution of $\Gt_\Delta(-0.3)$ is plotted in
Figure \ref{fig:Expdiff_pdf}. Again, good agreement can be observed when compared to the reference solutions.

\subsubsection{Trigonometric drift and diffusion}

\begin{figure}[htbp]
  \centering
  \label{fig:TrigDrift_data}
  \includegraphics[width=.43\textwidth]{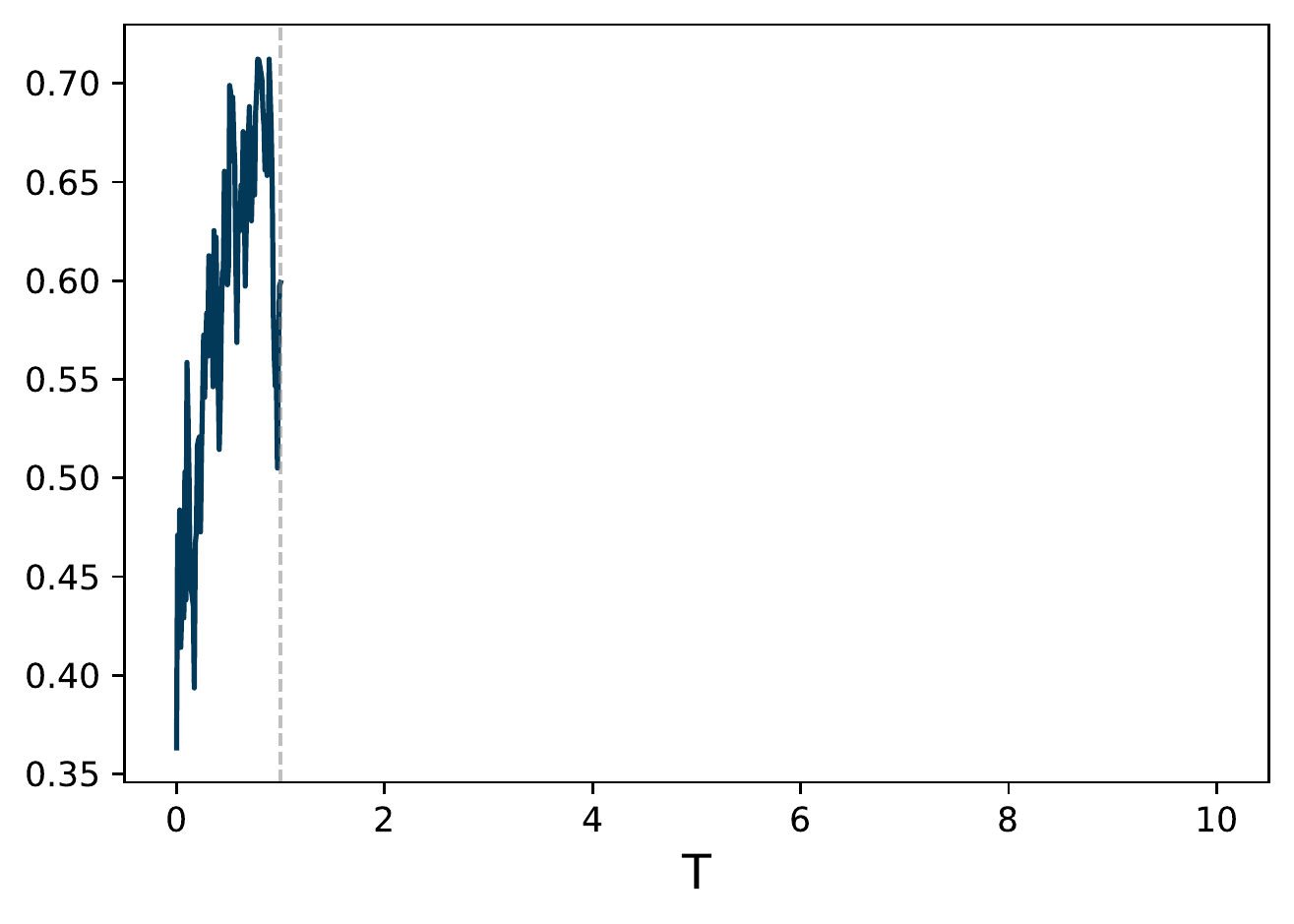}
  \includegraphics[width=.43\textwidth]{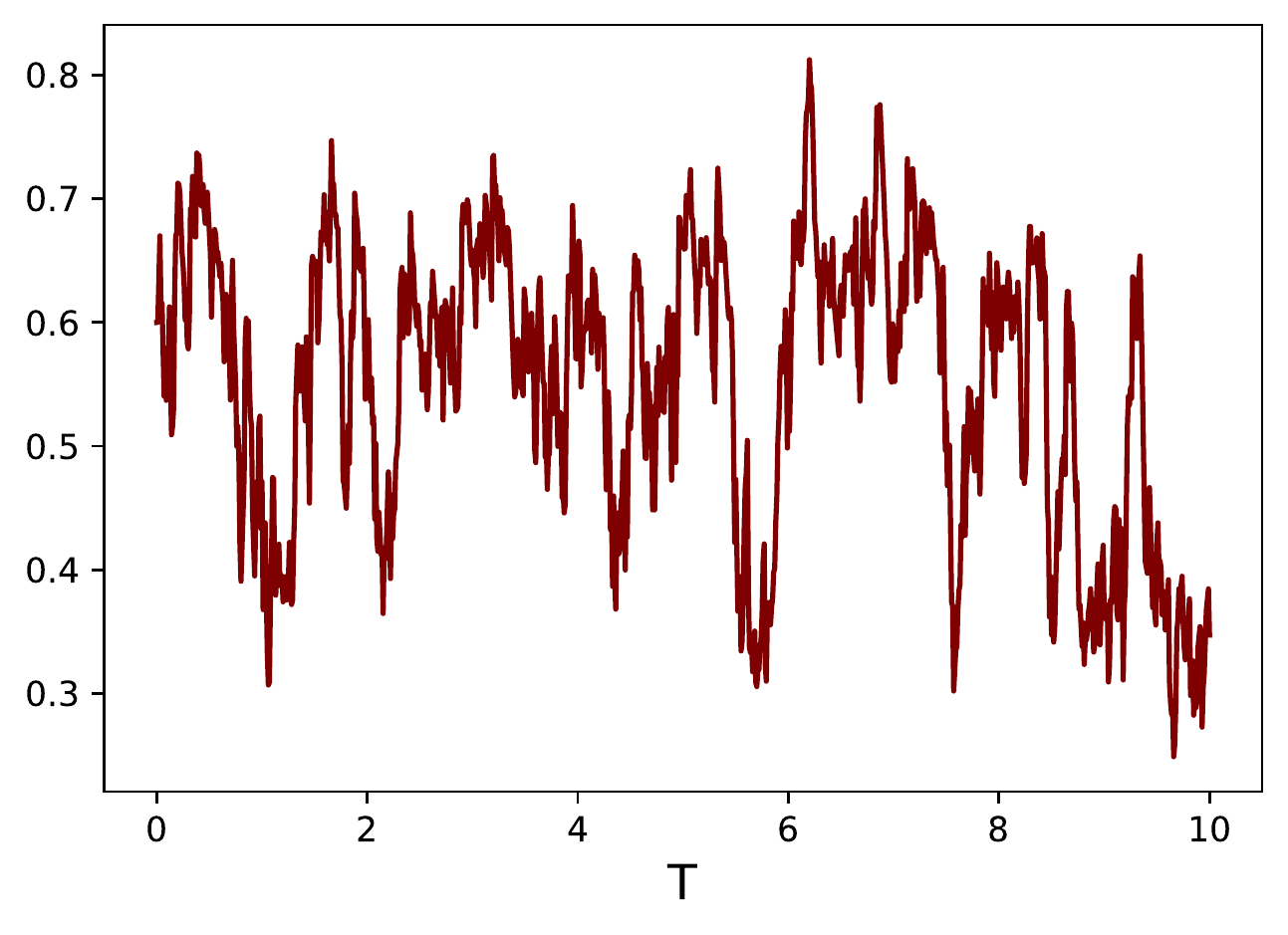}
  \caption{Nonlinear SDE \eqref{nonlinear2}. Left: a sample trajectory
    used in the training data; Right: a sample trajectory of the simulated solution by the learned sFML model for time up to $T=10.0$ with an initial condition $x_0=0.6$.}
\end{figure}

\begin{figure}[htbp]
  \centering
  \label{fig:TrigDrift_std}
  \includegraphics[width=.9\textwidth]{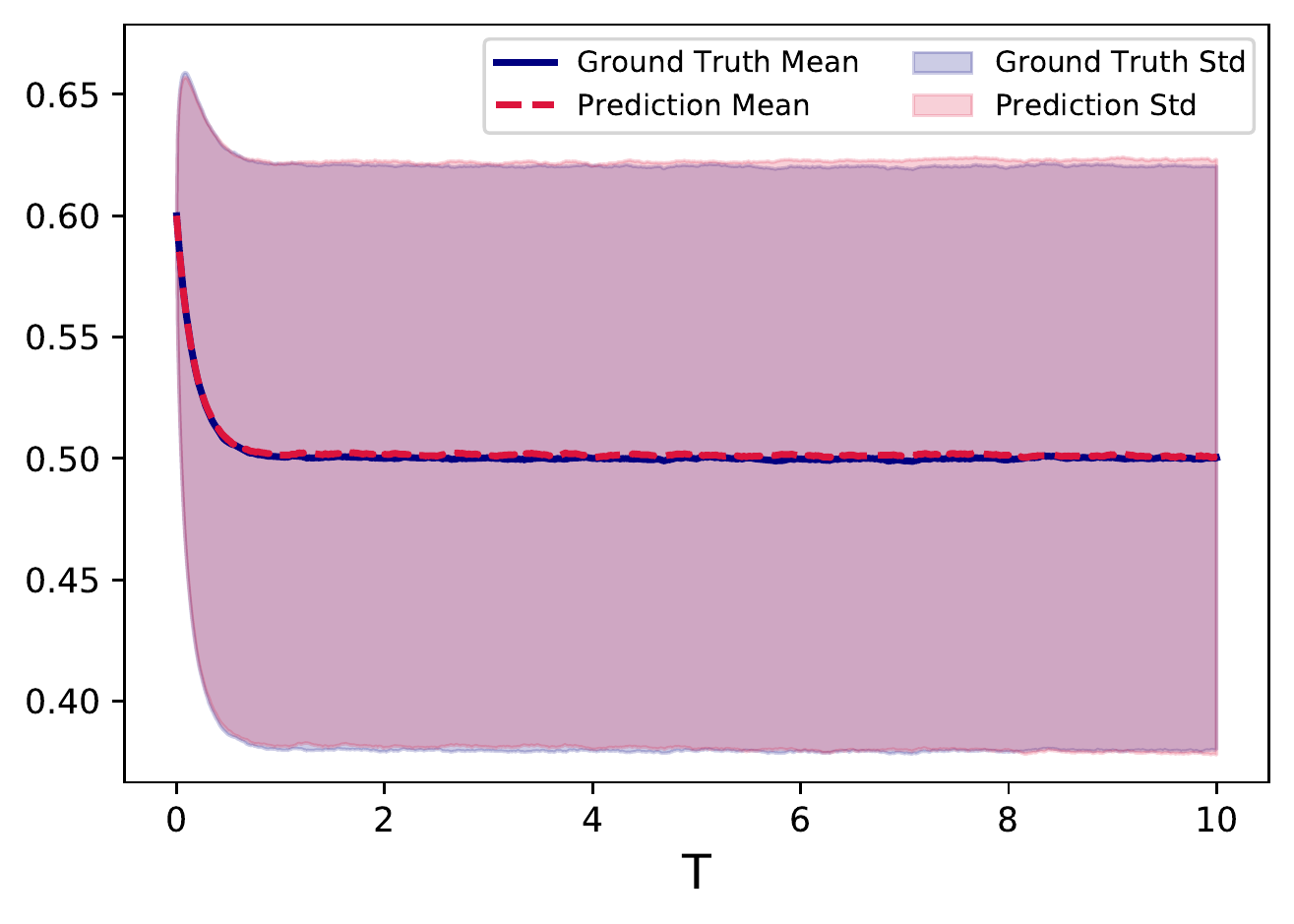}
  \caption{Mean and standard deviation of the learned sFML model for \eqref{nonlinear2}.}
\end{figure}

We now consider the following non-linear SDE:
\begin{equation} \label{nonlinear2}
    d x_t = \sin(2k\pi x_t) d t+\sigma \cos(2k\pi x_t) d W_{t},
\end{equation}
where the constant $k$ and $\sigma$ are set at $k=1$ and $\sigma=0.5$.
The training data are generated with initial conditions uniformly
distributed as $\mathcal{U}(0.35,0.7)$ for up to time $T=1.0$
as the training data).
See the left of Figure \ref{fig:TrigDrift_data} for an example trajectory.
Upon training the sFML model from the training data, we simulate the
learned process for a longer time up to $T=10$. A sample of the simulated trajectory is shown on the right of
Figure \ref{fig:TrigDrift_data}, under the initial condition $x_0=0.6$.
The mean and standard deviation of the solutions are estimated using
$100,000$ sFML model simulations and plotted Figure
\ref{fig:TrigDrift_std}. They agree very well with those of the
true solutions.
%

\begin{figure}[htbp]
  \centering
  \label{fig:TrigDrift_show}
  \includegraphics[width=.43\textwidth]{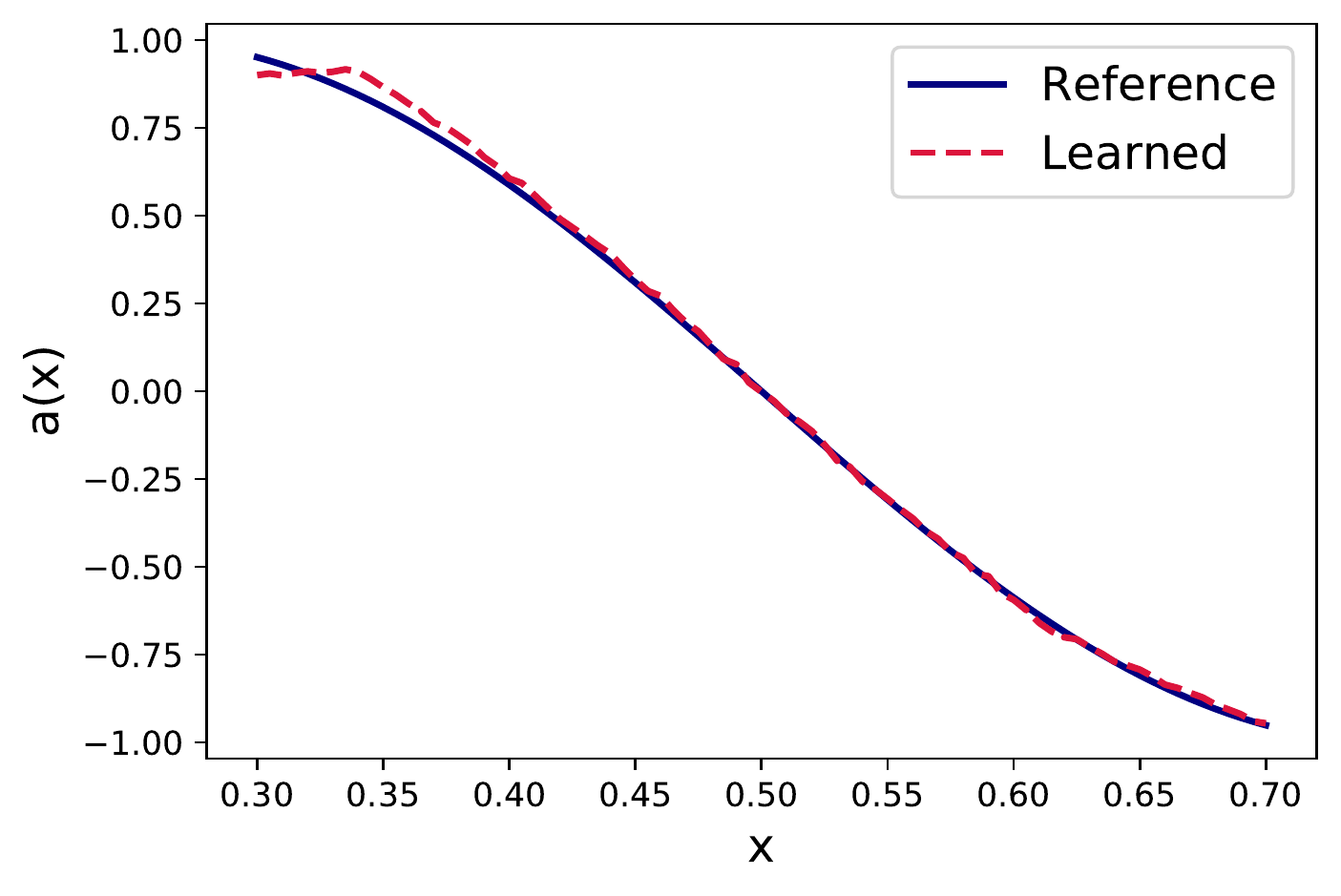}
  \includegraphics[width=.43\textwidth]{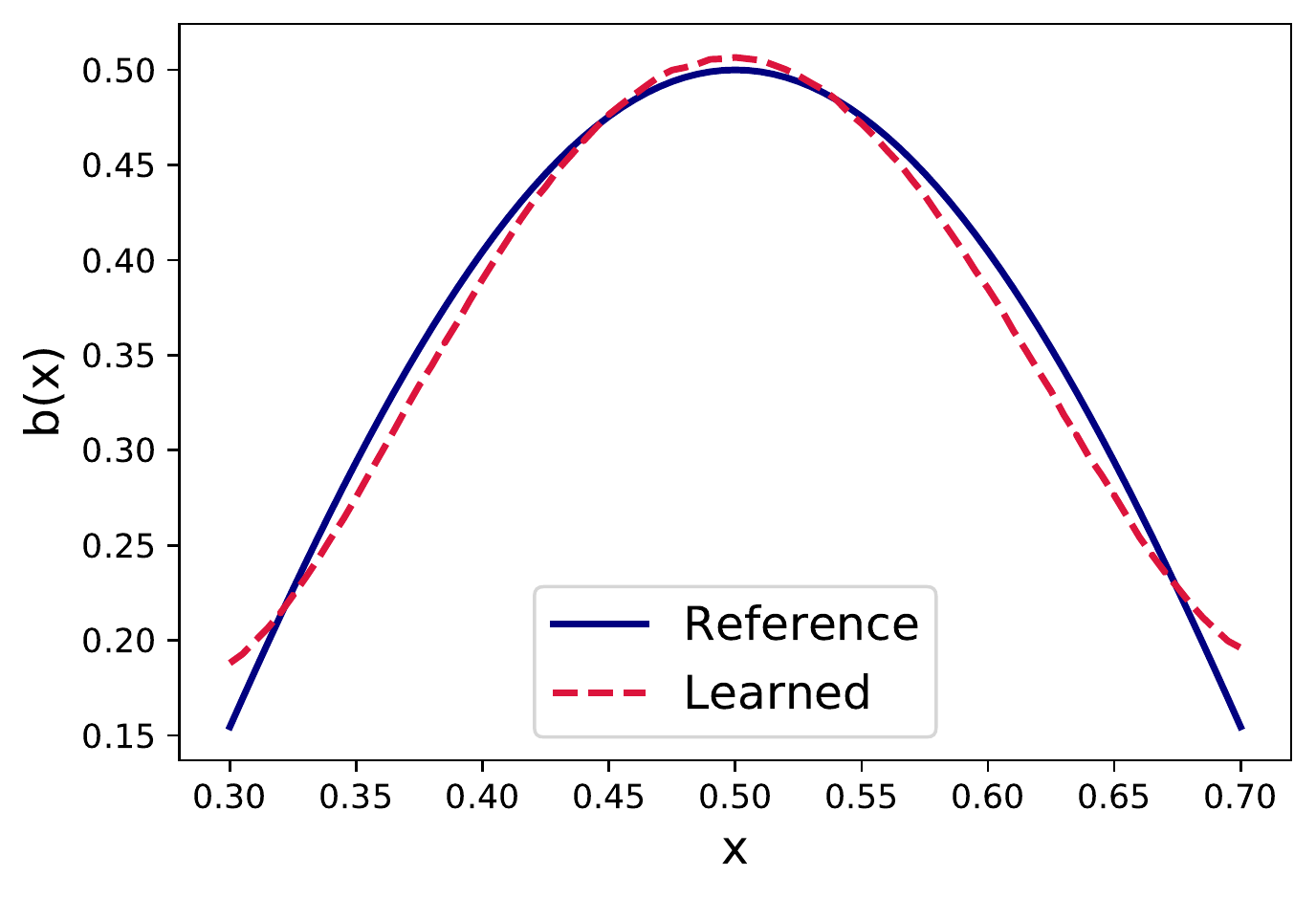}
  \caption{Nonlinear SDE \eqref{nonlinear2}. Left: recovery of the drift $a(x)=\sin(2k\pi x)$; Right: recovery of the diffusion $b(x)=\sigma \cos(2k\pi x)$.}
\end{figure}

We then compute the effective drift and diffusion functions  \eqref{abt} from the learned sFML model and compare them to the true drift and diffusion functions.
The comparison is shown in Figure \ref{fig:TrigDrift_show}, where good agreement can be seen. The accuracy starts to deterioriate near the end points, as
there are increasingly small number of samples towards the end points. Accuracy loss is thus expected. The conditional distribution by the learned sFML stochastic flow map
$\Gt_\Delta(x)$ is plotted for $x=0.5$ (an arbitrary choice for illustration purpose). It agrees well with the distribution of the true stochastic flow map $\G_\Delta(0.5)$.
\begin{figure}[htbp]
  \centering
  \label{fig:TrigDrift_pdf}
  \includegraphics[width=.9\textwidth]{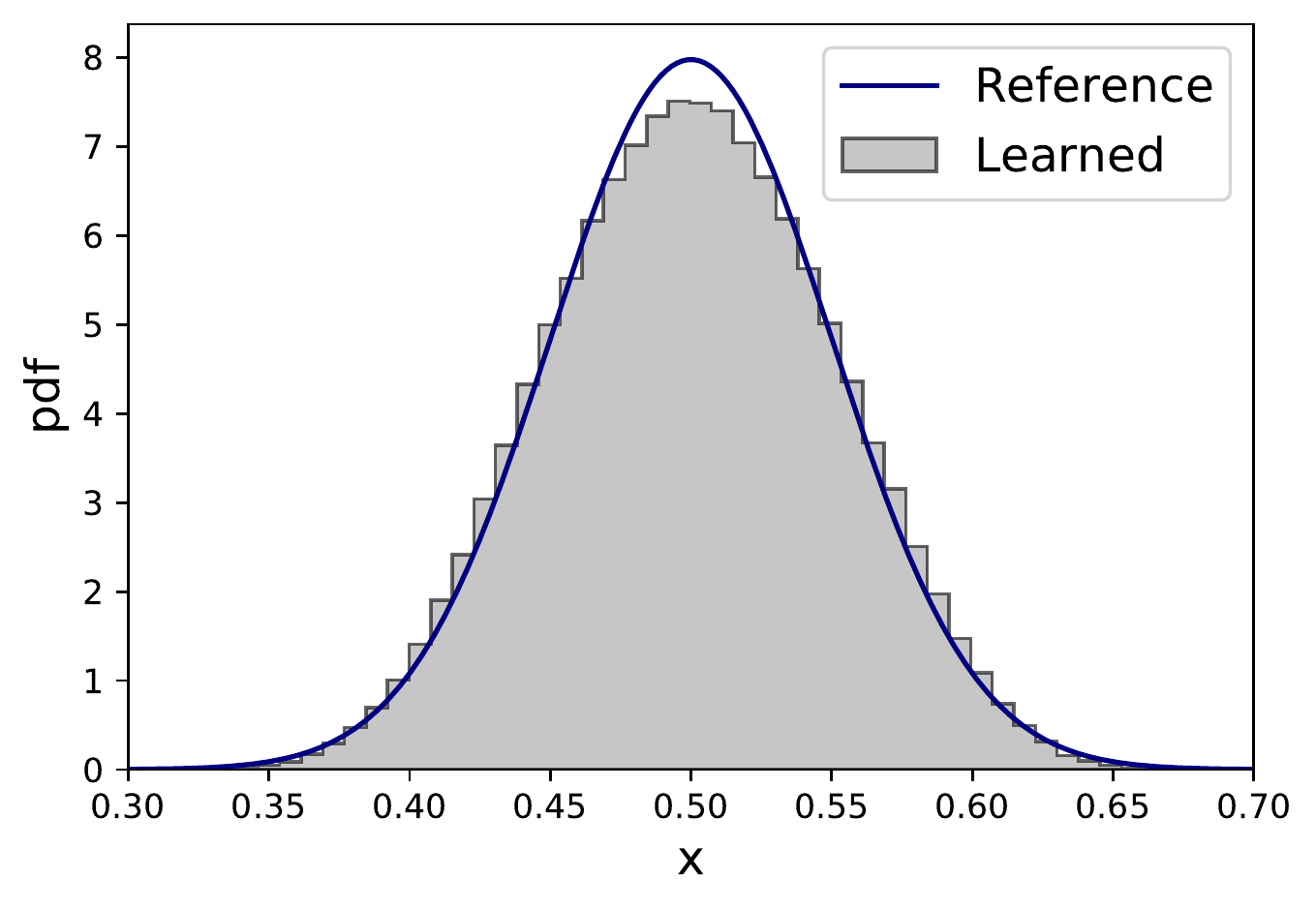}
  \caption{Nonlinear SDE \eqref{nonlinear2}: comparison of the conditional distribution of $\G_\Delta(0.5)$ and $\Gt_\Delta(0.5)$.}
\end{figure}

\subsubsection{SDE with double well potential}
We consider a one-dimensional SDE with a double well potential:
\begin{equation} \label{doublewell}
    d x_t = (x_t-x_t^3) d t+\sigma d W_{t},
\end{equation}
where the constant $\sigma$ is set as $\sigma=0.5$. The stochastic driving term is sufficiently large to induce random solution transitions between the two stable states of $x=1$ and $x=-1$.

\begin{figure}[htbp]
  \centering
  \label{fig:DW_data}
  \includegraphics[width=.34\textwidth]{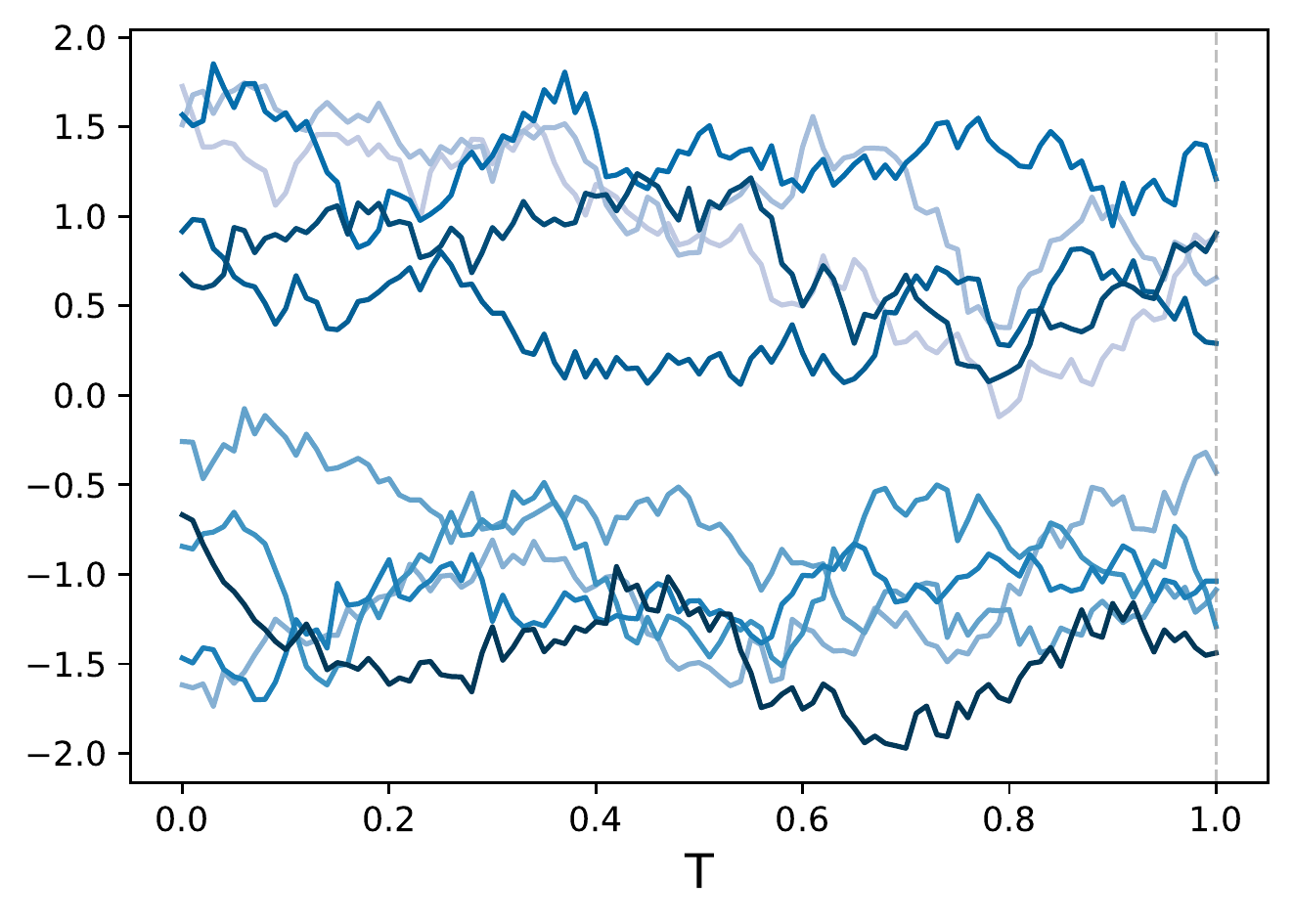}
  \includegraphics[width=.55\textwidth]{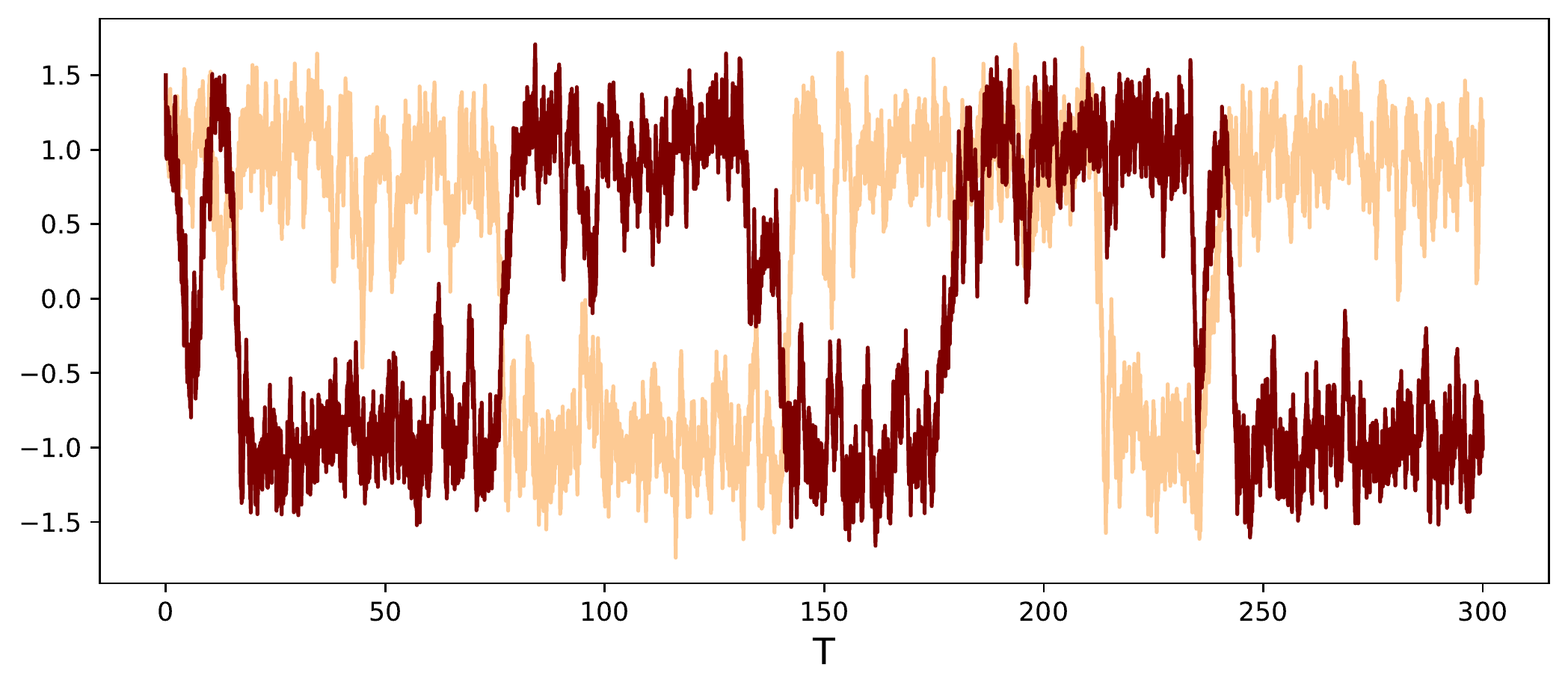}
  \caption{SDE with double well potential \eqref{doublewell}. Left:
    samples of the training date trajectories for $T=1.0$ (Note that
    there are no solution transitions.); Right: solution trajectories
    by the learned sFML model for time $T=300.0$, with an initial
    condition $x_0=1.5$. Note the state transitions occur over time.} 
\end{figure}

The training data are generated by solving the true SDE with initial
conditions uniformly sampled in $\mathcal{U}(-2.5,2.5)$ for up to
$T=1.0$. A few such sample
trajectories are shown on the left of Figure \ref{fig:DW_data}. We
emphasize that the trajectories in the training data set are so short
that they do not contain transitions between the two stable states. 
Using these short bursts of training data, we train the sFML model and
conduct long-term predictions. The results of two solution
trajectories with an initial condition $x_0=1.5$  for time up to
$T=300$ are shown 
on the right of Figure \ref{fig:DW_data}. We clearly observe the
solution transition between the two stable states. The transitions
occur with a small probability and over a time frame on the order of $O(10)$. Thus, the transitions are not observed in the training data, which span only $T=0.4$. This result demonstrates the learning capability of sFML -- the learned model is able
to produce the correct solution behaviors even though they may not be
present in the training data.  
%

\begin{figure}[htbp]
  \centering
  \label{fig:DW_show}
  \includegraphics[width=.43\textwidth]{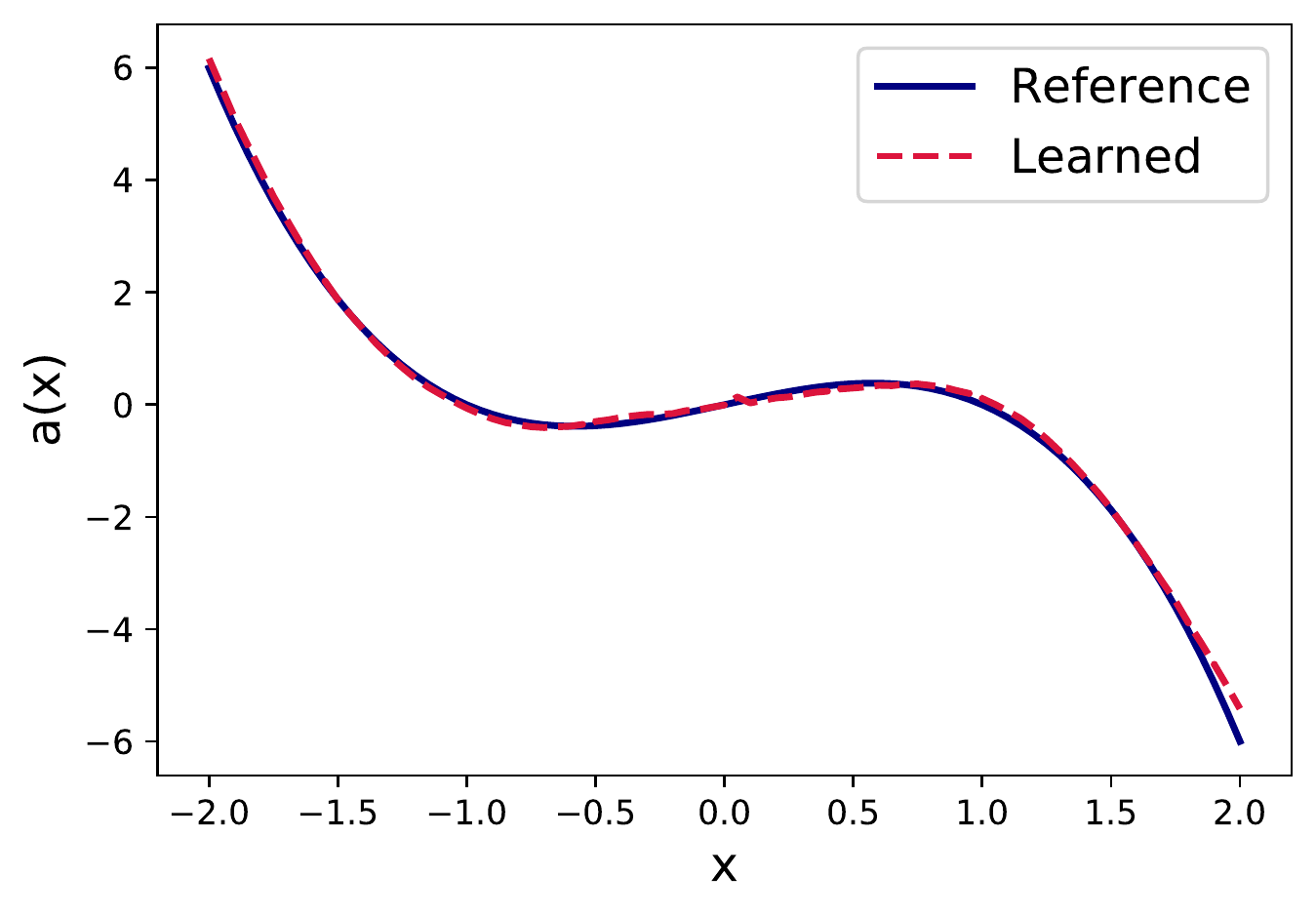}
  \includegraphics[width=.43\textwidth]{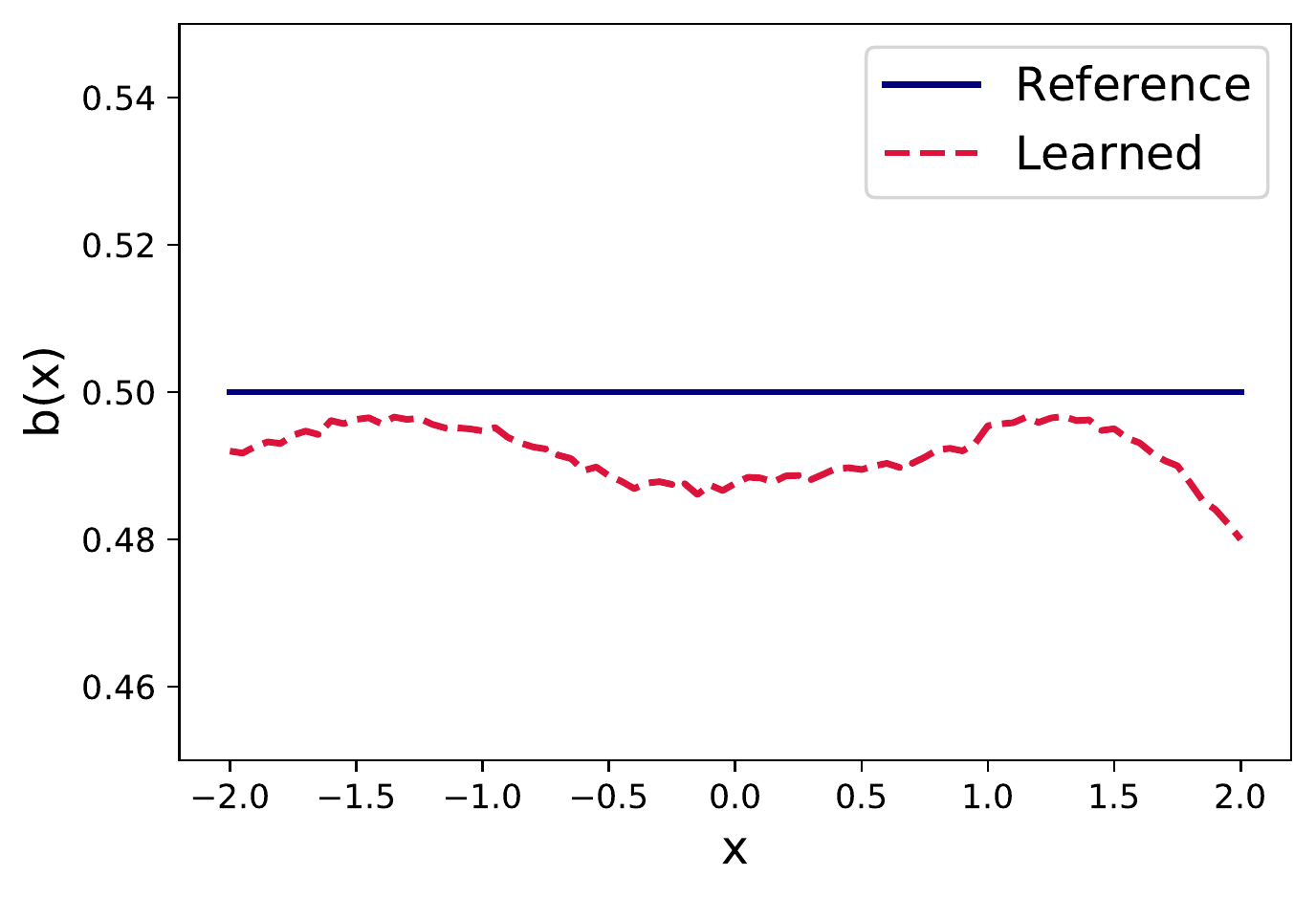}
  \caption{SDE with double well potential \eqref{doublewell}. Left: recovery of the drift $a(x)=x-x^3$; Right: recovery of diffusion $b(x)=0.5$.}
\end{figure}

\begin{figure}[htbp]
  \centering
  \label{fig:DW_pdf}
  \includegraphics[width=.24\textwidth]{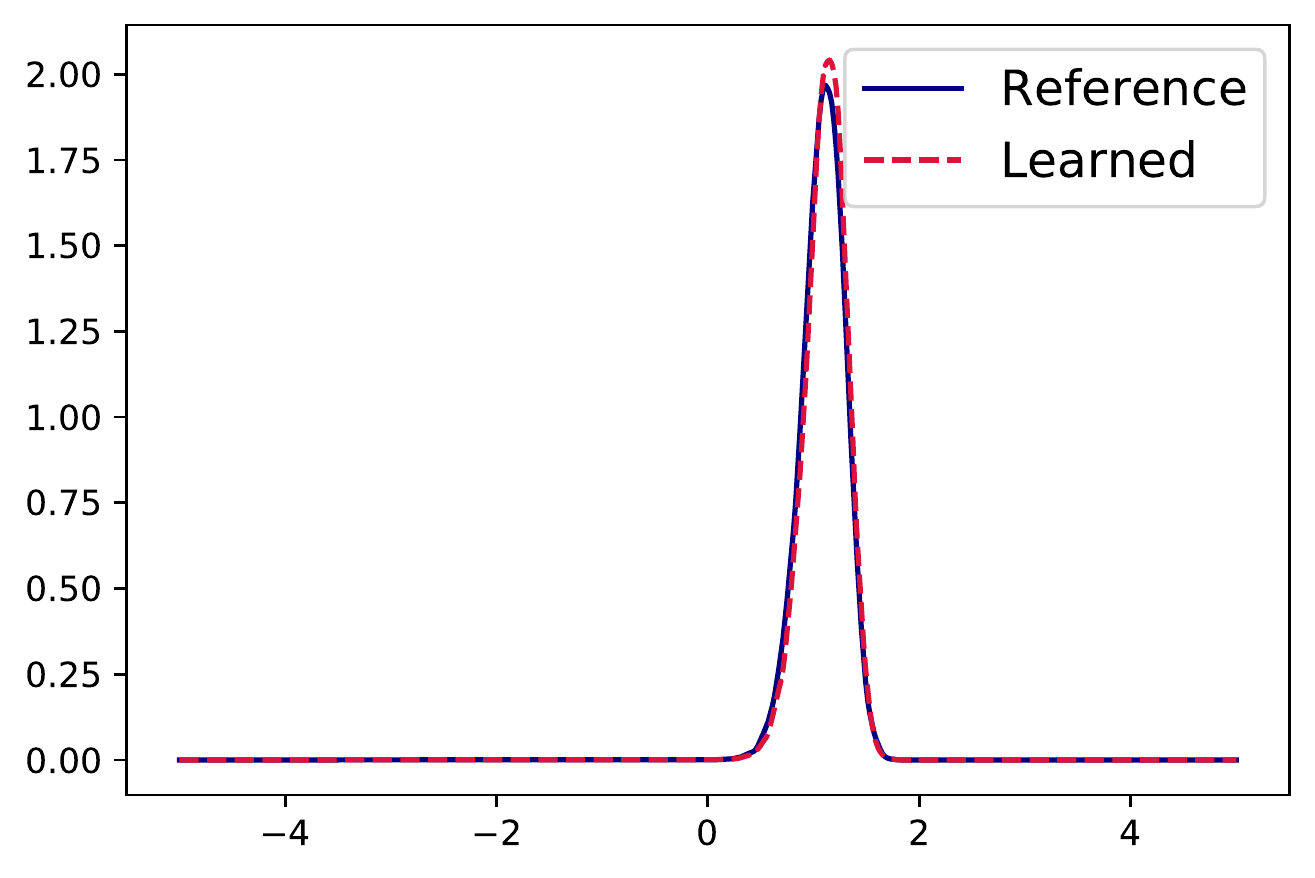}
  \includegraphics[width=.24\textwidth]{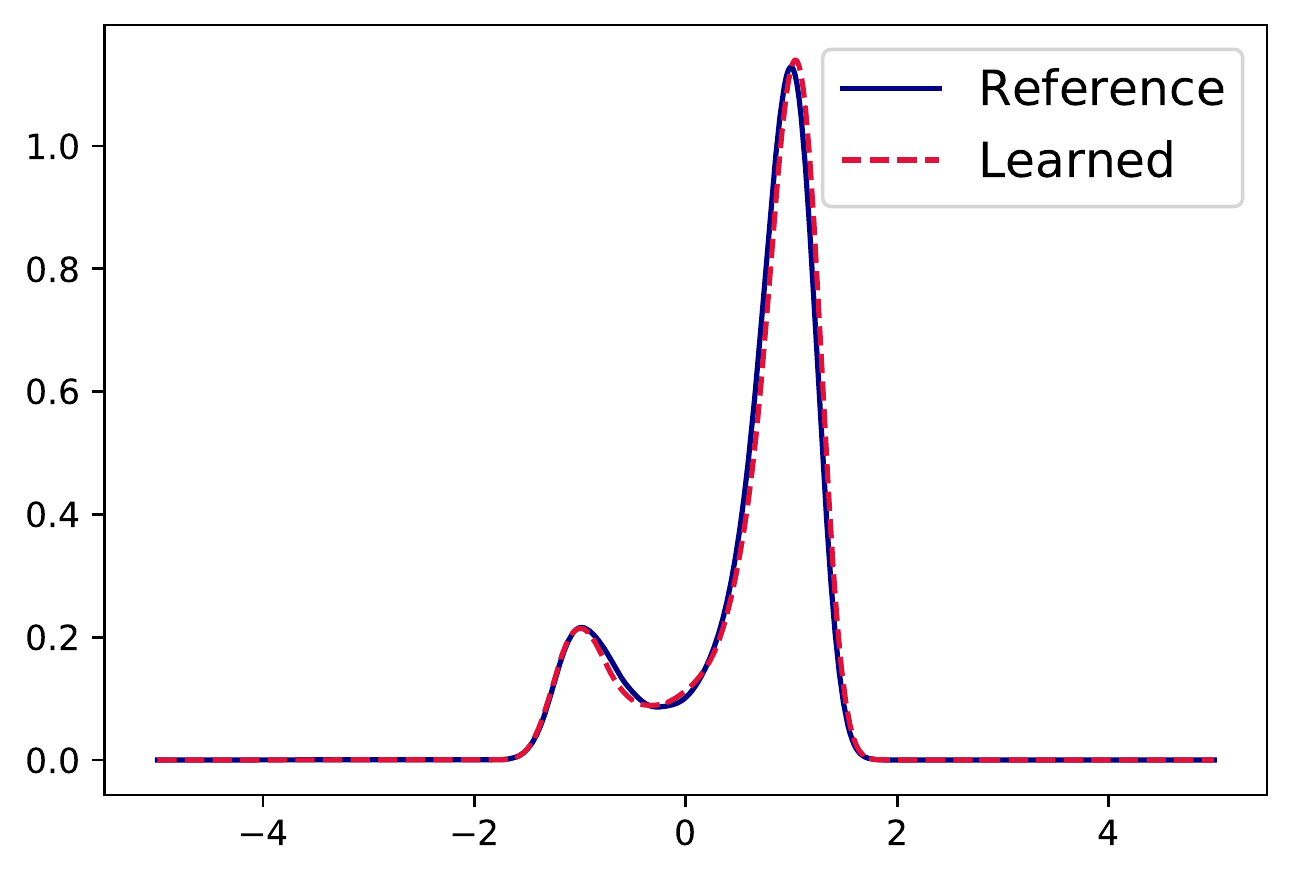}
  \includegraphics[width=.24\textwidth]{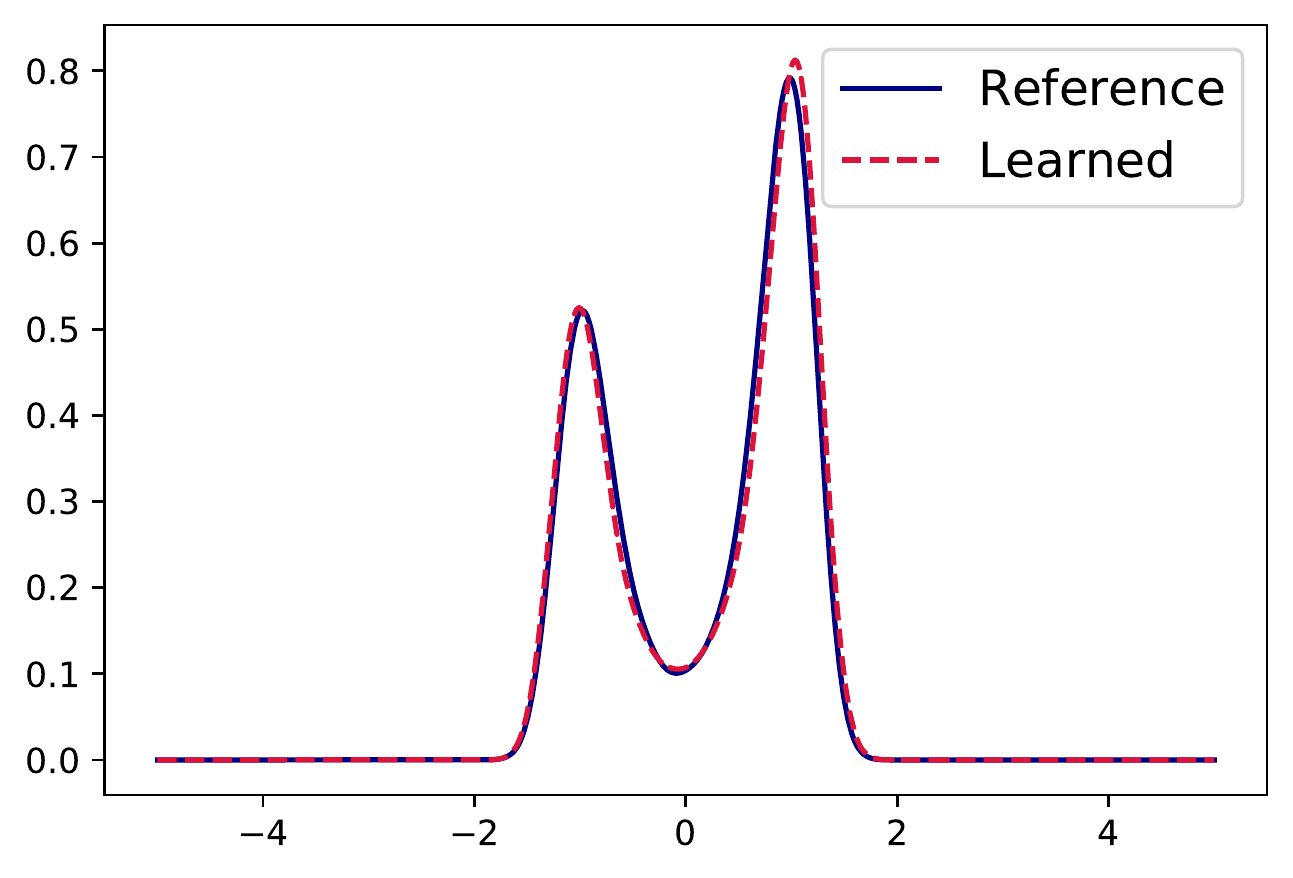}
  \includegraphics[width=.24\textwidth]{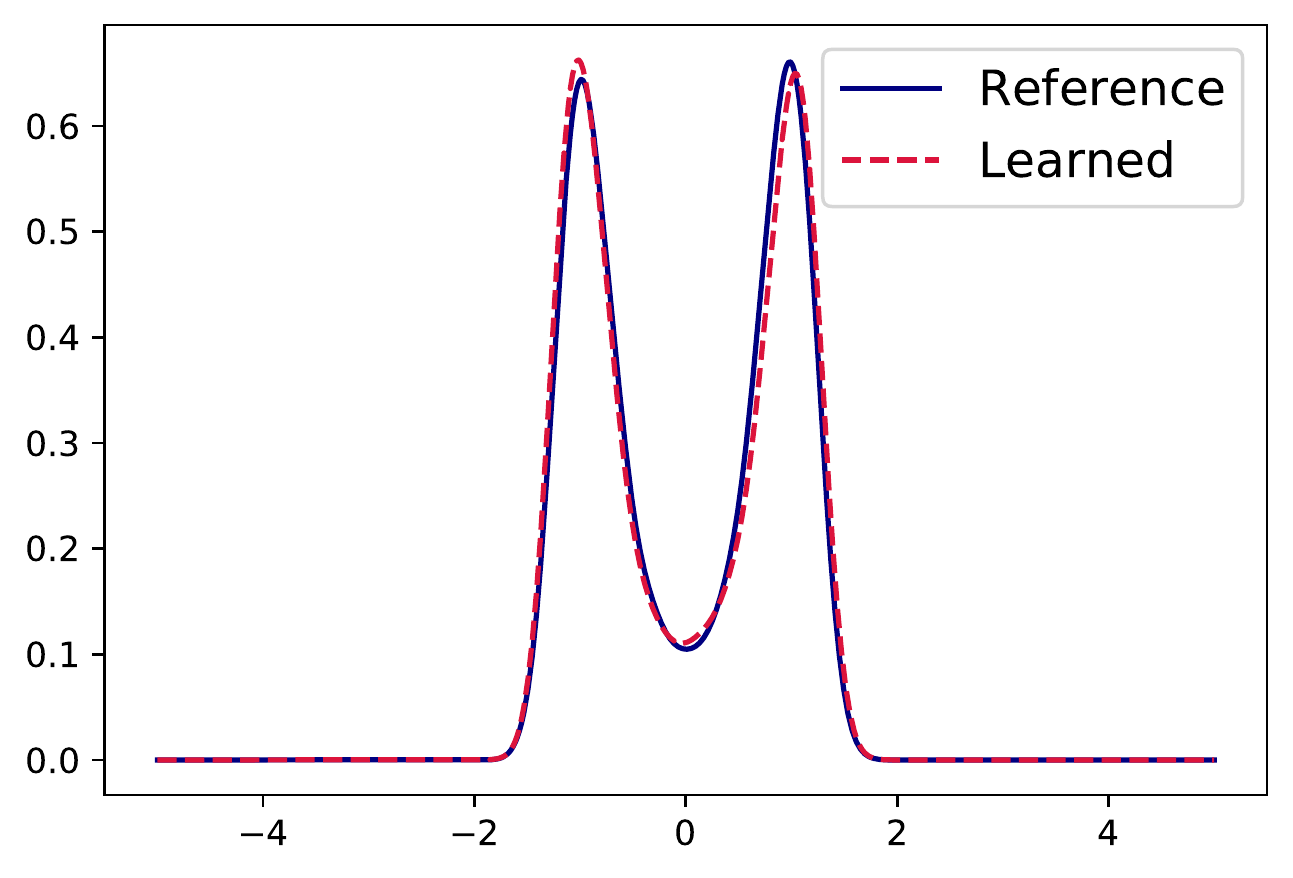}
  \caption{SDE with double well potential \eqref{doublewell}. The
    predicted solution pdf's with an initial condition $x_0=1.5$ against
    those of the reference true solution at time $T=0.5$, $10$,
    $30$ and $100$ (from left to right).}
\end{figure}

We then recover the drift and diffusion functions by computing the effective drift and diffusion functions \eqref{abt}. The results are shown in Figure \ref{fig:DW_show}, where good agreement
in the drift function is observed. Although the error in the constant diffusion function is visible, it is rather acceptable at about $2\%$.
The evolution of the probability distribution of the solution is shown in  
Figure \ref{fig:DW_pdf}, at various time levels $T=0.5, 10.0, 30.0, 100.0$. 
It is started from an initial condition $x_0=1.5$ and computed by
using $100,000$ simulated trajectories by the sFML model. Excellent
agreement with the reference
solution from the true SDE is evident.
%

\subsection{SDEs with Non-Gaussian Noises}

The applicability of the proposed sFML approach is not restricted
to modeling the classical SDE with Gaussian noises from Wiener process. To demonstrate this, we consider SDEs with
non-Gaussian noises.

\subsubsection{Noises with Exponential Distribution}
Let us consider a SDE with exponentially distributed noise inputs,
\begin{equation}
    d x_{t} = \mu  x_t dt + \sigma \sqrt{dt} \eta_t,\qquad \eta_t \sim \text{Exp}(1),
\end{equation}
where $\eta_t$ has an exponential pdf $f_{\eta}(x)=e^{-x}$, $x\geq
0$, and the constants are set as $\mu=-2.0$ and $\sigma=0.1$.

\begin{figure}[htbp]
  \centering
  \label{fig:Expdis_data}
  \includegraphics[width=.43\textwidth]{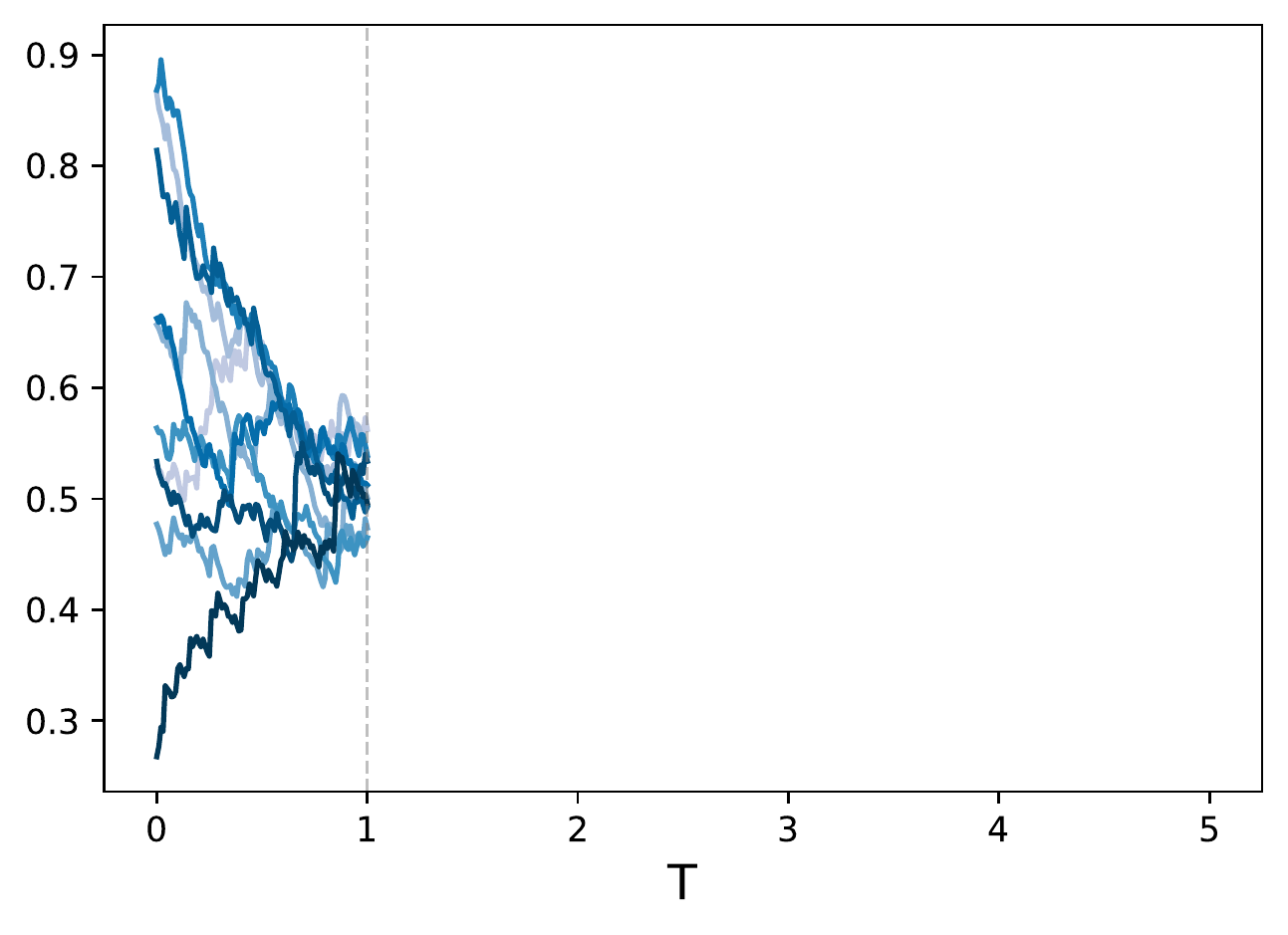}
  \includegraphics[width=.43\textwidth]{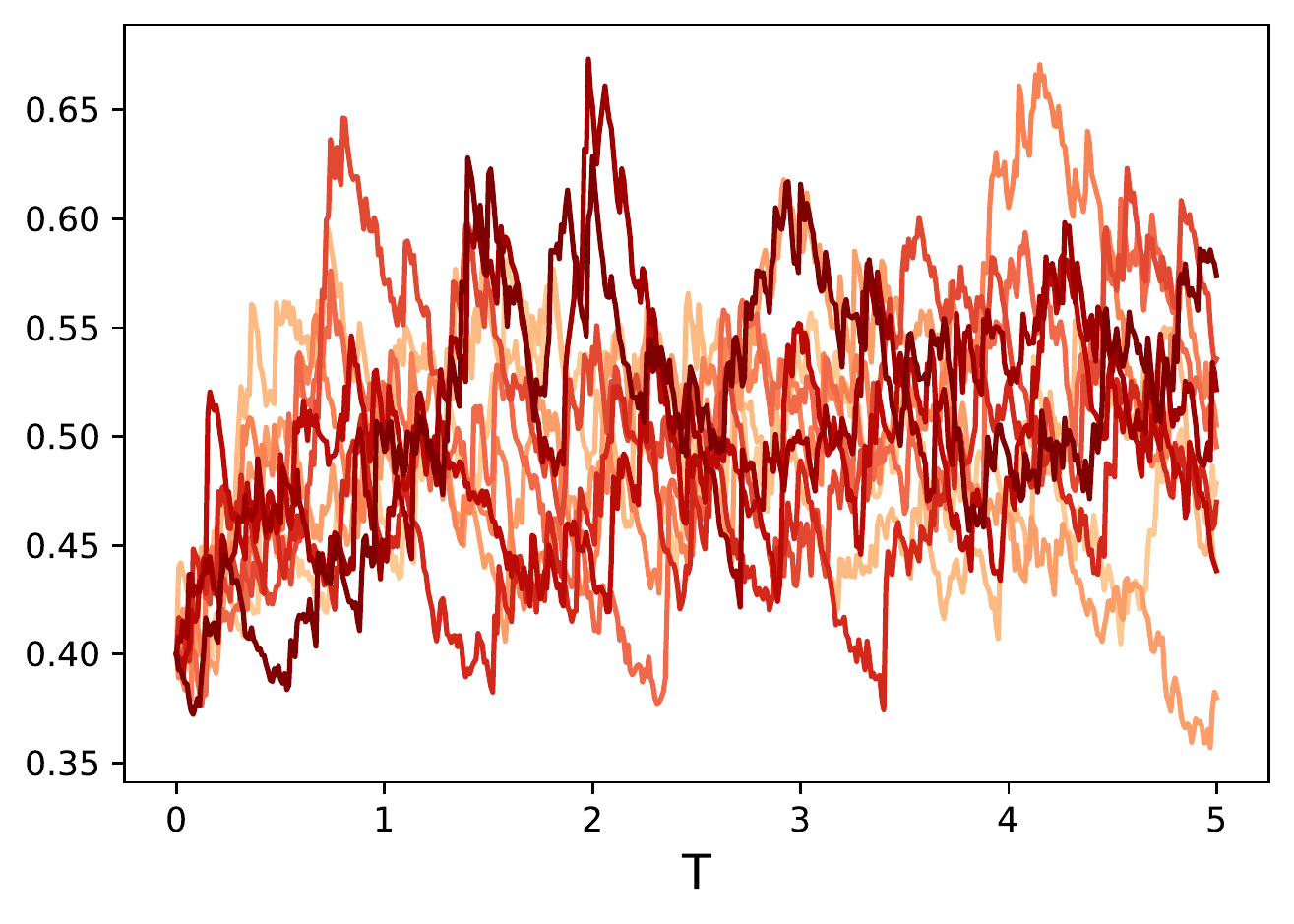}
  \caption{SDE with exponential noise. Left: samples of the training
    data trajectories; Right: Samples of the sFML model simulations
    with an initial condition $x_0=0.4$ for up to $T=5.0$.}
\end{figure}

\begin{figure}[htbp]
  \centering
  \label{fig:Expdis_pdf}
  \includegraphics[width=.43\textwidth]{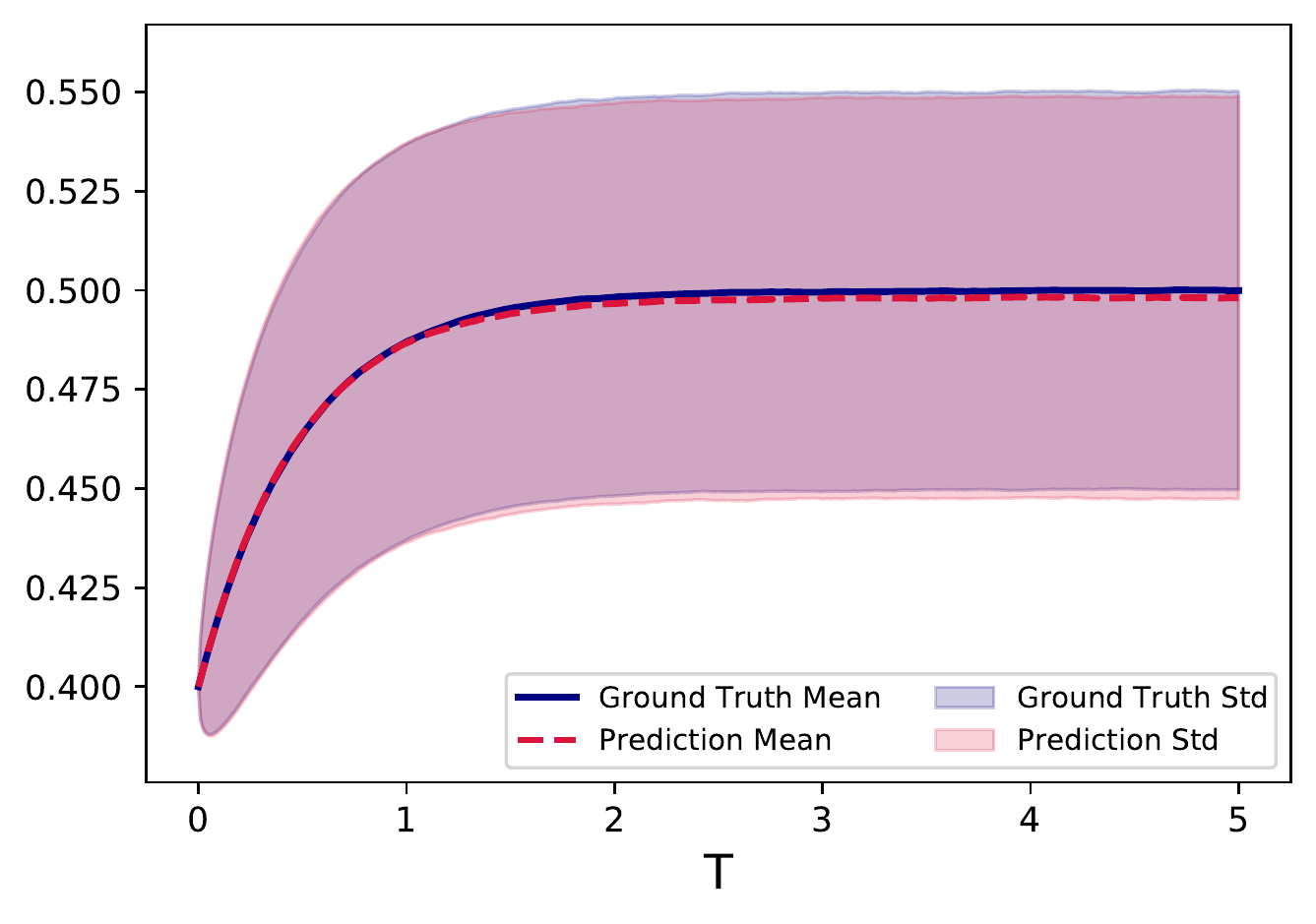}
    \includegraphics[width=.43\textwidth]{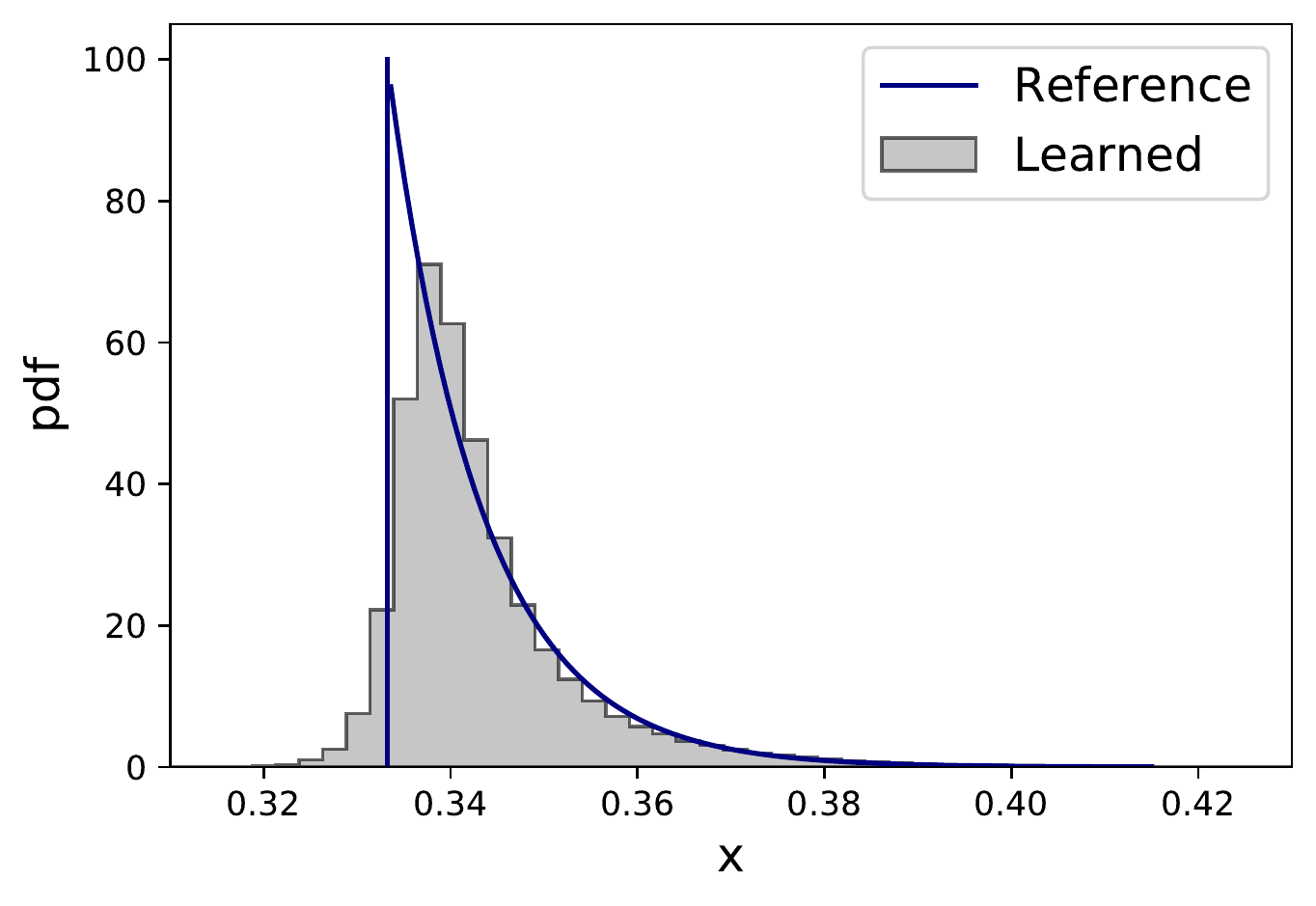}
  \caption{SDE with exponential noise. Left: mean and standard
    deviation of the sFML model predictions; Right: conditional
    distribution by the sFML model $\Gt_\Delta(x)$ against that of the
    true model $\G_\Delta(x)$ at $x=0.34$.}
\end{figure}

We plot a few samples of the training data in the left of Figure \ref{fig:Expdis_data}
up to $T=1.0$.
Some of the simulated trajectory samples by the learned sFML model are shown 
are on the right of Figure \ref{fig:Expdis_data} for up
to $T=5.0$.  The mean and standard deviation of the sFML prediction
are shown on the left of Figure \ref{fig:Expdis_pdf}, where good
agreement with those of the truth can be observed. On the right of
Figure \ref{fig:Expdis_pdf}, we show the conditional distribution by
the sFML model generator $\Gt_\Delta(x)$ at $x=0.34$ (an arbitrary
choice for demonstration). It agrees with the reference solution
$\G_\Delta(0.34)$ generally well, with visible discrepancy at the left
end of the domain. The reference distribution is exponential and has a
clear cut-off, whereas the sFML solution exhibits a small tail. This
is not surprising because the sFML generative model is built upon
stochastic input $\z$ with the standard Gaussian
distribution. Presumably the small discrepancy in the left tail can be
eliminated if one uses exceedingly high order accuracy in the
approximation. This is difficult to achieve with DNN training and we
did not pursue it further.

\begin{figure}[htbp]
  \centering
  \label{fig:Expdis_show}
  \includegraphics[width=.43\textwidth]{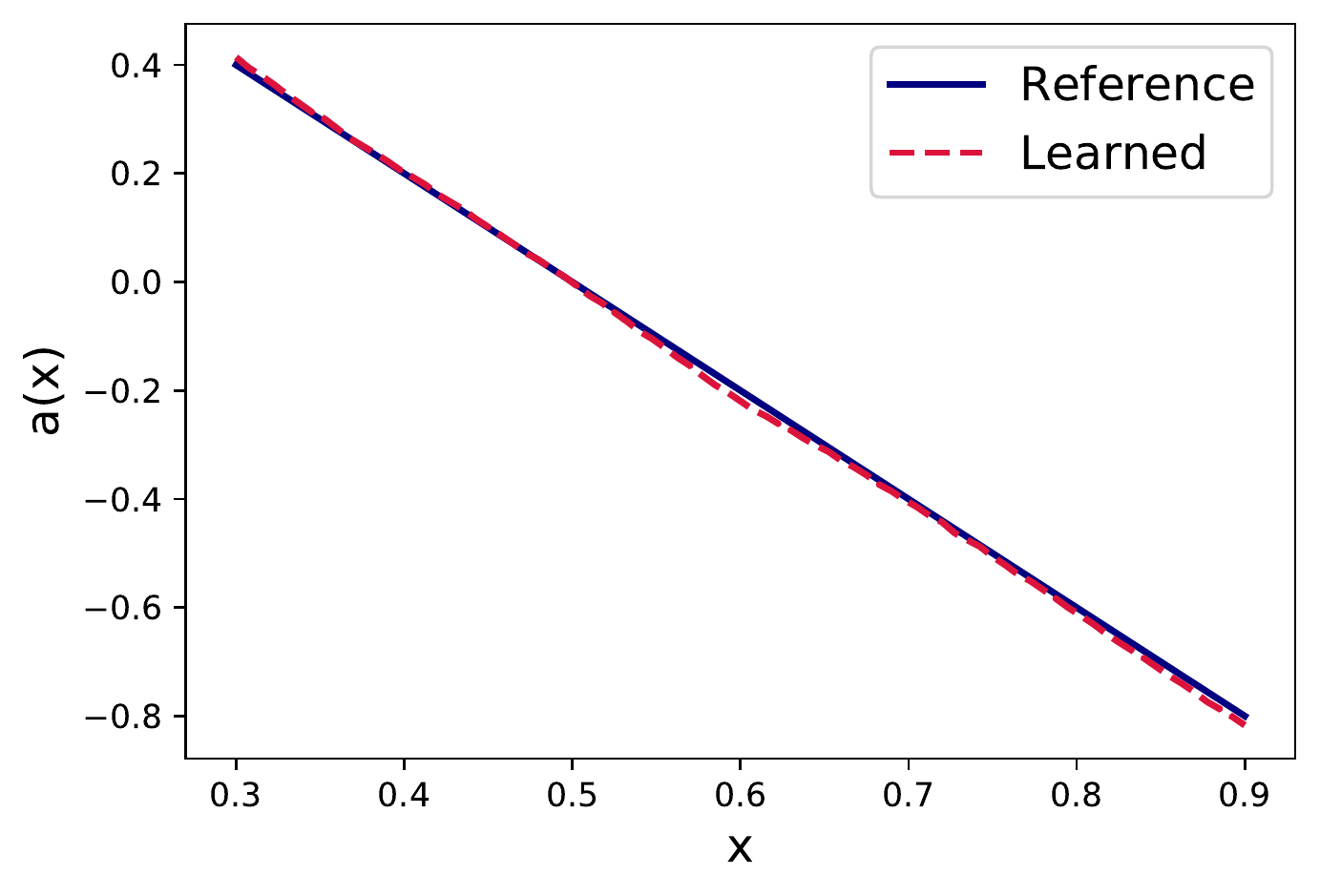}
  \includegraphics[width=.43\textwidth]{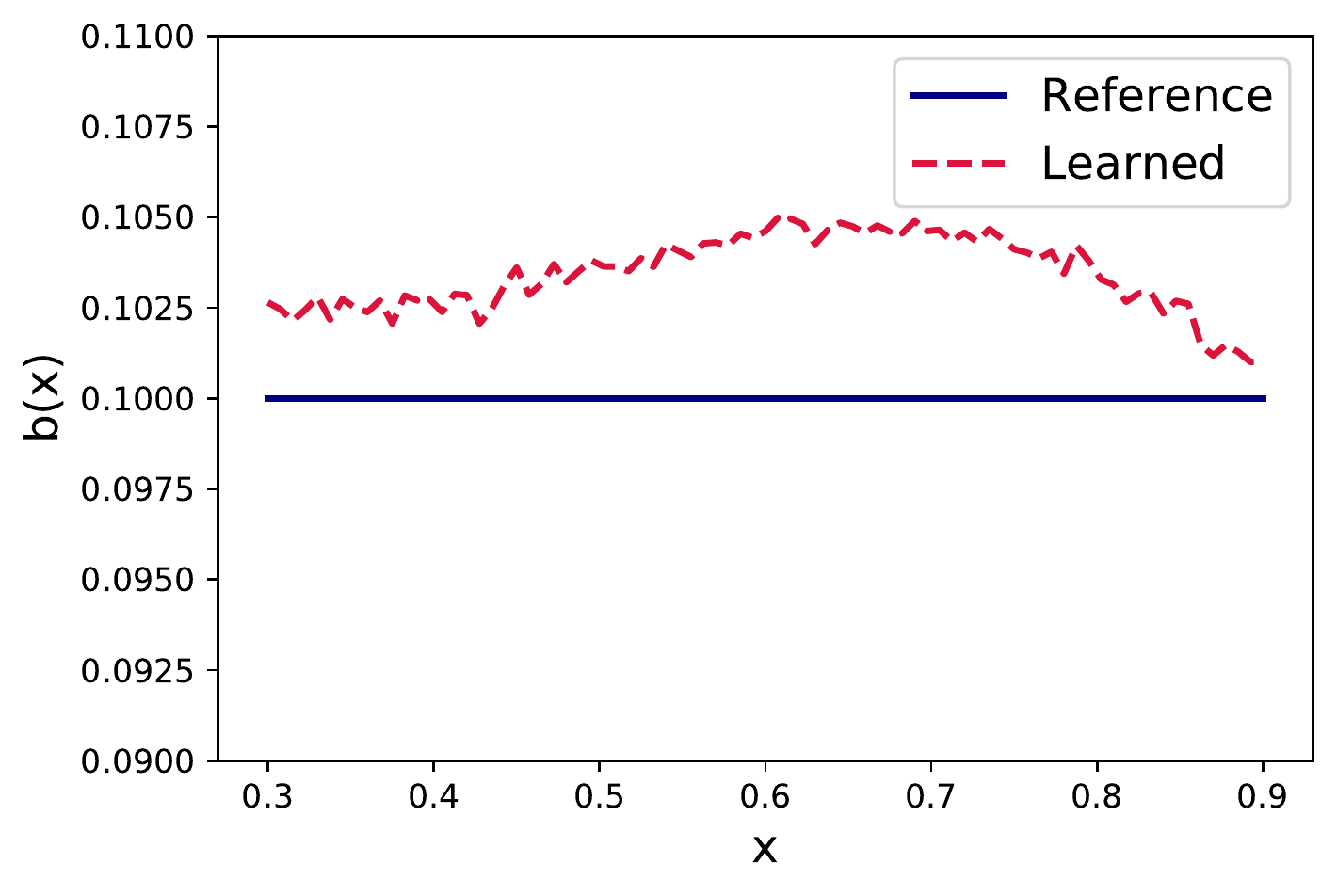}
  \caption{SDE with exponential noise. Left: Recovery of the drift $a(x)=\mu x +
    \sigma/\sqrt{\Delta}$; Right: recovery of the diffusion  $b(x)=\sigma$.}
\end{figure}

Upon moving the non-zero mean of the exponential distribution out of
the noise term, we can write the SDE in the form of the classical SDE \eqref{Ito} and
obtain the drift and diffusion as
\begin{equation}
    \begin{split}
        a(x_n)&=\mathbb{E}_\omega\left(\frac{x_{n+1}-x_{n}}{\Delta}\Big|x_n\right)=\mu x_n + \frac{\sigma}{\sqrt{\Delta}},\\
        b(x_n)&=\text{Std}_\omega\left(\frac{x_{n+1}-x_{n}}{\Delta}|x_n\right)=\sigma.
    \end{split}
\end{equation}
The effective drift and diffusion recovered by the sFML model are
shown in Figure \ref{fig:Expdis_show}, where we observe very good
agreement with the reference truth.

\subsubsection{Noise with Lognormal Distribution}
We now consider the following stochastic system
\begin{equation}
    d \log x_t = (\log m -\theta \log x_t) d t+\sigma d W_{t},
\end{equation}
where $m$, $\theta$ and $\sigma$ are parameters. This is effectively
an OU process after taking an exponential operation.
Its dynamics can
be solved by using Euler method with the following scheme:
\begin{equation}
    x_{n+1}= m^{\Delta} x_n^{1-\theta \Delta} \eta_n^{\sigma
      \sqrt{\Delta}}, \qquad \eta_n \sim \text{Lognormal}(0,1).
\end{equation}

\begin{figure}[htbp]
  \centering
  \label{fig:ExpOU_data}
  \includegraphics[width=.43\textwidth]{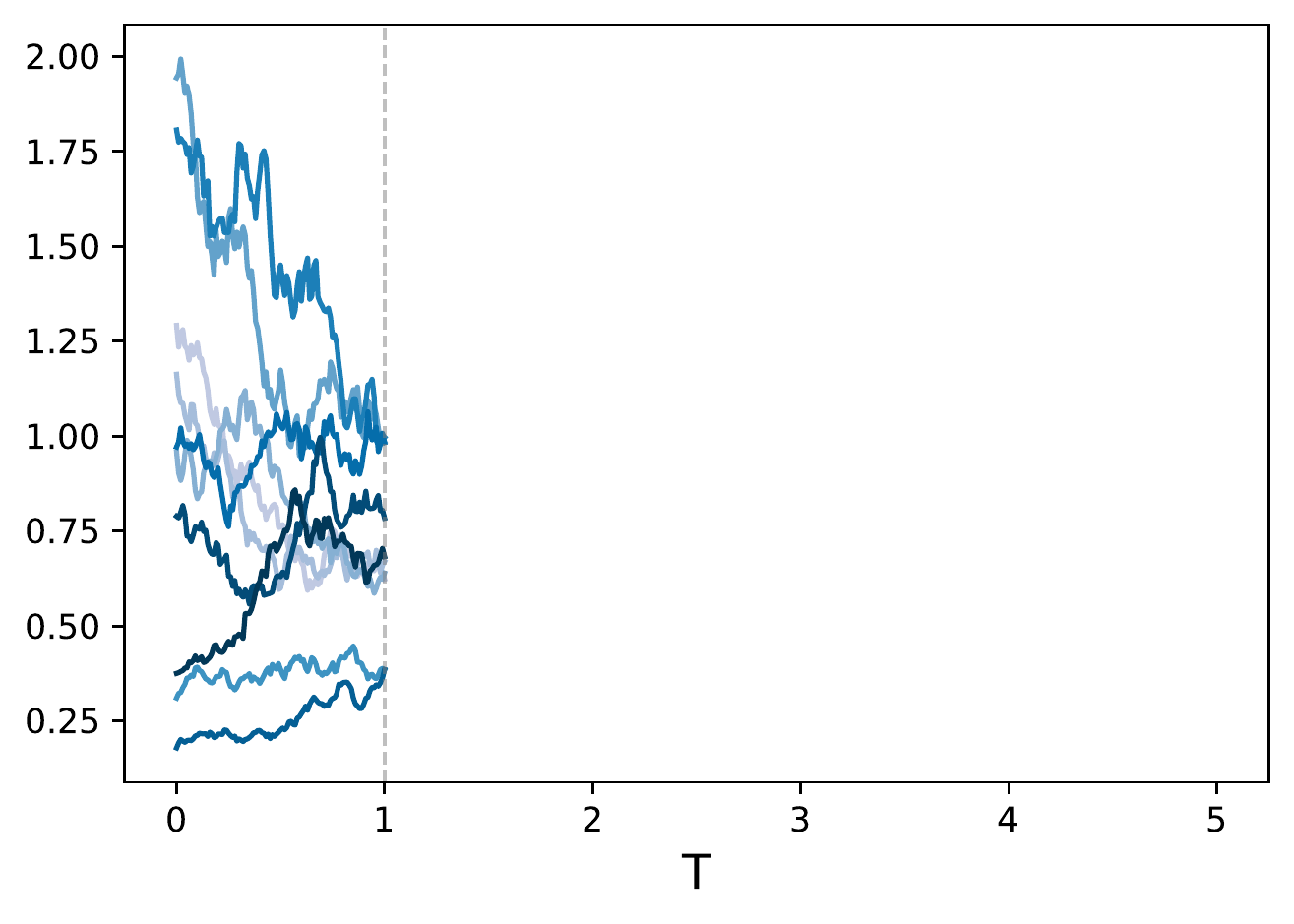}
  \includegraphics[width=.43\textwidth]{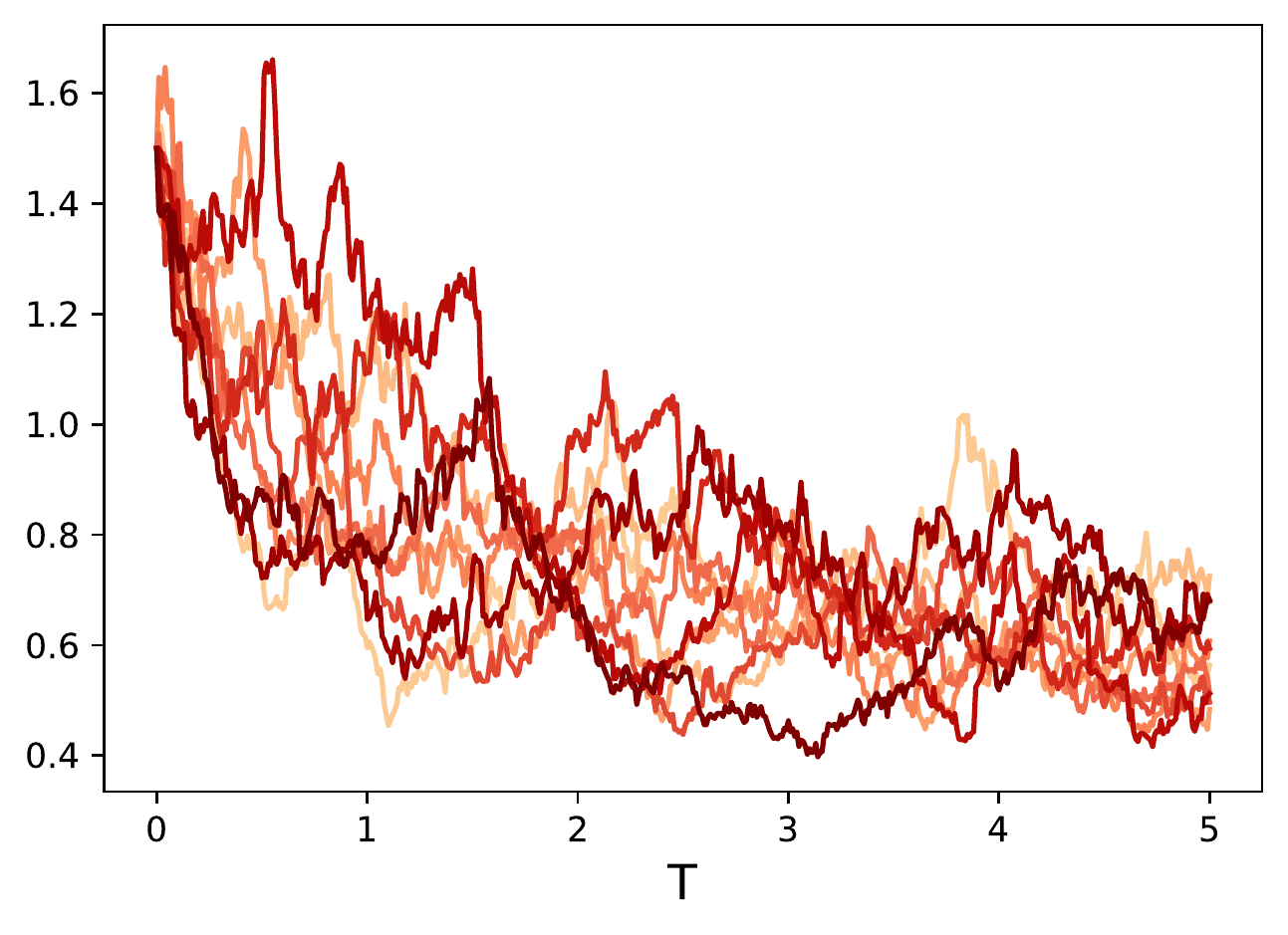}
  \caption{SDE with lognormal distribution. Left: Samples of the
    training trajectory data; Right: Simulation sample of the trained
    sFML model with an initial condition $x_0=1.5$.}
\end{figure}

In our test, we take $m=1/\sqrt{e}$, $\theta=1.0$ and $\sigma=0.3$.
It is obvious that the conditional distribution from the stochastic flow map $\G_{\Delta}(x)$ follows
lognormal distribution for any $x$.
The  training data are generated with initial condition uniformly
sampled from $\mathcal{U}(0.1,2.0)$ up to time $T=1.0$. See the left
of Figure \ref{fig:ExpOU_data} for some samples.
Upon training the sFML model, we conduct system prediction for time up
to $T=5.0$. Some simulation samples with an initial condition
$x_0=1.5$ are shown on the right of Figure \ref{fig:ExpOU_data}.
The mean and standard deviation from the sFML model prediction, as
well as the distribution of $\Gt_\Delta$, are shown in Figure
\ref{fig:ExpOU_data}. Good agreement with the truth reference
solutions can be observed.
%
\begin{figure}[htbp]
  \centering
  \label{fig:ExpOU_pdf}
  \includegraphics[width=.43\textwidth]{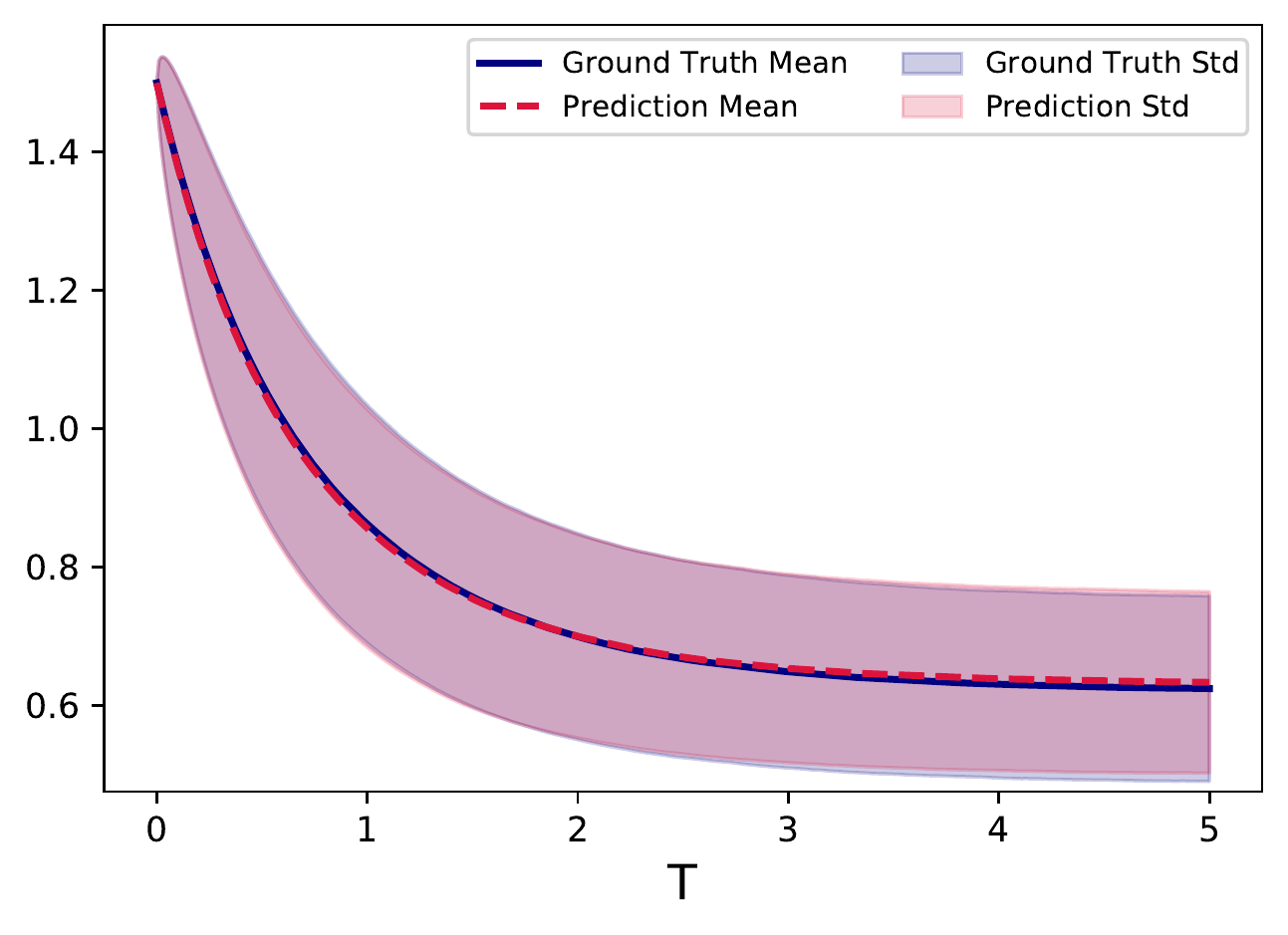}
  \includegraphics[width=.43\textwidth]{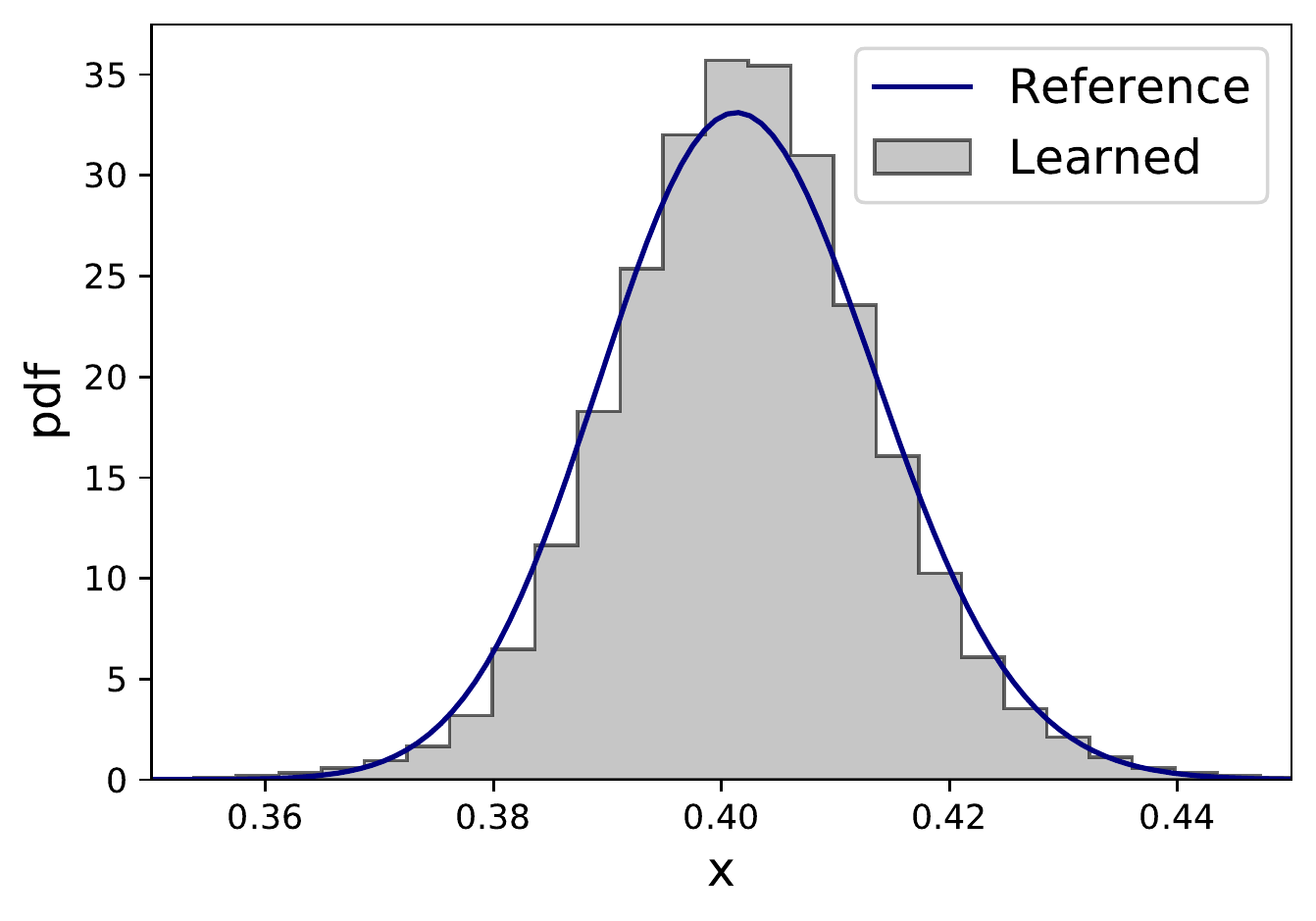}
  \caption{SDE with lognormal distribution. Left: mean and standard
    deviation of the sFML model prediction with the initial condition
    $x_0=1.5$; Right: conditional probability by the sFML generator
    $\Gt_\Delta(x)$ at $x=0.4$.}
\end{figure}

\begin{figure}[htbp]
  \centering
  \label{fig:ExpOU_show}
  \includegraphics[width=.43\textwidth]{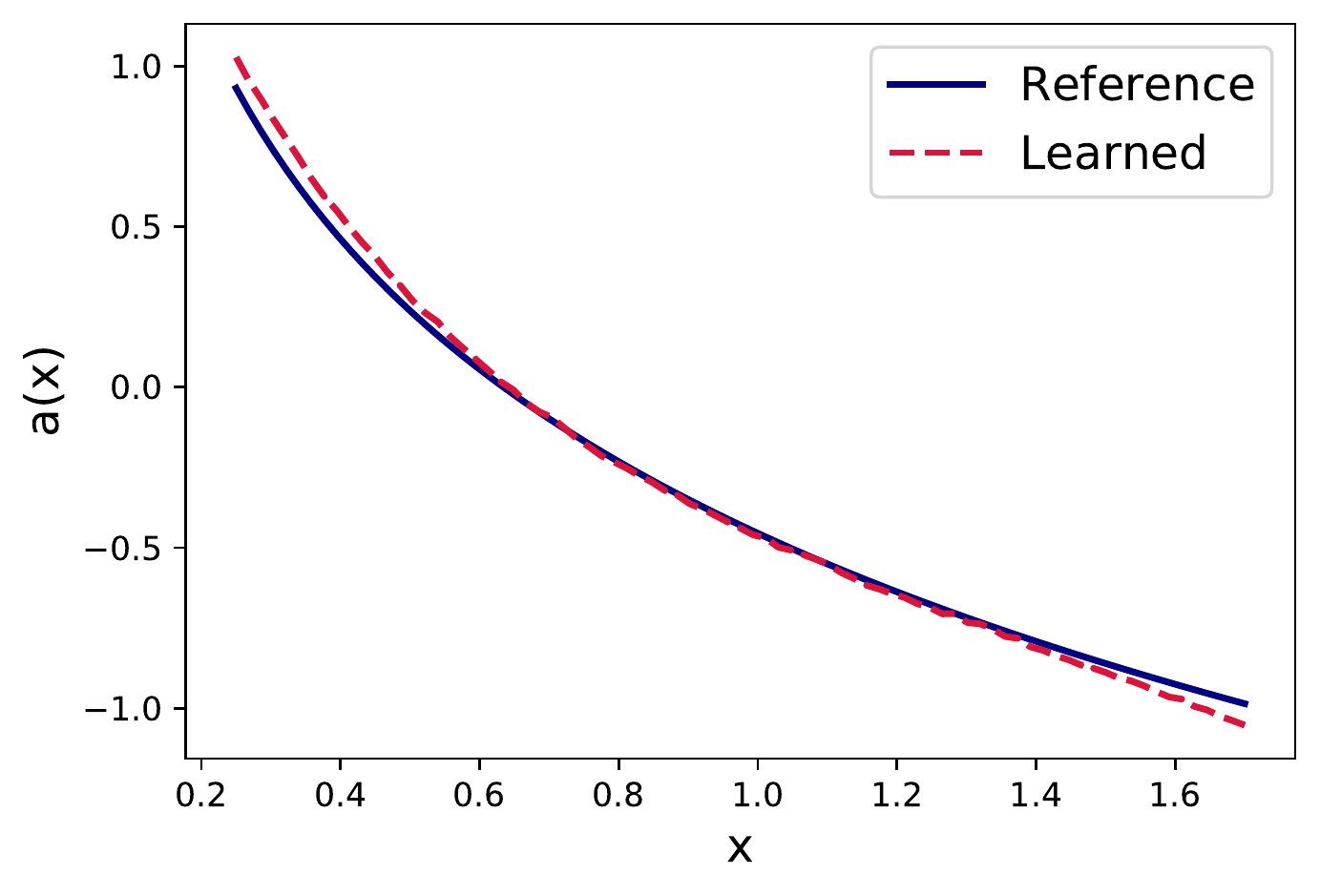}
  \includegraphics[width=.43\textwidth]{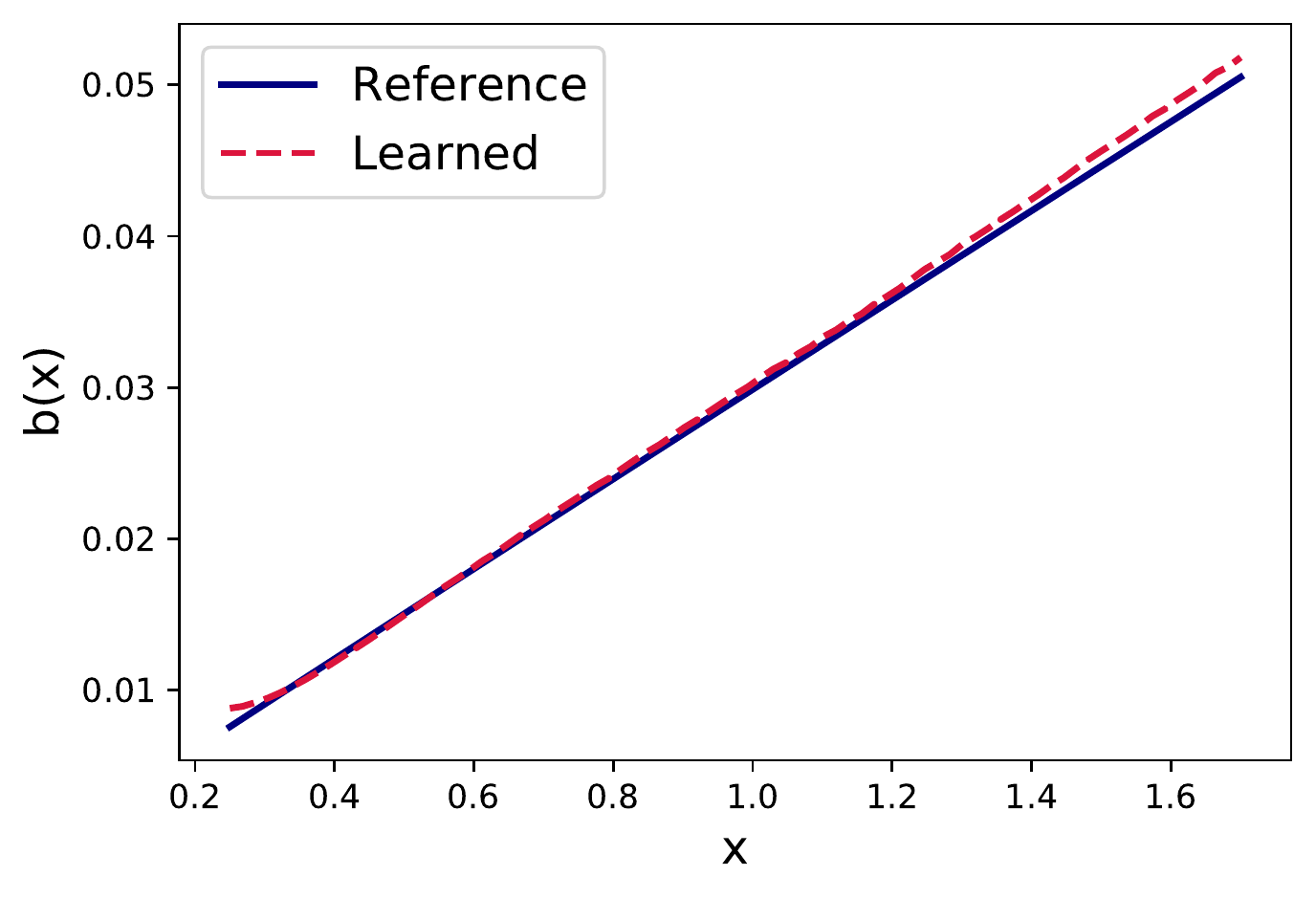}
  \caption{SDE with lognormal distribution. Left: recovery of the
    drift $a(x)$; Right: recovery of conditional standard deviation
    $b(x)$; (See \eqref{ab_lognormal} for the reference solution.)}
\end{figure}

Like in the previous example, we rewrite this SDE in the form of the
classical SDE \eqref{Ito} and obtain
\begin{equation} \label{ab_lognormal}
    \begin{split}
        a(x_n)&=\ln\left[\left(\mathbb{E}_\omega\left(\frac{x_{n+1}}{x_n}\Big|x_n\right)\right)^{1/\Delta}\right]=\ln(mx_n^{-\theta})+\frac{\sigma^2}{2},\\
        b(x_n)&=\text{Std}_\omega\left(x_{n+1}|x_n\right)=\sqrt{e^{\sigma^2\Delta}-1}(me^{\sigma^2/2})^\Delta (1-\theta\Delta)x_n.
    \end{split}
\end{equation}
From the learned sFML model, we estimate the effective drift and
diffusion via
\begin{equation}
    \hat{a}(x) = \ln
    \left[\left(\mathbb{E}_{\omega}\left(\frac{\Gt_\Delta(x,\omega)}{x}\right)\right)^{1/\Delta}\right],
    \qquad \hat{b}(x) = \text{Std}_{\omega}(\Gt_\Delta(x,\omega)).
\end{equation}
The comparison of functions for $a(x)$ and $b(x)$ are shown in Figure
\ref{fig:ExpOU_show}, where we observe good agreement between
the recovered functions by the learned sFML model and the reference \eqref{ab_lognormal}.

\subsection{Two-Dimensional SDEs}
In this section, we present examples of learning two-dimensional SDE systems.

\subsubsection{Two-dimensional Ornstein–Uhlenbeck process}
\begin{figure}[htbp]
  \centering
  \label{fig:MdOU_data}
  \includegraphics[width=.43\textwidth]{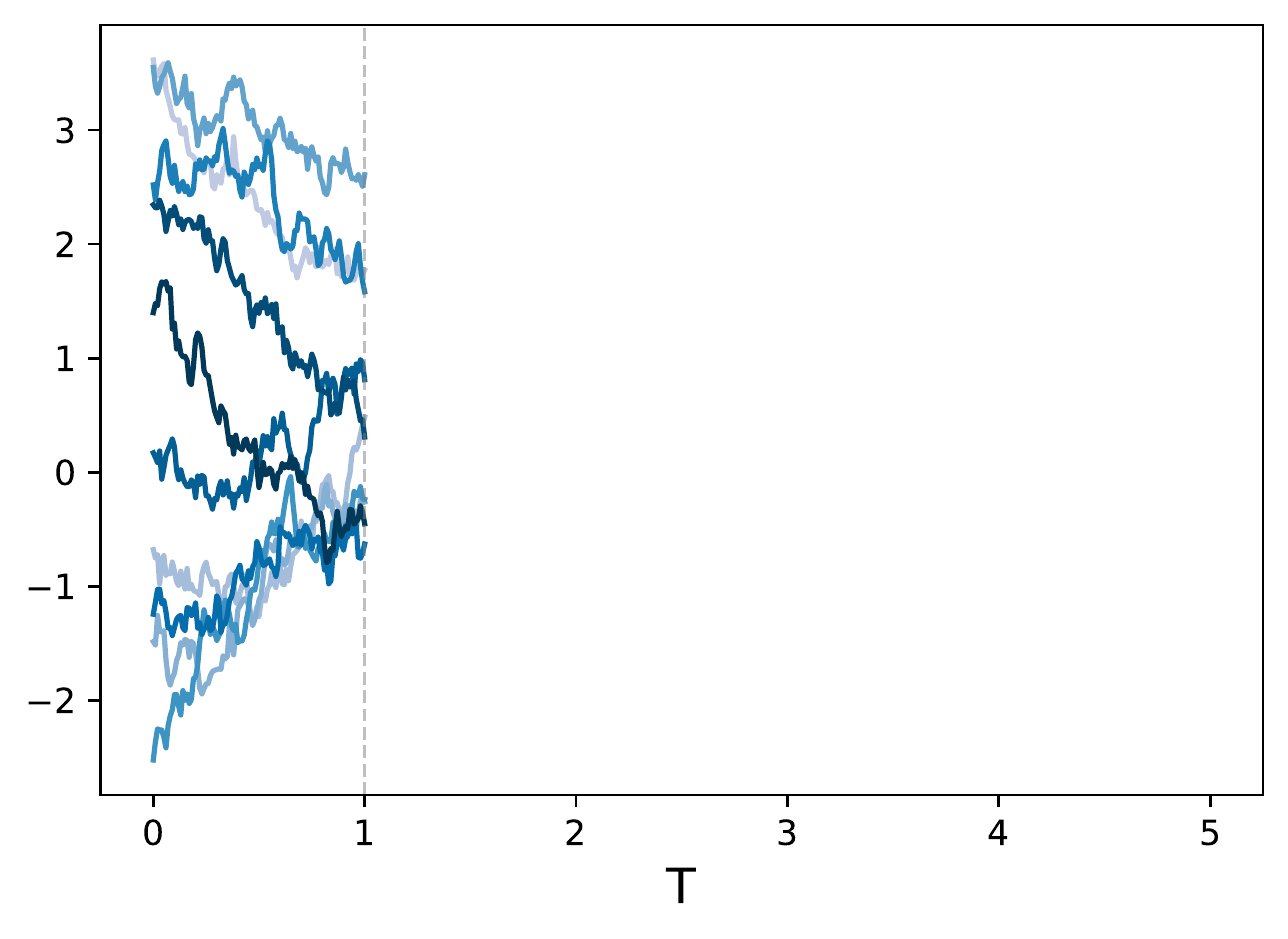}
  \includegraphics[width=.43\textwidth]{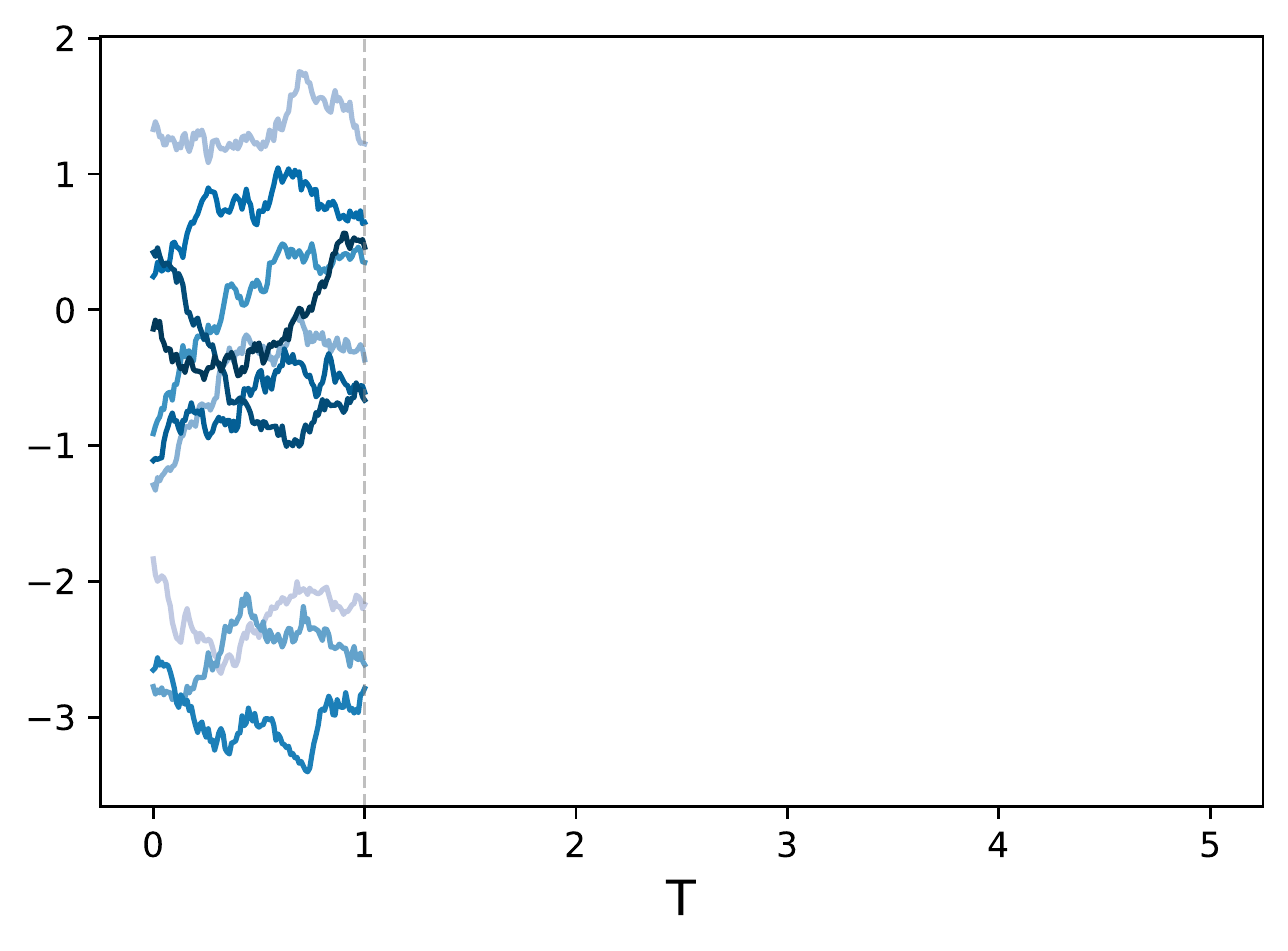}\\
  \includegraphics[width=.43\textwidth]{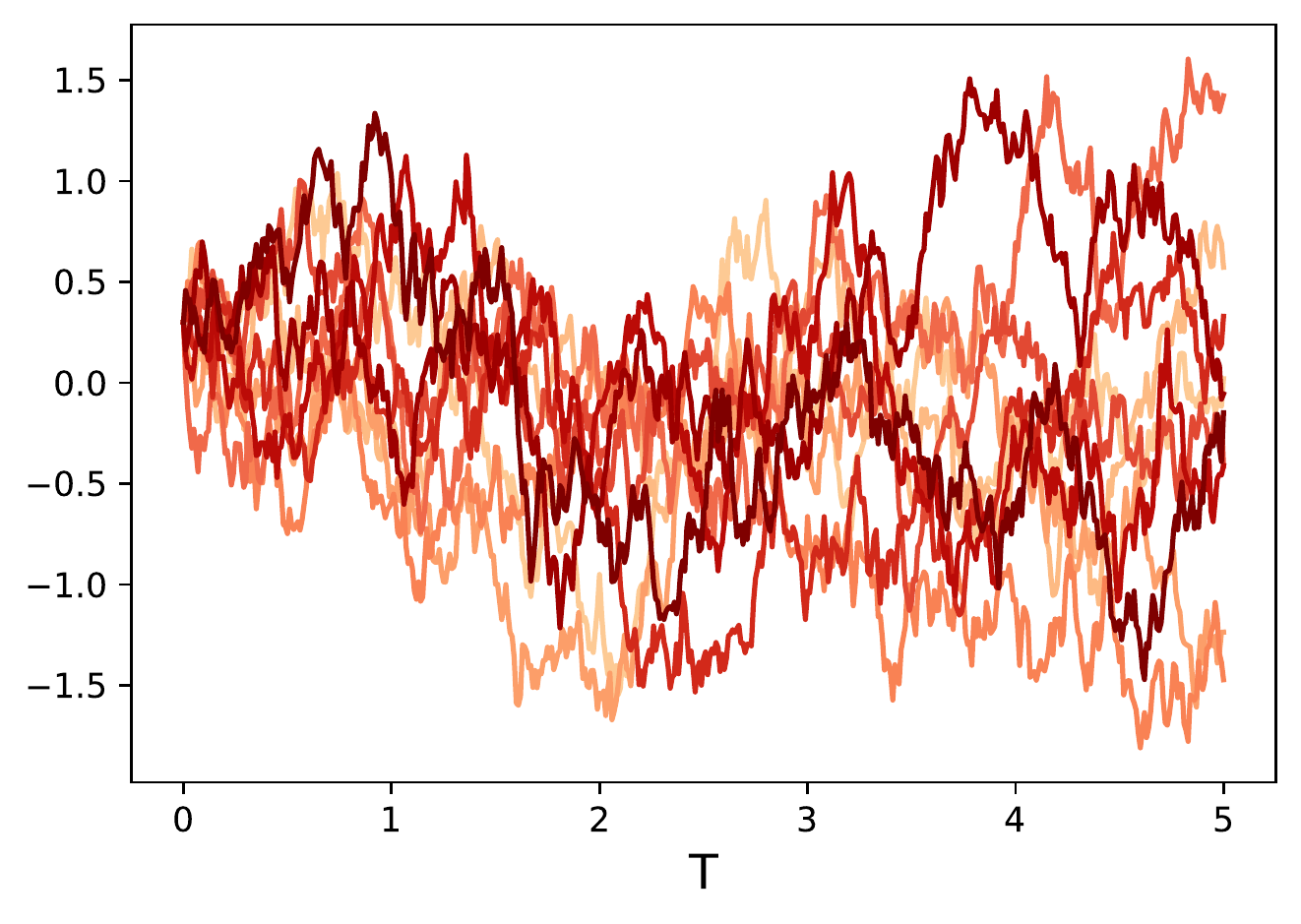}
  \includegraphics[width=.43\textwidth]{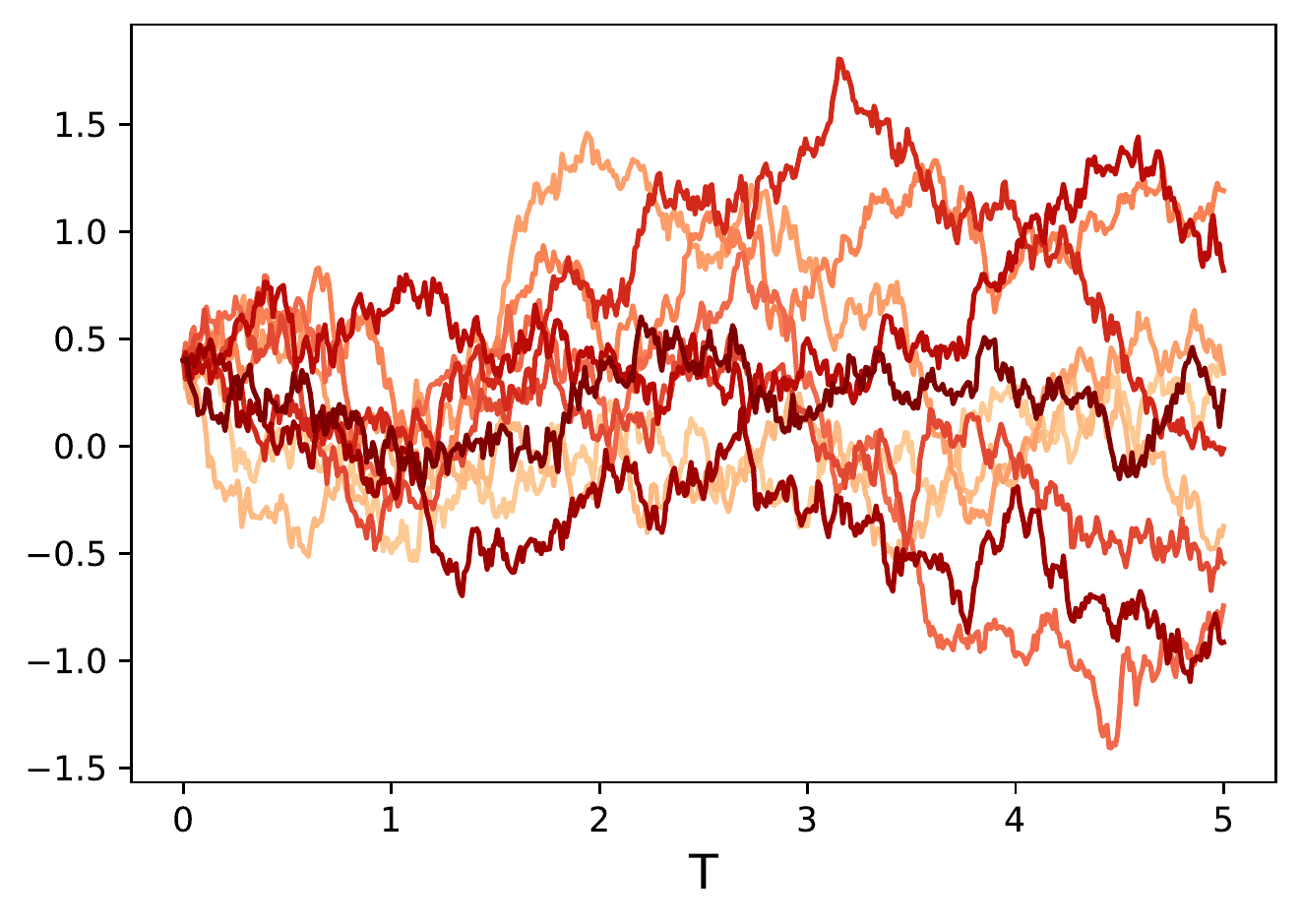}
  \caption{Learning the 2d OU process \eqref{OU2d}. Top row: samples of
    the training data for $x_1$ (left) and $x_2$ (right);
    Bottom row: The sFML model prediction trajectories for $x_1$ (left)
 and  $x_2$ (right), with an initial condition $\x_0=(0.3,0.4)$.}
\end{figure}

\begin{figure}[htbp]
  \centering
  \label{fig:MdOU_show1}
  \includegraphics[width=.43\textwidth]{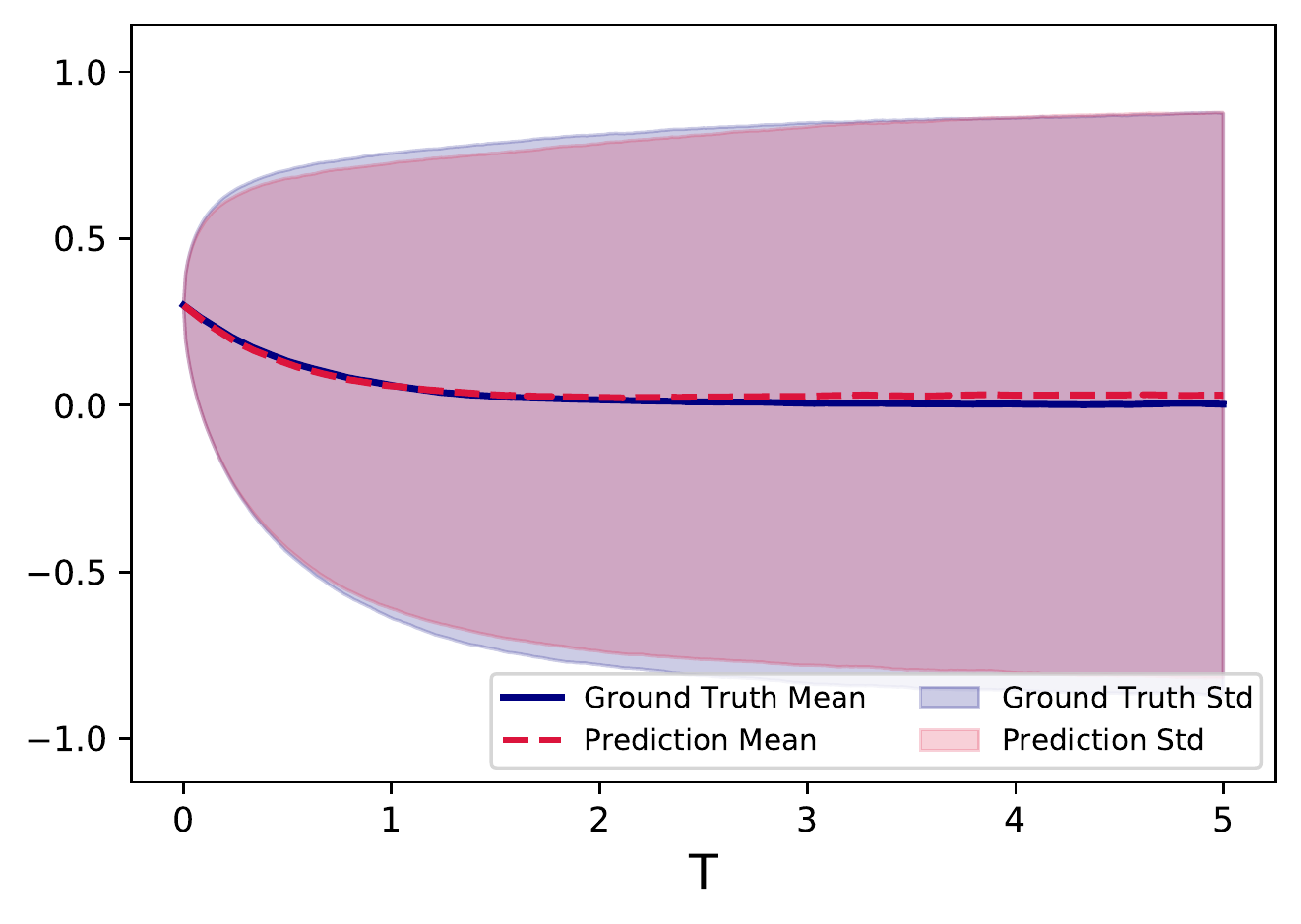}
  \includegraphics[width=.43\textwidth]{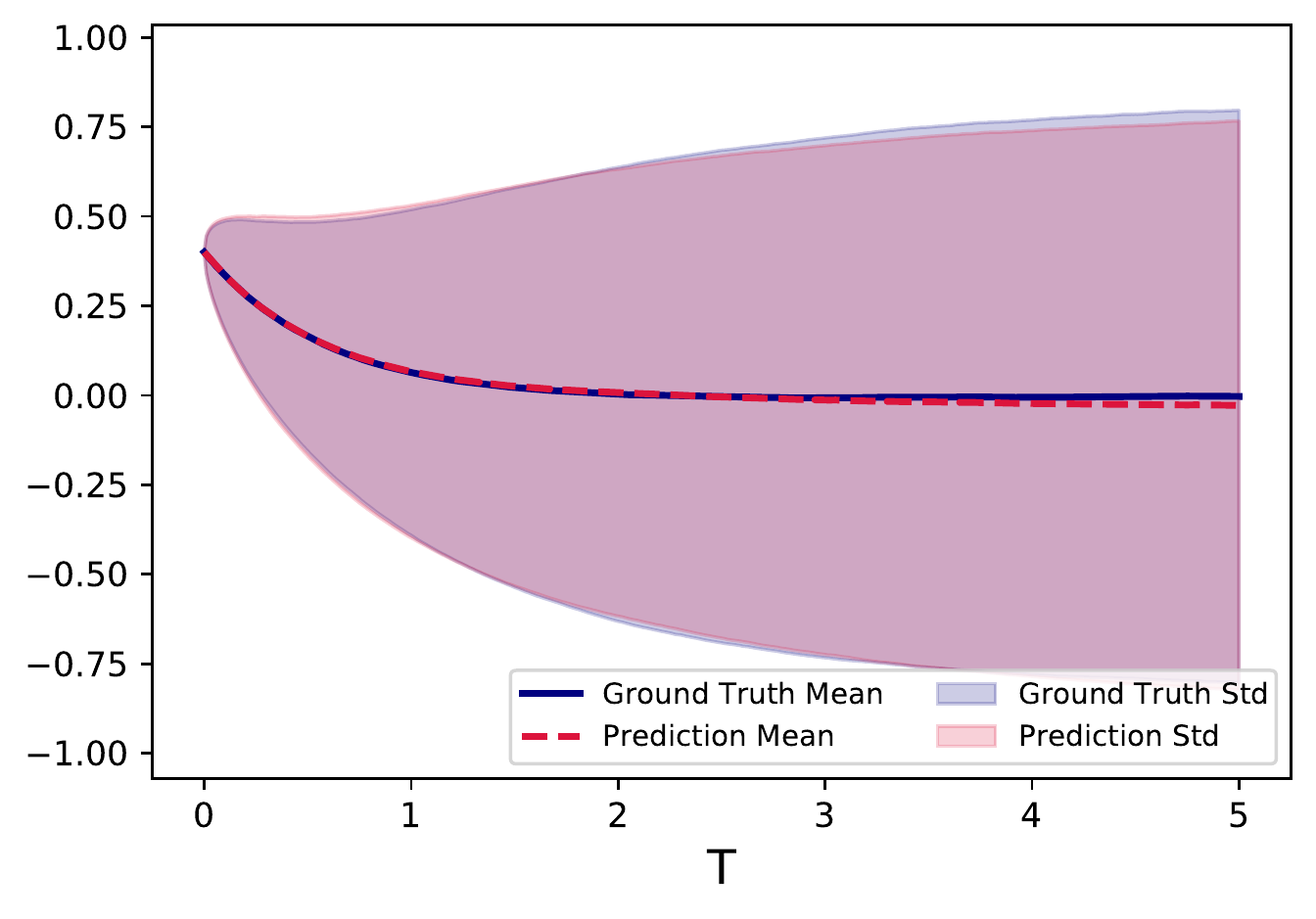}
  \caption{Solution mean and standard deviation of the sFML model
    prediction for the 2d OU process \eqref{OU2d}. Left: $x_1$; Right: $x_2$.}
\end{figure}

\begin{figure}[htbp]
  \centering
  \label{fig:MdOU_dis}
   \includegraphics[width=.43\textwidth]{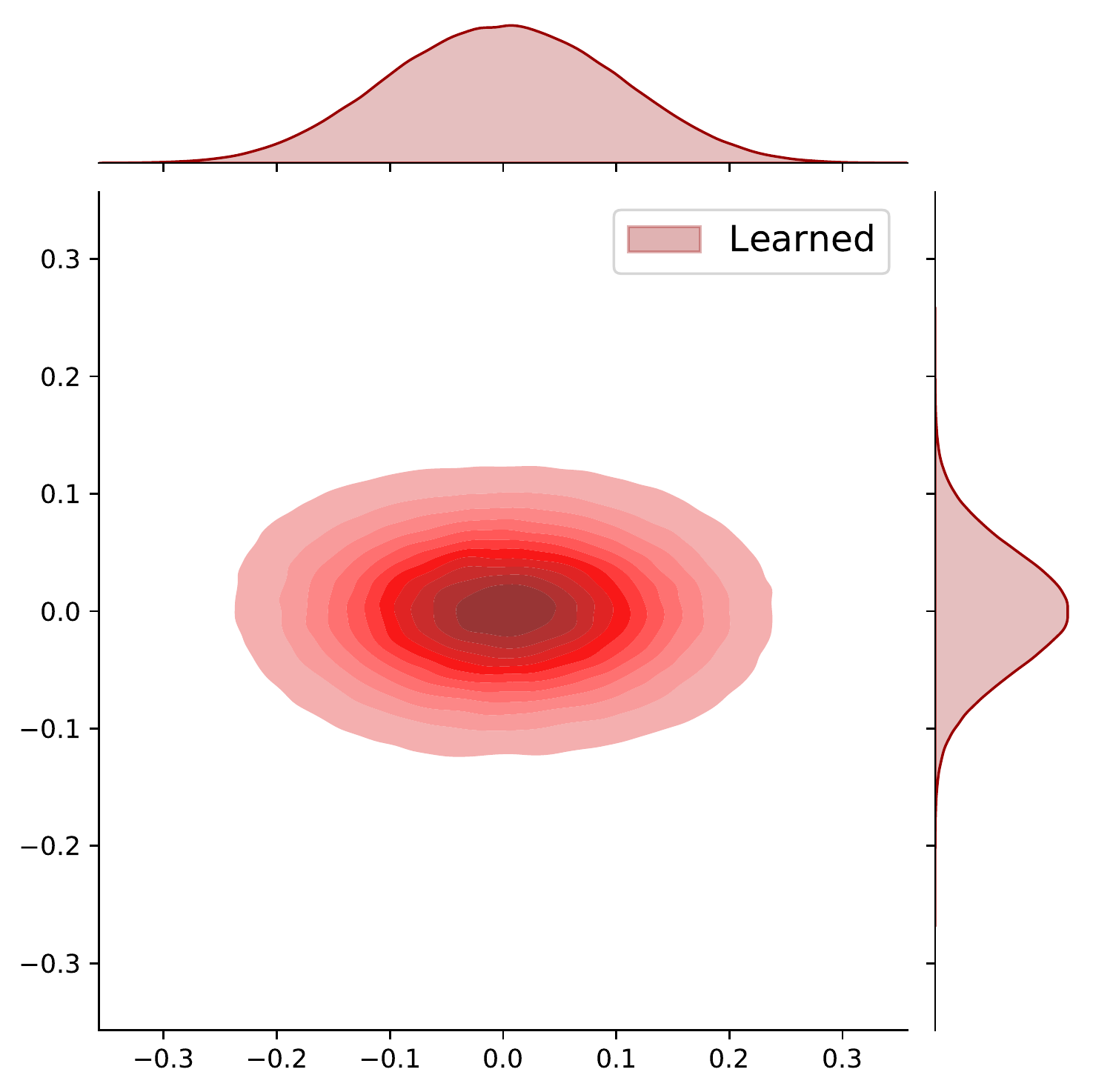}
 \includegraphics[width=.43\textwidth]{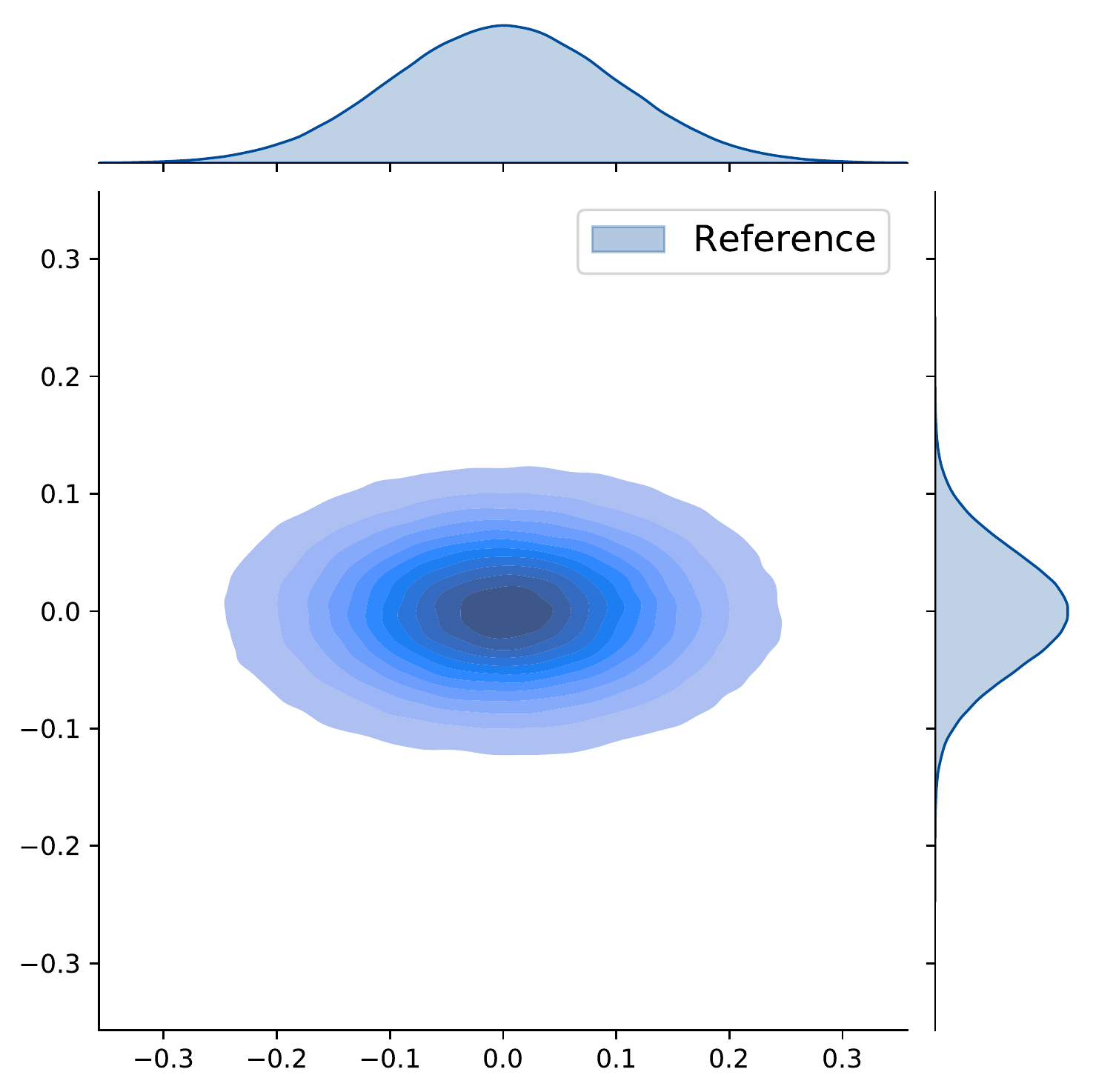}
  \caption{Learning the 2d OU process \eqref{OU2d}.: conditional probability distribution from the learned sFML
    generator $\Gt_\Delta(\x)$ (left) and from the true generator
    $\G_\Delta(\x)$ (right) at $\x=(0,0)$.}
\end{figure}

We first consider a 2-dimensional OU process,
\be \label{OU2d}
d\mathbf{x}_t = \mathbf{B}\mathbf{x}_t dt + \mathbf{\Sigma} d\mathbf{W}_t,
\ee
where $\mathbf{x}_t = (x_1,x_2) \in \Rs^2$ are the state variables,
$\mathbf{B}$ and $\mathbf{\Sigma}$ are ${(2\times 2)}$ are matrices.
In our test, we set
$$\mathbf{B}=\left(\begin{array}{cc} -1 & -0.5\\ -1 &
  -1 \end{array}\right) \qquad
\mathbf{\Sigma}=\left(\begin{array}{cc} 1 & 0\\ 0 & 0.5 \end{array}\right).
$$
Our training data are generated with random initial conditions
uniformly sampled from $\mathcal{U}([-4,4]\times [-3,3])$, for a
termination time up to $T=1.0$.
In the top row of Figure \ref{fig:MdOU_data}, a few sample training
data trajectories are shown. 
Upon training the sFML model, we conduct system predictions for time
up to $T=5$.
In the bottom row of Figure \ref{fig:MdOU_data}, we show a few
prediction trajectories by the trained sFML model, with an initial
condition of $\x_0=(0.3,0.4)$.
The mean and the standard deviation of the model prediction are shown
Figure \ref{fig:MdOU_show1}, where we observe very good agreement with
the reference solutions.
To examine the conditional probability distribution generated by the
learned sFML model, we present both the joint probability distribution
and its marginal distributions by the sFML generator $\Gt_\Delta(\x)$
at $\x=(0,0)$, along with the true distribution $\G_\Delta$ for
comparison. The results are in Figure \ref{fig:MdOU_dis}, where good
agreement with the true conditional distribution can be seen.
%
%
%
%

\subsubsection{Stochastic Oscillator}
\begin{figure}[htbp]
  \centering
  \label{fig:SO_data}
  \includegraphics[width=.43\textwidth]{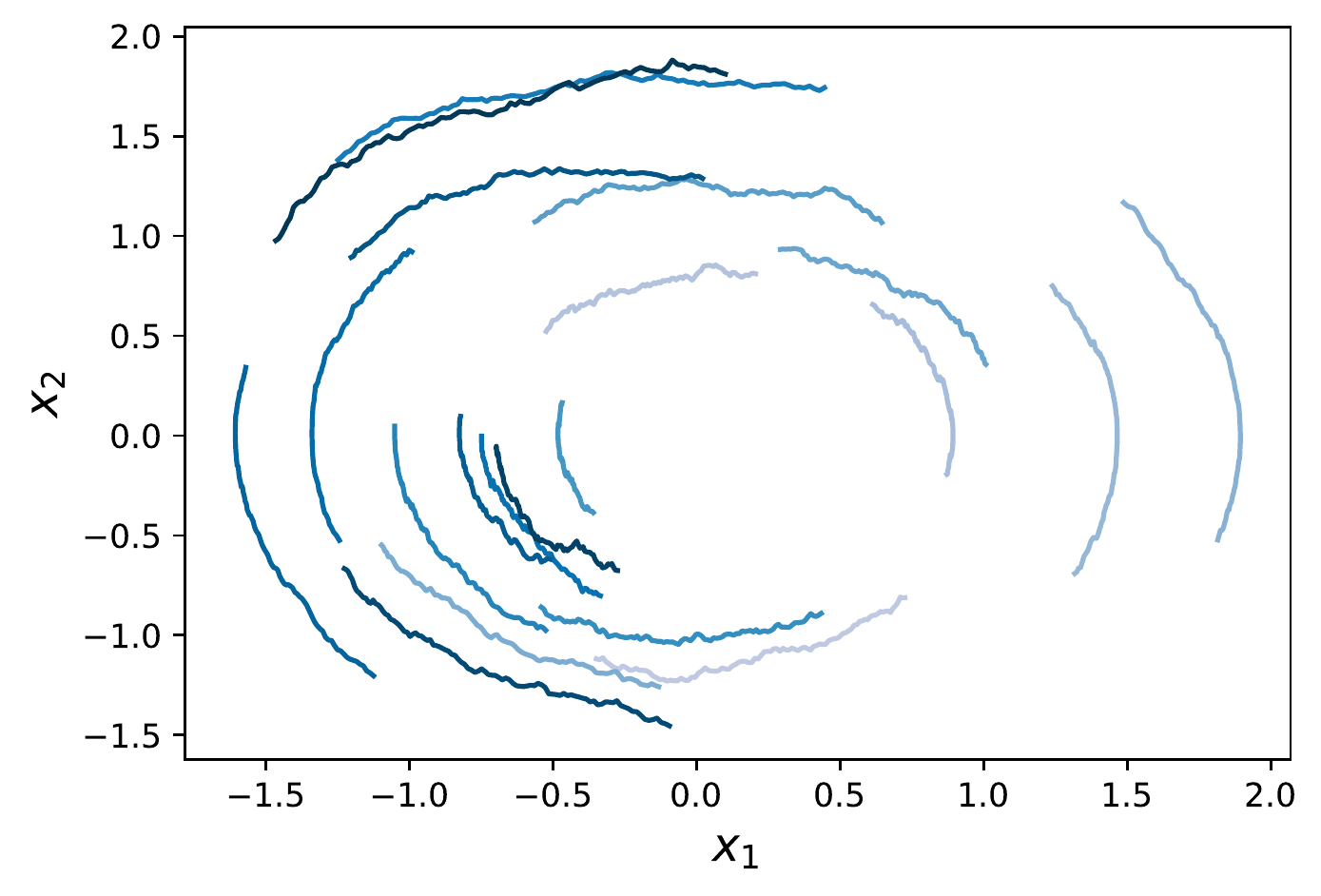}
  \includegraphics[width=.43\textwidth]{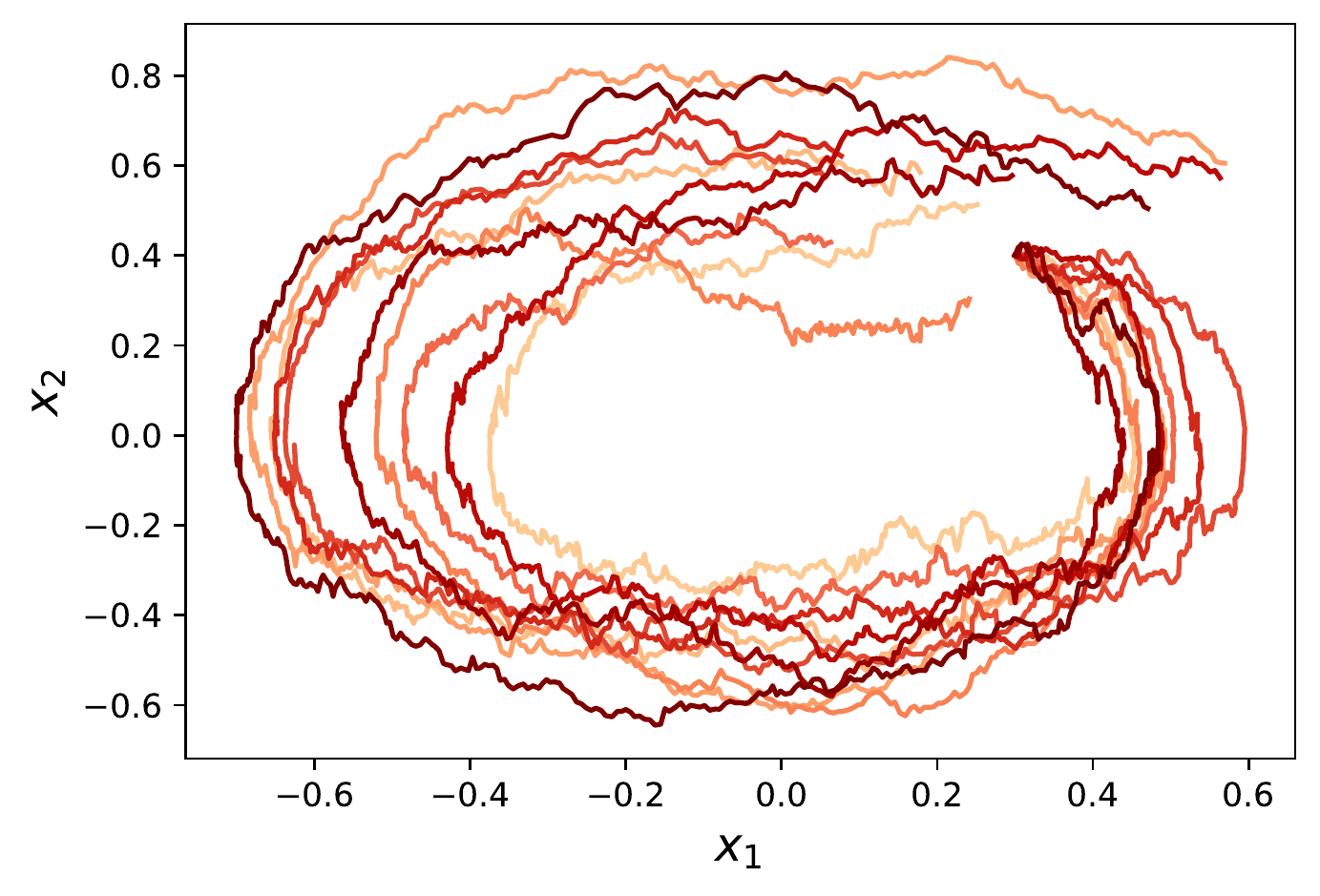}
  \caption{Stochastic oscillator \eqref{oscillator}. Left: samples of the phase portrait
    of the training trajectory data (up to $T=1.0$); Right:  phase
    portait of samples of the sFML prediction solution for up to
    $T=6.5$, with the initial condition $\x_0=(0.3,0.4)$.}
\end{figure}

In this example, we consider the following stochastic oscillator, 
\begin{equation} \label{oscillator}
\left\{\begin{array}{l}
\dot{x}_1 = x_2, \\
\dot{x}_2 = -x_1+\sigma \dot{W}_t.
\end{array}\right.
\end{equation}
Here $\sigma$ is the perturbation parameter and set as $\sigma=0.1$ in
our test. Our training data are generated with initial conditions uniformly
sampled from $\mathcal{U}([-1.5,1.5]\times [-1.5,1.5])$, for time up
to $T=1.0$. Some of the training data trajectories are shown on the
left of Figure  \ref{fig:SO_data}, for visualization purpose.
For the sFML model training, we use 100 recurrent steps for 5,000
epochs for the deterministic sub-map $\Dt_\Delta$. For the stochastic
sub-map, we use GANs the 4-layer, each of which with 80 nodes.
Once trained, we conduct system prediction using the learned sFML
model for time up to $T=6.5$. Some sample trajectories are shown on
the right of Figure \ref{fig:SO_data}, for an initial condition of
$\x_0=(0.3, 0.4)$.

\begin{figure}[htbp]
  \centering
  \label{fig:SO_show}
  \includegraphics[width=.43\textwidth]{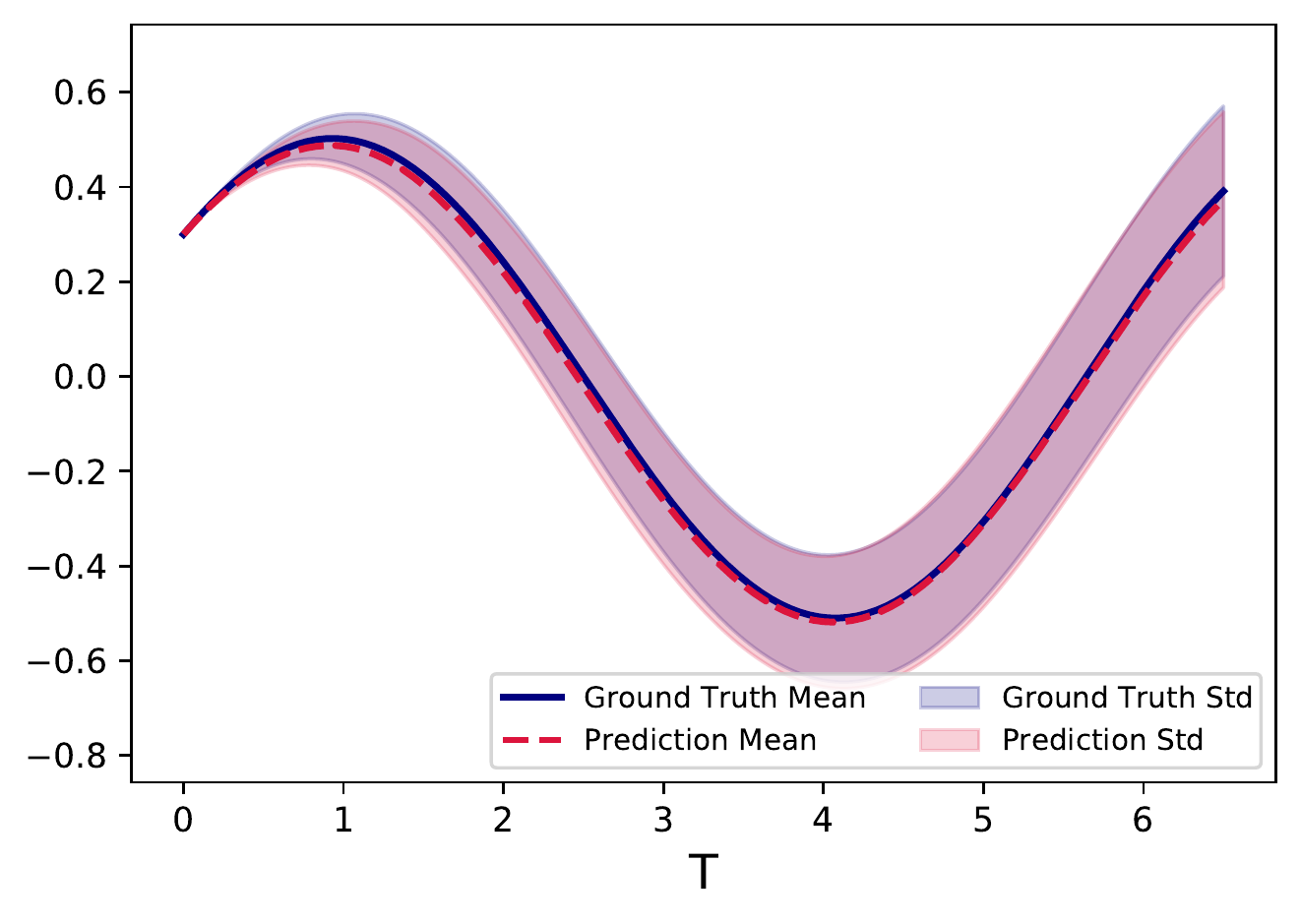}
  \includegraphics[width=.43\textwidth]{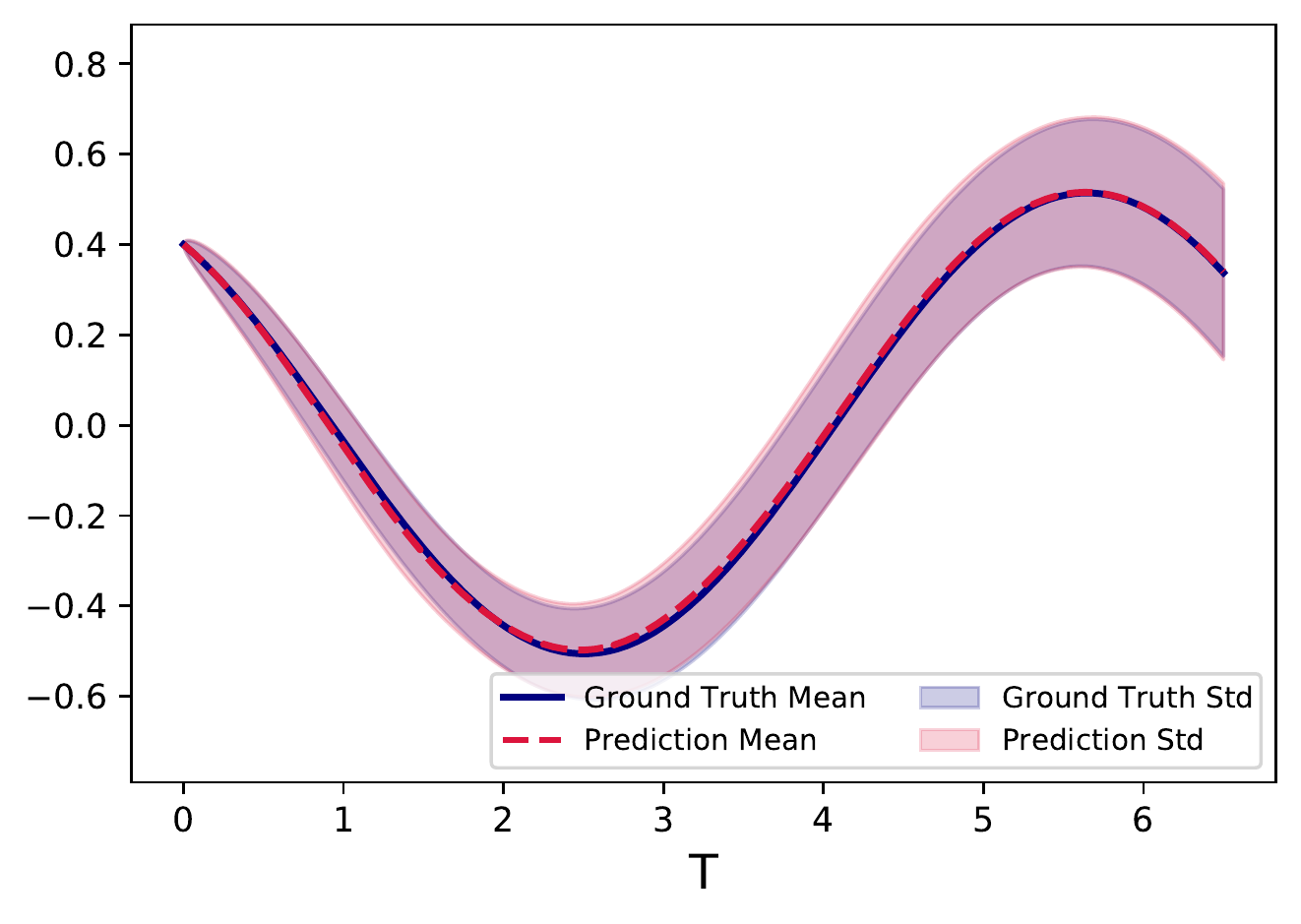}
  \caption{Mean and standard deviation of the stochastic oscillator
    \eqref{oscillator}. Left: $x_1$; Right: $x_2$.}
\end{figure}

\begin{figure}[htbp]
  \centering
  \label{fig:SO_show2}
  \includegraphics[width=.9\textwidth]{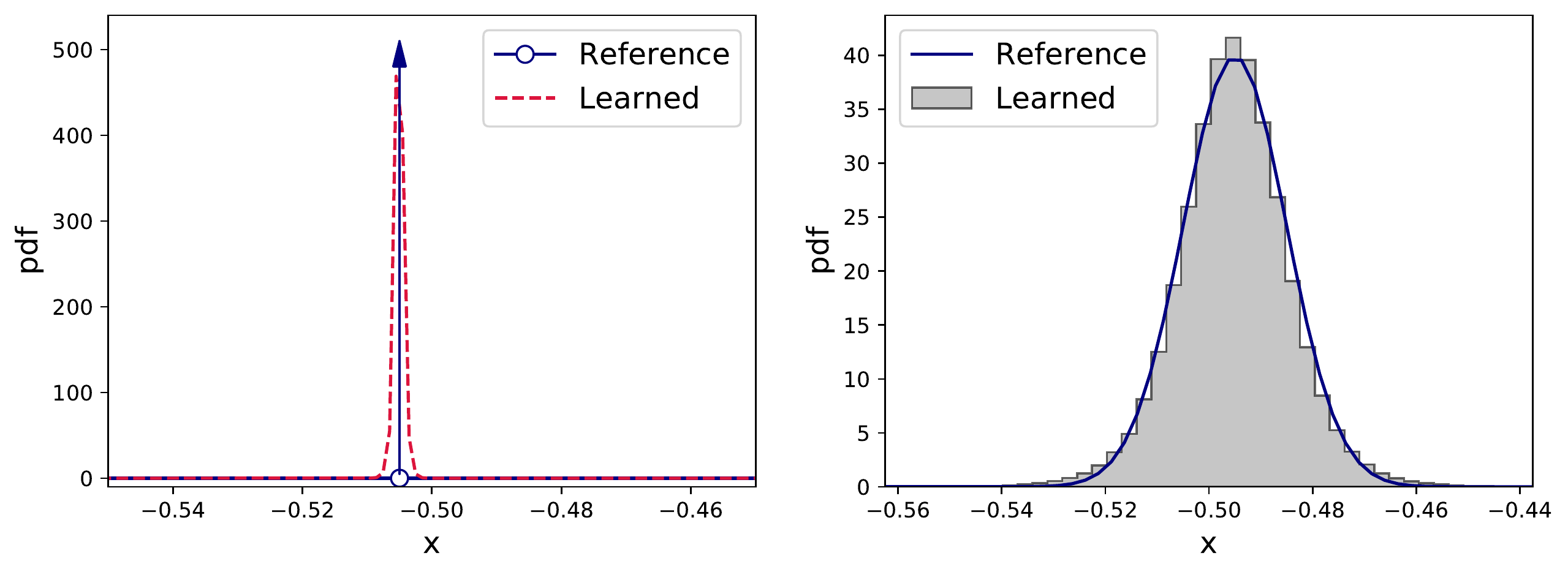}
  \caption{Stochastic oscillator \eqref{oscillator}: marginal
    distributions for $x_1$ (left) and $x_2$ (right) generated by the
    learned model  $\Gt_\Delta(\x)$ and the true model $\G_\Delta(\x)$
    at $\x=(-0.5, -0.5)$.}
\end{figure}

The mean and standard deviation of the system prediction by the
learned sFML model are shown in Figure \ref{fig:SO_show}, along with
those of the true solution. Good accuracy of the model prediction is
observed.
In Figure \ref{fig:SO_show2}, we present the marginal distributions of
the conditional probability
distribution produced by the learned stochastic flow map
$\Gt_\Delta(\x)$ at $\x=(-0.5, -0.5)$, in comparison with those of the true
conditional distribution from $\G_\Delta$. In this particularly case,
the marginal distribution in $x_1$ is a Dirac measure, as the equation
of $x_1$ is deterministic. We observe that the learned sFML model
produces very accurate prediction for such an extreme case. 
%
%

\section{Conclusions}
\label{sec:conclu}
In this paper, we proposed a general numerical framework for modeling unknown stochastic system by using observed trajectory data.
Our method is based on constructing an approximation of the underlying flow map of the unknown stochastic system. The constructed flow map consists
of two parts: a deterministic sub-map and a stochastic sub-map, both of which are learned by using the same observation data set.
While a standard feedforward DNN is  used to learn the deterministic sub-map, which predicts the conditional mean of the underlying
dynamics, a generative DNN is used to learn the stochastic sub-map, which captures the noisy dynamics. In this paper, we employ
GANs as the stochastic sub-map generator. By using a comprehensive set of numerical examples, we demonstrated that the proposed approach
is highly effective and accurate to model a variety of stochastic systems. The learned generative model is able to conduct long-term system predictions beyond the time domain of the training data.
More research is needed to further examine and understand the proposed method. For example, other generative models such as normalizing
flow can be used for the construction of the stochastic sub-map. These issues will be studied and reported in future work.

\bibliographystyle{siamplain}
\bibliography{references}
\end{document}